\documentclass[twoside,11pt,preprint]{article}

\usepackage{amsthm} 
\usepackage{jmlr2e_nolicensefooter}
\usepackage{amsmath}
\usepackage{amsfonts}
\usepackage{caption} 
\usepackage{subcaption} 
\usepackage{algorithm}
\usepackage{algpseudocode}
\usepackage{wrapfig}
\usepackage[section]{placeins} 
\usepackage{multirow}
\usepackage{booktabs}
\usepackage[table,xcdraw]{xcolor}         
\usepackage{rotating} 

\usepackage{makecell} 
\usepackage{hyperref}
\usepackage{fp} 
\usepackage{tikz}
\usepackage{lscape}

\newcommand{\dd}[2]{\frac{\partial #1}{\partial #2}}
\newcommand{\ddtfrac}[2]{\tfrac{\partial #1}{\partial #2}}

\jmlrheading{Vol}{2021}{x-x}{19/10/2021}{dd/mm/yyyy}{xxxx}{Timo M. Deist and Monika Grewal and Frank J.W.M. Dankers and Tanja Alderliesten and Peter A.N. Bosman}

\ShortHeadings{Multi-Objective Learning using HV Maximization}{Deist, Grewal et al.}
\firstpageno{1}

\begin{document}

\title{Multi-Objective Learning to Predict Pareto Fronts Using Hypervolume Maximization}

\author{\name Timo M. Deist\textsuperscript{1}\thanks{equal contribution}
        \email timo.deist@cwi.nl \\
       \AND
       \name Monika Grewal\textsuperscript{1}\textsuperscript{*}
       \email monika.grewal@cwi.nl \\
       \AND
       \name Frank J.W.M. Dankers\textsuperscript{2}
       \email f.j.w.m.dankers@lumc.nl \\
       \AND
       \name Tanja Alderliesten\textsuperscript{2}
       \email t.alderliesten@lumc.nl \\
       \AND
       \name Peter A.N. Bosman\textsuperscript{1, 3}
       \email peter.bosman@cwi.nl \\
       
       \addr \textsuperscript{1}Life Sciences \& Health Research Group,\\
       Centrum Wiskunde \& Informatica, Amsterdam, The Netherlands\\
       
       \addr \textsuperscript{2}Department of Radiation Oncology,\\
       Leiden University Medical Center, Leiden, The Netherlands\\
       
       \addr \textsuperscript{3}Faculty of Electrical Engineering, Mathematics and Computer Science,\\
       Delft University of Technology, Delft, The Netherlands
        }


\maketitle

\begin{abstract}
Real-world problems are often multi-objective with decision-makers unable to specify a priori which trade-off between the conflicting objectives is preferable. Intuitively, building machine learning solutions in such cases would entail providing multiple predictions that span and uniformly cover the Pareto front of all optimal trade-off solutions. We propose a novel approach for multi-objective training of neural networks to approximate the Pareto front during inference. In our approach, the neural networks are trained multi-objectively using a dynamic loss function, wherein each network's losses (corresponding to multiple objectives) are weighted by their hypervolume maximizing gradients. We discuss and illustrate why training processes to approximate Pareto fronts need to optimize on fronts of individual training samples instead of on only the front of average losses. Experiments on three multi-objective problems show that our approach returns outputs that are well-spread across different trade-offs on the approximated Pareto front without requiring the trade-off vectors to be specified a priori. Further, results of comparisons with the state-of-the-art approaches highlight the added value of our proposed approach, especially in asymmetric Pareto fronts.
\end{abstract}


\section{Introduction}
\label{sec:introduction}
Multi-objective (MO) optimization refers to finding Pareto optimal solutions according to multiple, often conflicting, objectives. In MO optimization, a solution is called Pareto optimal if none of the objectives can be improved without a simultaneous detriment in performance on at least one of the other objectives \citep{van2000multiobjective}. MO optimization is used for MO decision-making in many real-world applications \citep{stewart2008real} e.g., e-commerce recommendation \citep{lin2019pareto}, treatment plan optimization \citep{maree2019evaluation, Mller2017MulticriteriaPO}, and aerospace engineering \citep{oyama2002multiobjective}. In this paper, we focus on learning-based MO decision-making i.e., MO training of machine learning (ML) models such that MO decision-making is possible during inference. Furthermore, we specifically focus on generating Pareto fronts\footnote{Note that only \emph{near} Pareto front solutions can be generated during inference due to the generalization gap between training and inference.} (the Pareto front is the set of losses corresponding to all Pareto optimal solutions) for each sample separately during inference because decisions are made on a per-sample basis.

The most straightforward approach for MO optimization is linear scalarization, i.e., employing single-objective formulations of the problem as linear combinations of different objectives according to scalarization weights. The scalarization weights are based on the desired trade-off between multiple objectives which is often referred to as `user-preference'. A major issue with linear scalarization is that the user-preferences cannot always be straightforwardly translated to linear scalarization weights. Recently proposed approaches have tackled this issue and find solutions on the Pareto front (of average losses) for conflicting objectives according to a pre-specified user-preference vector \citep{lin2019paretomtl, mahapatra2020multi}. However, in many real world problems, the user-preference vector cannot be known a priori and decision-making is only possible \emph{a posteriori}, i.e., after multiple solutions are generated that are (near) Pareto optimal for a specific sample. For example, in neural style transfer \citep{gatys2016image} where photos are manipulated to imitate an art style from a selected painting, the user-preference between the amount of semantic information (the photo's content) and artistic style can only be decided by looking at multiple different resultant images on the Pareto front (Figure~\ref{fig:style_transfer_2d_example_corrected}). Therefore, to enable a posteriori decision-making per sample, an approximation set consisting of multiple solutions on the Pareto front needs to be generated, each representing a different trade-off between multiple objectives. For more information on a posteriori decision-making, please refer to \cite{hwang2012multiple}. 

\begin{figure}[h!]
    \centering
    \includegraphics[width=0.45\textwidth]{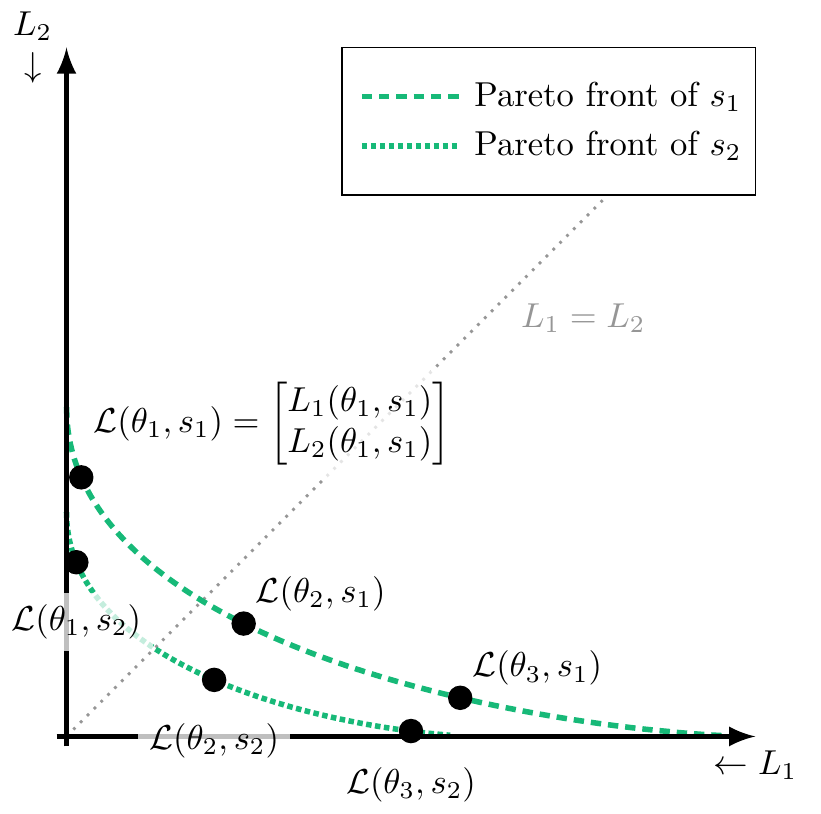}
    \caption{Pareto fronts in loss space for two samples and Pareto optimal predictions by three networks $\theta_{1}$, $\theta_{2}$, $\theta_{3}$. Both fronts are asymmetric in $L_{1}=L_{2}$.}
    \label{fig:pareto_fronts}
\end{figure}

Moreover, defining multiple trade-offs, typically by defining multiple scalarizations, to evenly cover the Pareto front is far from trivial, for example, if the Pareto front is asymmetric. Here, we define asymmetry in Pareto fronts as asymmetry in the distribution of Pareto optimal solutions in the objective space on either side of the 45$^{\circ}$-line, the line which represents the trade-off of equal marginal benefit along all objectives (Figure~\ref{fig:pareto_fronts}, notation will be explained in Section~\ref{sec:approach}). We demonstrate and discuss this further in Section~\ref{sec:experiments}. We will also demonstrate through experiments (in Section~\ref{sec:batch_vs_persample}) that MO training of an ML model for a trade-off between losses, which are averaged over multiple training samples, may not yield the same trade-off for each sample's losses in certain scenarios. Therefore, existing methods that train ML models to generate solutions which approximate the Pareto front \emph{for average losses of the training samples} are not sufficient when the goal is to generate models that approximate the Pareto front \emph{for losses of individual samples}.

Therefore, in learning-based a posteriori decision-making scenarios, it is crucial to have an MO approach for training ML models that can provide multiple diversely distributed outputs on the Pareto front per sample without requiring the user-preference vectors up front. Despite many developments in the direction of MO training of neural networks with pre-specified user-preferences, research in the direction of MO learning allowing for a posteriori decision-making is still scarce. In this paper, we present a novel method to multi-objectively train a set of neural networks to this end, leveraging the concept of hypervolume. Although we present our approach for training neural networks, the proposed formulation can be used for a wide range of ML models. 

The hypervolume (HV) -- the objective space dominated by a given set of solutions \citep{zitzler1999multiobjective} -- is a popular metric to compare the quality of different sets of solutions approximating the Pareto front. Theoretically, if the HV is maximal for a set of solutions, these solutions are on the Pareto front \citep{fleischer2003measure}. Additionally, HV not only encodes the proximity of a set of solutions to the Pareto front but also their diversity, which means that HV maximization provides a straightforward way for finding diverse solutions on the Pareto front. Therefore, we use hypervolume maximization for MO training of neural networks. We train the set of neural networks with a dynamically weighted combination of loss functions corresponding to multiple objectives, wherein the weight of each loss is based on the HV-maximizing gradients. In summary, our paper has the following main contributions:
\begin{itemize}
    \item An MO approach for training neural networks 
    \begin{itemize}
        \item using gradient-based HV maximization
        \item predicting Pareto optimal and diverse solutions on the Pareto front per sample without requiring specification of user-preferences
        \item enabling learning-based a posteriori decision-making.
    \end{itemize}
    \item An analysis highlighting the advantage of learning per sample over average-loss Pareto front approximations for differently shaped fronts.
    \item Experiments in real-world scenarios to demonstrate the added value of the proposed approach, specifically in asymmetric Pareto fronts.
\end{itemize}

\section{Related Work}
\textbf{MO optimization has been used in machine learning} for hyperparameter tuning of machine learning models \citep{koch2015efficient, avent2020automatic}, multi-objective classification of imbalanced data \citep{tari2020automatic}, and discovering the complete Pareto set starting from a single Pareto optimal solution \citep{ma2020efficient}. \cite{iqbal2020flexibo} used MO optimization for finding configurations of deep neural networks for conflicting objectives. \cite{gong2015multiobjective} proposed optimizing the weights of an autoencoder multi-objectively for finding the Pareto front of sparsity and reconstruction error. \cite{mao2020tchebycheff} used the Tchebycheff procedure for multi-objective optimization of a single neural network with multiple heads for multi-task text classification. Although we do not focus in these directions, our proposed approach can be used in similar applications.

\textbf{MO training of neural networks} has been researched widely, especially, for multi-task learning (MTL) \citep{sener2018multi, lin2019paretomtl, mahapatra2020multi, ma2020efficient, lin2020controllable, navon2020learning}. MO training of a set of neural networks such that their predictions approximate the Pareto front of multiple objectives is closely related to the work presented in this paper. Similar to our work, \cite{lin2019paretomtl,mahapatra2020multi} describe approaches with dynamic loss formulations to train multiple networks such that the predictions from these multiple networks together approximate the Pareto front. However, in these approaches, the trade-offs between conflicting objectives are required to be known in advance whereas our proposed approach does not require knowing the set of trade-offs beforehand. Other approaches \citep{navon2020learning,lin2020controllable} for MO training of neural networks involve training a ``hypernetwork'' to predict weights of another neural network based on a user-specified trade-off. A recent work has proposed to condition a neural network for an input user-preference vector to allow for predicting multiple points at the Pareto front during inference \citep{DBLP:journals/corr/abs-2103-13392}. While these approaches can approximate the Pareto front by iteratively predicting neural network weights or outputs based on multiple user-preference vectors, the process of sampling the user-preference vectors may still be intensive for an unknown Pareto front shape. Another key distinction of our approach from the abovementioned approaches is that these approaches learn to approximate the Pareto front of average losses for a set of training samples, while our approach learns to approximate the Pareto front for each individual sample in the training set. 

\textbf{Gradient-based HV maximization} is a key component of our proposed approach. \cite{miranda2016single} have described gradient-based HV maximization for single networks and formulated a dynamic loss function treating each sample's error as a separate loss. \cite{albuquerque2019multi} applied this concept for training in generative adversarial networks. HV maximization is also applied in reinforcement learning \citep{van2014multi,xu2020prediction}. While these approaches use HV maximizing gradients to optimize the weights of a single neural network, our proposed approach formulates a dynamic loss based on HV maximizing gradients for a \emph{set} of neural networks. Different from our approach, other concurrent approaches for HV maximization are based on transformation to $(m-1)D$ (where m is the number of objectives) integral by use of polar coordinates \cite{deng2019approximating}, random scalarization \citep{golovin2020random}, and q-Expected hypervolume improvement function \citep{daulton2020differentiable}.
\section{Approach}
\label{sec:approach}
MO learning of a network parameterized by a vector $\theta$ can be formulated as minimizing a vector of $n$ losses $\mathcal{L}(\theta,s_{k})=\left[L_{1}(\theta,s_{k}),\dots,L_{n}(\theta,s_{k})\right]$ for a given set of samples $S = \{s_{1},\dots, s_{k},\dots, s_{|S|}\}$. These loss functions form the loss space in which the subspace attainable per sample $s_{k}$ is bounded by its Pareto front, i.e., the combination of loss values $L_{j}(\theta,s_{k})$ of which none can be decreased without simultaneously increasing another loss (shown in Figure~\ref{fig:pareto_fronts} for two losses and two samples).
To learn multiple networks with loss vectors on each sample's Pareto front, we replace $\theta$ by a set of parameters $\Theta = \{\theta_{1},\dots,\theta_{p} \}$, where each parameter vector $\theta_{i}$ represents a network. The corresponding set of loss vectors is $\left\{\mathcal{L}(\theta_{1},s_{k}),\dots,\mathcal{L}(\theta_{p},s_{k}) \right\}$ and is represented by a stacked loss vector $\mathfrak{L}(\Theta,s_{k})=\left[\mathcal{L}(\theta_{1},s_{k}),\dots,\mathcal{L}(\theta_{p},s_{k}) \right]$.
\textbf{Our goal is to learn a set of $p$ networks such that loss vectors in $\mathfrak{L}(\Theta,s_{k})$ corresponding to the networks' predictions for sample $s_{k}$ lie on and span the Pareto front of loss functions for sample $s_{k}$}. That is, each network's loss vector is Pareto optimal and lies in a distinct subsection of the Pareto front. To achieve this goal, we train networks so that the loss subspace Pareto dominated by the networks' predictions is maximal.

\begin{figure*}
\begin{subfigure}{0.32\textwidth}
    \centering
    \includegraphics[width=\textwidth]{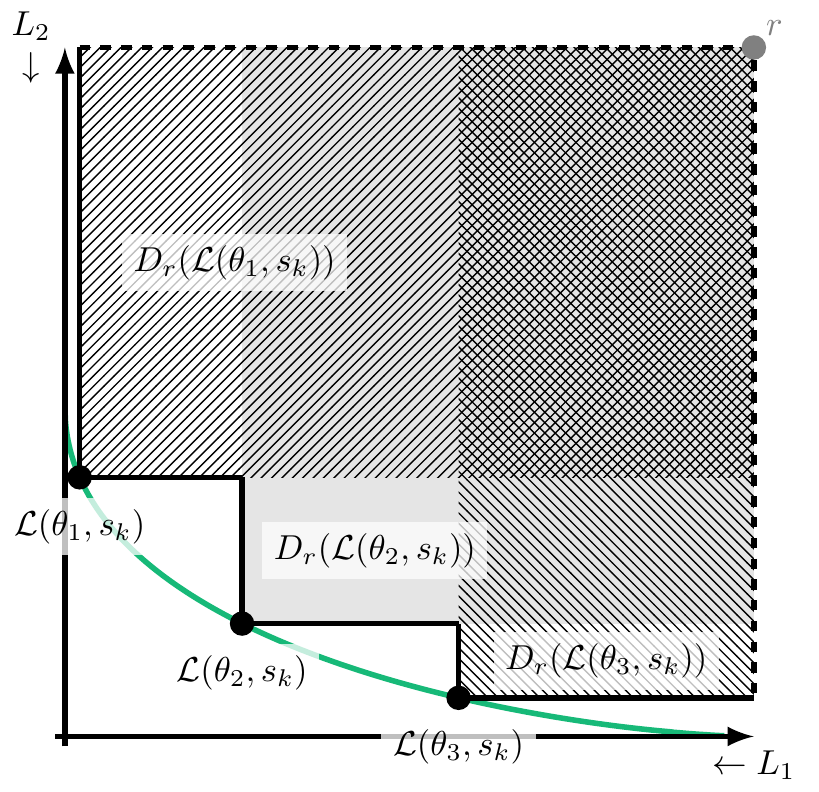}
    \caption{Dominated subspaces}
    \label{fig:hv_intuition}
\end{subfigure}
\begin{subfigure}{0.32\textwidth}
    \centering
    \includegraphics[width=\textwidth]{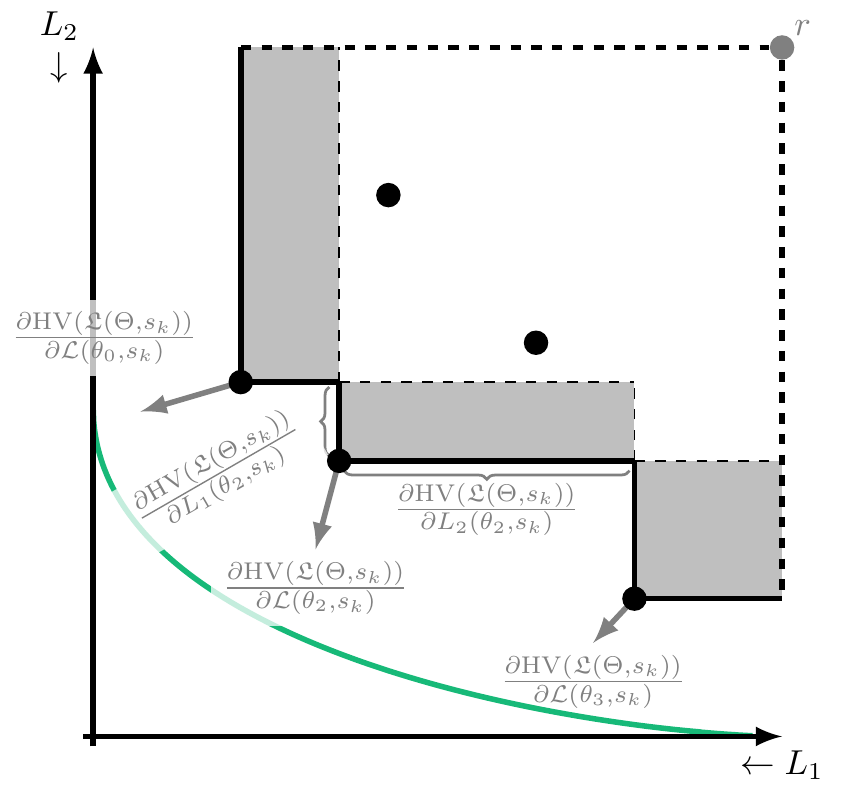}
    \caption{HV gradients}
    \label{fig:hv_grad_1front}
\end{subfigure}
\begin{subfigure}{0.32\textwidth}
    \centering
    \includegraphics[width=\textwidth]{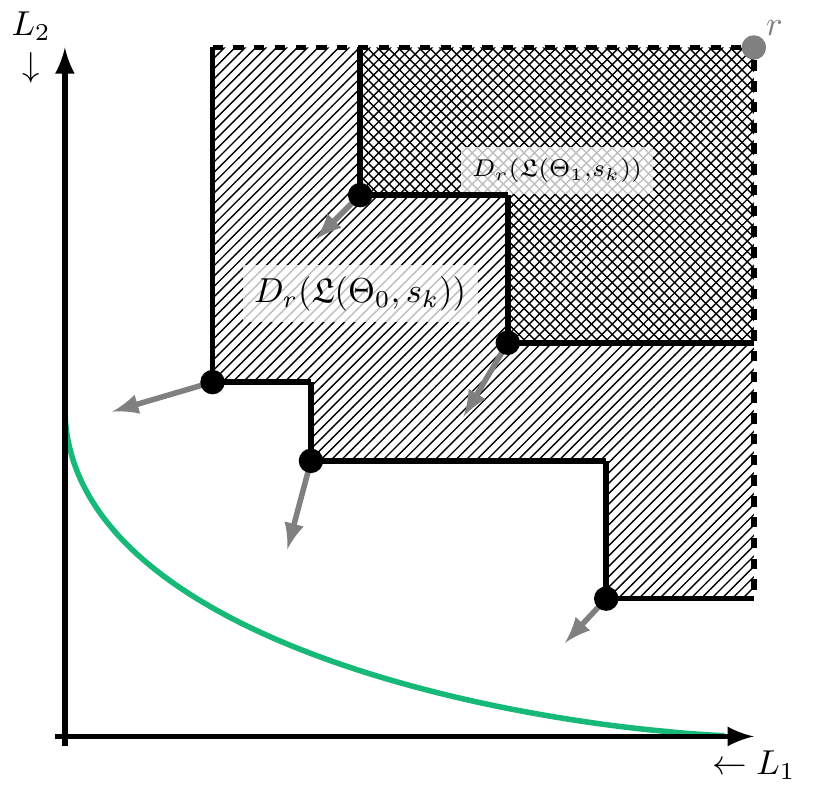}
    \caption{Domination-ranked fronts}
    \label{fig:hv_grad_intuition}
\end{subfigure}
    \centering
    \caption{\textbf{(a)} Three Pareto optimal loss vectors $\mathcal{L}(\theta_{i},s)$ on the Pareto front (green) with dominated subspaces $D_{r}(\mathcal{L}(\theta_{i},s_{k}))$ with respect to reference point $r$. The union of dominated subspaces is the dominated hypervolume (HV) of $\mathfrak{L}(\Theta,s_{k})$.
    \textbf{(b)} Gray markings illustrate the computation of the HV gradients $\dd{\mathrm{HV(\mathfrak{L}(\Theta,s))}}{\mathcal{L}(\theta_{i},s)}$ (gray arrows) in the three non-dominated solutions.
    \textbf{(c)} The same five solutions grouped into two domination-ranked fronts $\Theta_{0}$ and $\Theta_{1}$ with corresponding HV, equal to their dominated subspaces $D_{r}(\mathcal{L}(\theta_{i},s_{k}))$, and HV gradients.}
    \label{fig:hv}
\end{figure*}

\subsection{MO Learning by HV Maximization}
The HV of a loss vector  $\mathcal{L}(\theta_{i},s_{k})$ for a sample $s_{k}$ is the volume of the subspace $D_{r}(\mathcal{L}(\theta_{i},s_{k}))$ in loss space dominated by $\mathcal{L}(\theta_{i},s_{k})$. This is illustrated in Figure \ref{fig:hv_intuition}. To keep this volume finite, the HV is computed with respect to a reference point $r$ which bounds the space to the region of interest\footnote{The reference point is generally set to large coordinates in loss space to ensure that it is always dominated by all loss vectors.}. Subsequently, the HV of multiple loss vectors $\mathfrak{L}(\Theta,s_{k})$ is the HV of the union of dominated subspaces $D_{r}(\mathcal{L}(\theta_{i},s_{k})), \forall i \in \{1, 2, ..., p\}$. The MO learning problem to maximize the mean HV over all $|S|$ samples is as follows:
\begin{align}
    &\text{maximize} \frac{1}{|S|}\sum_{k=1}^{|S|}\mathrm{HV}\left(\mathfrak{L}(\Theta,s_{k})\right)\label{eq:max_mean_hv}
\end{align}
Concordantly, the update direction of gradient ascent for parameter vector $\theta_{i}$ of network $i$ is:
\begin{equation}
\dd{\frac{1}{|S|}\sum_{k=1}^{|S|}\mathrm{HV}(\mathfrak{L}(\Theta,s_{k}))}{\theta_{i}} \label{eq:training_set_gradients1}
\end{equation}
By exploiting the chain rule decomposition of HV gradients as described in \cite{emmerich2014time}, the update direction in Equation~\eqref{eq:training_set_gradients1} for parameter vector $\theta_{i}$ of network $i$ can be written as follows:
\begin{equation}
\frac{1}{|S|} \sum_{k=1}^{|S|}\dd{\mathrm{HV}\left( \mathfrak{L}(\Theta,s_{k}) \right)}{\mathcal{L}(\theta_{i},s_{k})}\cdot\dd{\mathcal{L}(\theta_{i},s_{k})}{\theta_{i}} \quad\forall i\in\{1,\dots,p\} \label{eq:training_set_gradients_dot_product}
\end{equation}

The dot product of $\dd{\mathrm{HV}\left( \mathfrak{L}(\Theta,s_{k}) \right)}{\mathcal{L}(\theta_{i},s_{k})}$ (the HV gradients with respect to loss vector $\mathcal{L}(\theta_{i},s_{k})$) in loss space,  
and $\dd{\mathcal{L}(\theta_{i},s_{k})}{\theta_{i}}$ (the matrix of loss vector gradients in the network $i$'s parameters $\theta_{i}$) in parameter space, can be decomposed to
\begin{equation}
 \frac{1}{|S|} \sum_{k=1}^{|S|}\sum_{j=1}^{n}\dd{\mathrm{HV}\left( \mathfrak{L}(\Theta,s_{k}) \right)}{L_{j}(\theta_{i},s_{k})}\dd{L_{j}(\theta_{i},s_{k})}{\theta_{i}} \quad\forall i\in\{1,\dots,p\} \label{eq:training_set_gradients}
\end{equation}
where $\dd{\mathrm{HV}(\mathfrak{L}(\Theta,s_{k}))}{L_{j}(\theta_{i},s_{k})}$ is the scalar HV gradient in the single loss function $L_{j}(\theta_{i},s_{k})$, and $\dd{L_{j}(\theta_{i},s_{k})}{\theta_{i}}$ are the gradients used in gradient descent for single-objective training of network $i$ for loss $L_{j}(\theta_{i}, s_{k})$. Based on Equation~\eqref{eq:training_set_gradients}, one can observe that mean HV maximization of loss vectors from a set of $p$ networks for $|S|$ samples can be achieved by weighting their gradient descent directions for loss functions $L_{j}(\theta_{i}, s_{k})$ with their corresponding HV gradients $\dd{\mathrm{HV}(\mathfrak{L}(\Theta,s_{k}))}{L_{j}(\theta_{i},s_{k})}$ for all $i$, $j$. In other terms, the MO learning of a set of $p$ networks can be achieved by minimizing\footnote{Minimizing (instead of maximizing) the dynamic loss function maximizes the HV because the reference point $r$ is in the positive quadrant (``to the right and above 0'').} the following dynamic loss function for each network $i$:
\begin{equation}
    \frac{1}{|S|}
    \sum_{k=1}^{|S|}\sum_{j=1}^{n}\dd{\mathrm{HV}\left( \mathfrak{L}(\Theta,s_{k}) \right)}{L_{j}(\theta_{i},s_{k})}L_{j}(\theta_{i},s_{k}) \quad\forall i\in\{1,\dots,p\} \label{eq:joint_loss_max_mean_hv_multi_sample}
\end{equation}
The computation of the HV gradients $\dd{\mathrm{HV}(\mathfrak{L}(\Theta,s_{k}))}{L_{j}(\theta_{i},s_{k})}$ is illustrated in Figure~\ref{fig:hv_grad_1front}. These HV gradients are equal to the marginal decrease in the subspace dominated only by $\mathcal{L}(\theta_{i},s_{k})$ when increasing $L_{j}(\theta_{i},s_{k})$.

\subsection{HV Maximization of Domination-Ranked Fronts}
\label{sec:domination_ranking}
A relevant caveat of gradient-based HV maximization is that HV gradients $\dd{\mathrm{HV}\left( \mathfrak{L}(\Theta,s_{k}) \right)}{L_{j}(\theta_{i},s_{k})}$ in strongly dominated solutions, i.e., solutions in the interior of the dominated HV, are zero \citep{emmerich2014time} because no movement in any direction will affect the HV (Figure~\ref{fig:hv_grad_1front}). Further, gradients in weakly dominated solutions are undefined \citep{emmerich2014time}. As a consequence, HV gradients cannot be used for optimizing (weakly or strongly) dominated solutions. To resolve this issue, we follow \cite{wang2017hypervolume}'s approach to gradient-based HV optimization. Other strategies to handle dominated solutions exist \citep{wang2017steering,deist2020multi}, but \cite{wang2017hypervolume} was selected as it only requires HV computation and non-dominated sorting and a comparison had shown that it performs similar to a competing approach \citep{deist2020multi}. The selected approach avoids the problem of dominated solutions by sorting all loss vectors into separate fronts $\Theta_{l}$ of mutually non-dominated loss vectors and optimizing each front separately (Figure \ref{fig:hv_grad_intuition}).
$l$ is the domination rank and $q(i)$ is the mapping of network $i$ to domination rank $l$. By maximizing the HV of each front, trailing fronts with domination rank $>0$ eventually merge with the non-dominated front $\Theta_{0}$ and a single front is maximized by determining optimal locations for each loss vector on the Pareto front.

Furthermore, we normalize the HV gradients $\ddtfrac{\mathrm{HV}\left(\mathfrak{L}(\Theta_{q(i)},s_{k}) \right)}{\mathcal{L}(\theta_{i},s_{k})}$ as in \cite{deist2020multi} such that their length in loss space is 1. The dynamic loss function including domination-ranking of fronts and HV gradient normalization is:
\begin{equation}
\frac{1}{|S|}
 \sum_{k=1}^{|S|}
 \sum_{j=1}^{n}
 \frac{1}{w_{i}}\dd{\mathrm{HV}\left(\mathfrak{L}(\Theta_{q(i)},s_{k}) \right)}{L_{j}(\theta_{i},s_{k})}
 L_{j}(\theta_{i},s_{k})
 \label{eq:final_joint_loss}
 \quad \forall i\in\{1,\dots,p\}
\end{equation}
where $w_{i} = \left\Vert \ddtfrac{\mathrm{HV}\left(\mathfrak{L}(\Theta_{q(i)},s_{k}) \right)}{\mathcal{L}(\theta_{i},s_{k})} \right\Vert$. The algorithm is summarized in Algorithm~\ref{alg:pseudocode}.

\begin{algorithm}[htbp]
\caption{Training networks $\Theta$ for Pareto front prediction by HV maximization of domination-ranked fronts}
\begin{algorithmic}
\State Initialize $p$ networks $\Theta=\{\theta_{1},\dots,\theta_{p}\}$
\For{each batch $\tilde{S}$}
    \For{each network $\theta_{i}$}
        \For{each sample $s_{k}\in \tilde{S}$}
            \State Compute loss vector $\mathcal{L}(\theta_{i},s_{k})$
        \EndFor
    \EndFor
    \For{each sample $s_{k}\in \tilde{S}$}
        \State Stack loss vectors $\mathcal{L}(\theta_{i},s_{k})$ into $\mathfrak{L}(\Theta,s_{k})$
        \State Sort $\mathfrak{L}(\Theta,s_{k})$ into multiple fronts $\mathfrak{L}(\Theta_{l},s_{k})$ by domination ranking (Section~\ref{sec:domination_ranking})
        \For{each front $l$}
            \State Compute loss weights $\dd{\mathrm{HV}\left(\mathfrak{L}(\Theta_{q(i)},s_{k}) \right)}{L_{j}(\theta_{i},s_{k})}\forall i,j$ using algorithm by
            \State \cite{emmerich2014time} 
        \EndFor
    \EndFor
    \For{each network $\theta_{i}$}
        \State Backpropagate on joint loss from Equation~\eqref{eq:final_joint_loss}
    \EndFor
    \State Update $\Theta$ by stepping into gradient direction
\EndFor
\end{algorithmic}
\label{alg:pseudocode}
\end{algorithm}

\subsection{Average vs per-sample dynamic loss formulations}
\label{sec:batch_vs_persample}
The dynamic loss in \eqref{eq:final_joint_loss} maximizes the HV for each sample's loss vectors by weighting the loss vectors with the corresponding HV maximizing gradients -- which is the ideal objective for MO training of the set of neural networks. However, this means that one network $\theta_{i}$ is not necessarily trained exclusively for a specific trade-off. Instead, across different samples $s_{k}$, one network could generate outputs corresponding to different trade-offs. This may have three practical implications: 1) HV maximizing gradients need to be calculated for each sample, which could be expensive, 2) the predictions on the different unseen samples may not follow a similar ordering along the Pareto front and a surrogate method might be required to ascertain the trade-offs of the different outputs comprising the approximated Pareto front during inference, 3) since the networks may learn a different trade-off corresponding to different samples, the optimization problem underlying the joint training of the $p$ learners becomes more complex.

A simple workaround to avoid the abovementioned implications is to reformulate the dynamic loss in \eqref{eq:final_joint_loss} such that the HV maximizing gradients are calculated for average losses of multiple samples. Existing approaches \citep{sener2018multi, lin2019paretomtl, mahapatra2020multi} for MO training of neural networks also optimize for average losses (but not their HV). With the use of average losses, the dynamic~loss~\eqref{eq:final_joint_loss} would be simplified to:
\begin{equation}
 \sum_{j=1}^{n}\frac{1}{w_{i}}\dd{\mathrm{HV}\left(\overline{\mathfrak{L}(\Theta_{q(i)},S)} \right)}{\overline{L_{j}(\theta_{i},S)}}\overline{L_{j}(\theta_{i},S)} \label{eq:joint_loss_max_hv_of_mean_losses}\quad\forall i\in\{1,\dots,p\}
\end{equation}
where the different loss vectors are replaced by their corresponding averages over the training samples (or a batch of samples when training batchwise). Changing the formulation from \eqref{eq:final_joint_loss} to \eqref{eq:joint_loss_max_hv_of_mean_losses} has two practical benefits: 1) the number of HV gradient computations $\ddtfrac{\mathrm{HV}\left(\cdot \right)}{{L_{j}(\cdot)}}$ reduces from $n|S|$ to $n$, which gives a considerable speed-up especially in the cases of large batchsize and shallow neural network training, 2) since the weights of a neural network are updated according to one trade-off (dynamically estimated as HV gradients) for all samples, the prediction ordering remains same across samples during inference making the presentation of approximated Pareto fronts to the decision-maker easier. However, this simplification in \eqref{eq:final_joint_loss} may lead to imperfect Pareto front approximations on individual samples in certain cases, examples of which are outlined below.
\begin{figure}
    \centering
    \begin{tabular}{ccc}
        & \textbf{Training per sample} & \textbf{Training on average losses} \\
    & $\max \tfrac{1}{|S|}\sum_{k=1}^{|S|} \mathrm{HV}\left(\mathfrak{L}(\Theta,s_{k})\right)$ & $\max \mathrm{HV}\left(\overline{\mathfrak{L}(\Theta,S)}\right)$  \\
    & (Dynamic~loss~\eqref{eq:final_joint_loss}) & (Dynamic~loss~\eqref{eq:joint_loss_max_hv_of_mean_losses})  \\
    \multirow[t]{1}{*}{
    \rotatebox[origin=c]{90}{\textbf{Strictly convex}}}
    &
\begin{subfigure}{0.30\textwidth}
    \centering
    \includegraphics[width=\textwidth]{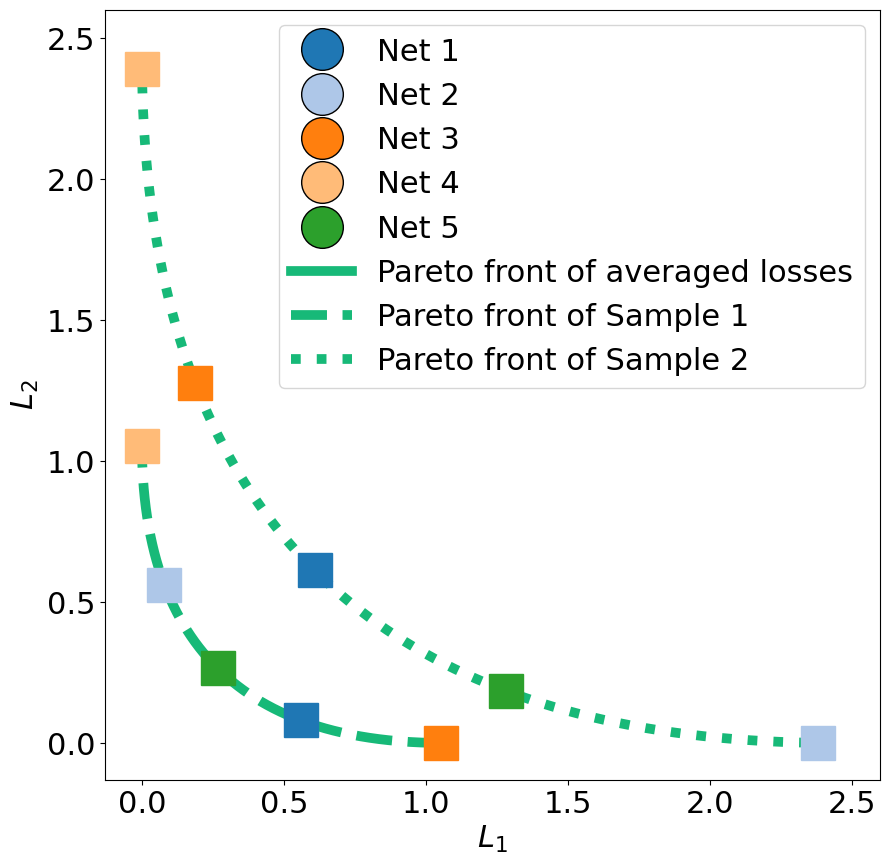}
    \caption{}
    \label{fig:opt_ex_loss_per_sample_convex_higamo}
\end{subfigure}
    &
\begin{subfigure}{0.30\textwidth}
    \centering
    \includegraphics[width=\textwidth]{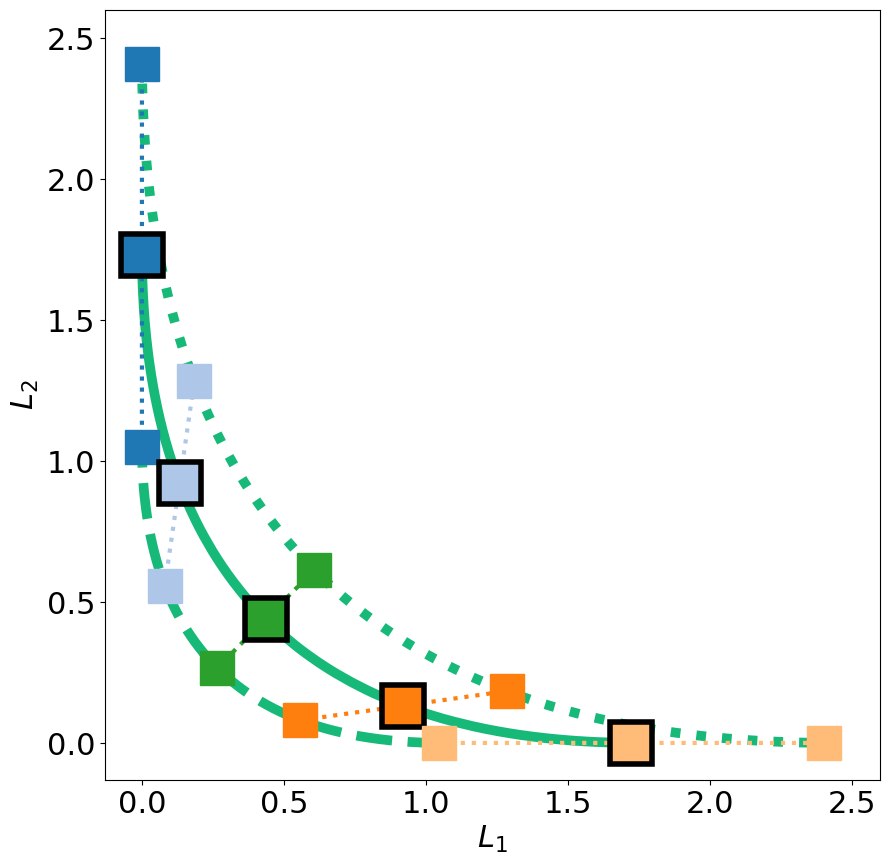}
    \caption{}
    \label{fig:opt_ex_mean_losses_convex_higamo}
\end{subfigure}
    \\    
    \multirow[t]{1}{*}{
    \rotatebox[origin=c]{90}{\textbf{Linear}}}
    &
\begin{subfigure}{0.30\textwidth}
    \centering
    \includegraphics[width=\textwidth]{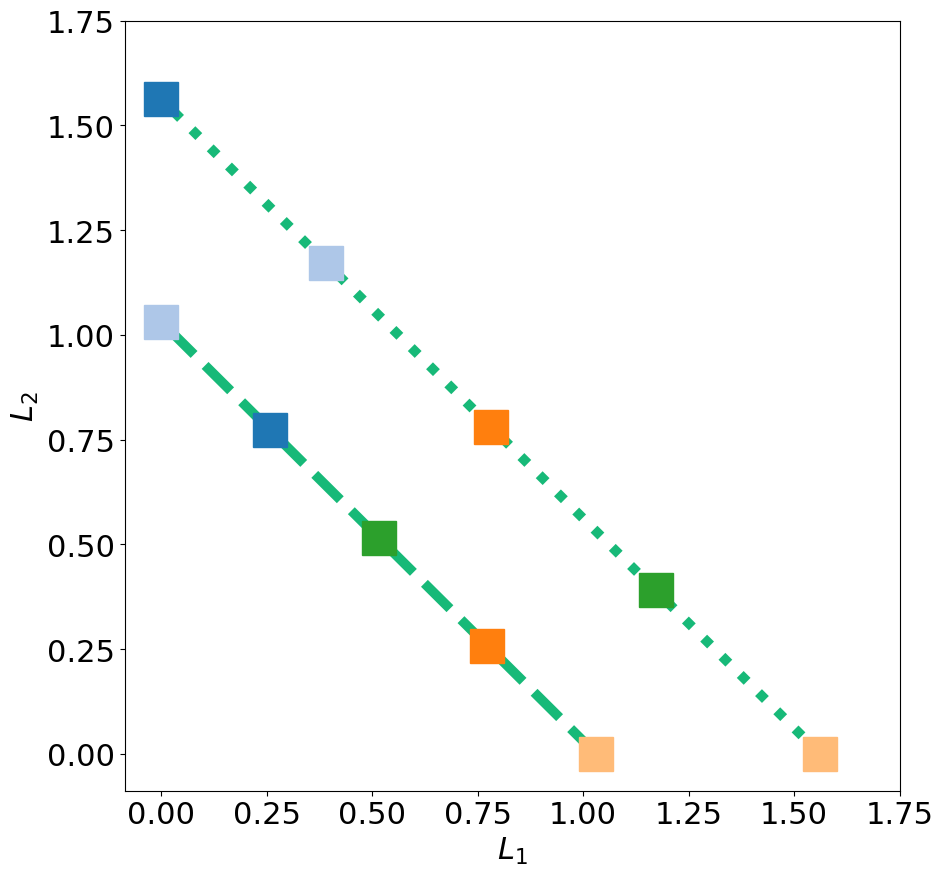}
    \caption{}
    \label{fig:opt_ex_loss_per_sample_line_higamo}
\end{subfigure}
    &
\begin{subfigure}{0.30\textwidth}
    \centering
    \includegraphics[width=\textwidth]{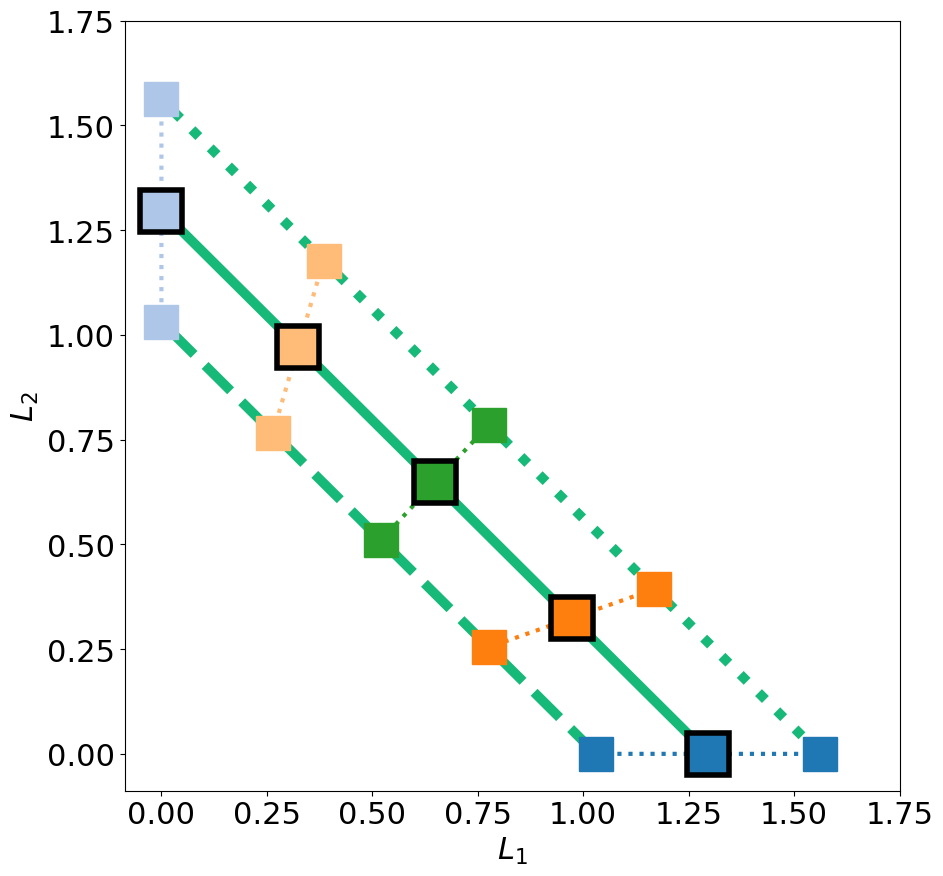}
    \caption{}
    \label{fig:opt_ex_mean_losses_line_higamo}
\end{subfigure}
\\
    \multirow[t]{1}{*}{
    \rotatebox[origin=c]{90}{\textbf{Non-convex}}}
    &
\begin{subfigure}{0.30\textwidth}
    \centering
    \includegraphics[width=\textwidth]{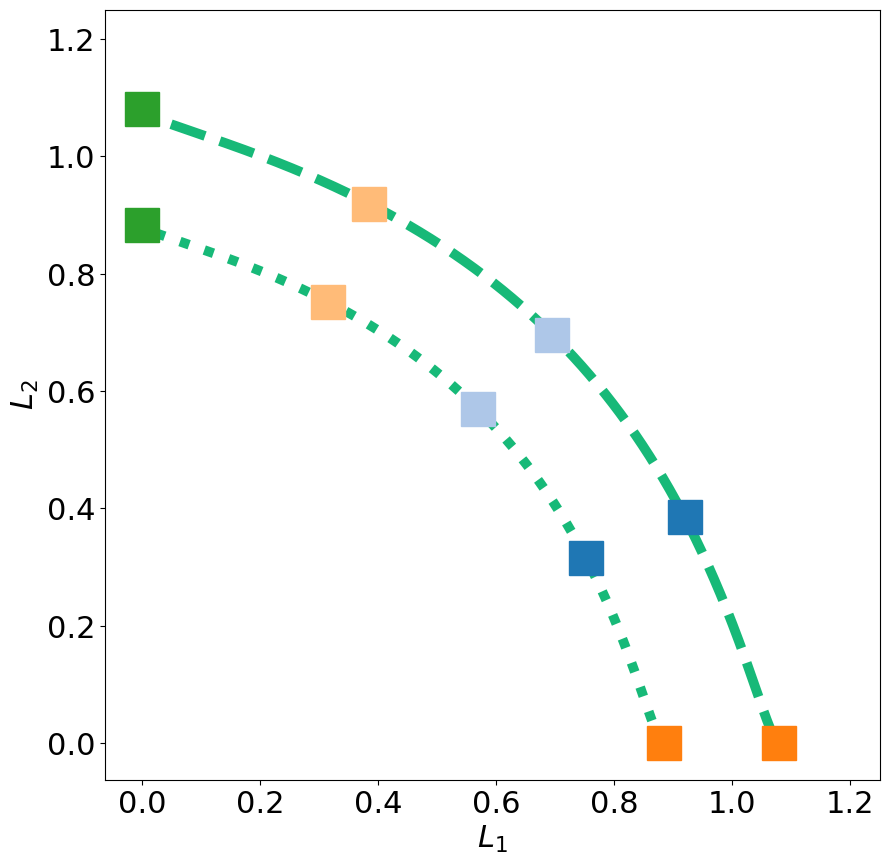}
    \caption{}
    \label{fig:opt_ex_loss_per_sample_concave_higamo}
\end{subfigure}
&
\begin{subfigure}{0.30\textwidth}
    \centering
    \includegraphics[width=\textwidth]{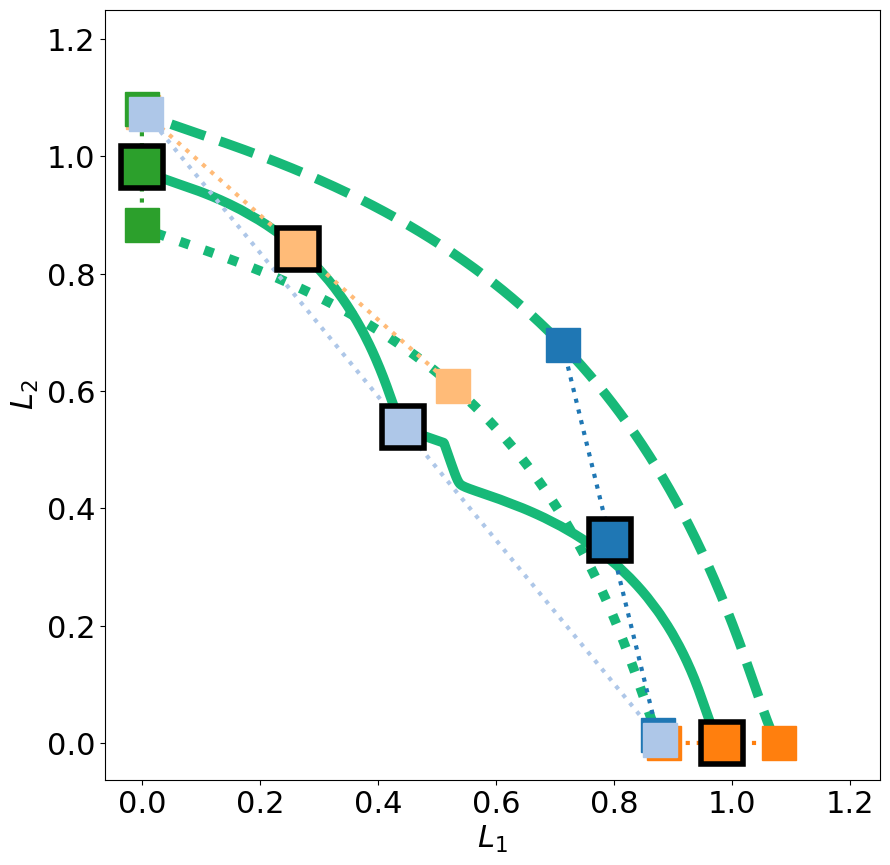}
    \caption{}
    \label{fig:opt_ex_mean_losses_concave_higamo}
\end{subfigure}

\\
    \end{tabular}    
    \caption{HV maximization of each sample's losses (left) and of average losses (right) for a two sample problem with different Pareto front shapes (rows).
    }
    \label{fig:optimization_example_higamo_hv}
\end{figure}

Consider the following three cases of MO training of five neural networks to predict strictly convex, linear, or non-convex fronts for two training samples ($k \in \{1, 2\} $) each.
\begin{enumerate}

\item \textbf{Strictly convex:} given the centres $x_{k}^{1}$ and $x_{k}^{2}$ of two circles as inputs, each neural network $\theta_{i}$ in the set outputs the coordinates $z_{k}$ 
with minimal squared Euclidean distance from both the circle centres such that the set of the outputs approximate the Pareto front. This results in a strictly convex Pareto front. The corresponding losses to minimize are:
\begin{equation*}
    L_{1} = \left\lVert z_{k} - x_{k}^{1} \right\rVert_{2}^{2}  \quad L_{2} = \left\lVert z_{k} - x_{k}^{2}\right\rVert_{2}^{2}
\end{equation*}
\item \textbf{Linear (simultaneously convex \& concave):} For the problem specified above, if the Euclidean distance from the two centres is minimized, the shape of the corresponding Pareto front will become linear. The losses to minimize are:
\begin{equation*}
    L_{1} = \left\lVert z_{k} - x_{k}^{1} \right\rVert_{2}  \quad L_{2} = \left\lVert z_{k} - x_{k}^{2}\right\rVert_{2}
\end{equation*}
\item \textbf{Non-convex:} To generate a non-convex front\footnote{It is named \emph{non-convex} instead of \emph{concave} or \emph{strictly concave} because it is concave for the most part but, when $L_{1}$ or $L_{2}$ are close to 0, the front is locally convex.}, we adapted the optimization example used in  \cite{lin2019paretomtl} and \cite{mahapatra2020multi} to a learning problem with two samples. The problem is formulated as: given a scalar input $x_{k}$, each network outputs a scalar $z_{k}$ 
such that they minimize the following losses simultaneously. 
\begin{equation*}
    L_{1} = x_{k} \left(1- \exp{\left(-\left\lVert z_{k}-1\right\rVert_{2}^{2}\right)} \right)  \, L_{2} = x_{k} \left(1- \exp{\left(-\left\lVert z_{k}+1\right\rVert_{2}^{2}\right)} \right)
\end{equation*}

\end{enumerate}

Figure~\ref{fig:optimization_example_higamo_hv} (left column) shows results for maximizing the average of each sample's HV (Dynamic~loss~\eqref{eq:final_joint_loss}) and Figure~\ref{fig:optimization_example_higamo_hv} (right column) shows results for maximizing the HV of average losses (Dynamic~loss~\eqref{eq:joint_loss_max_hv_of_mean_losses}). The approximated Pareto fronts per sample (small rectangles with different colors per network $\theta_{i}$) and each sample's Pareto fronts (dashed/dotted green lines) are displayed. Additionally on the right, because the networks are trained on the average losses, the averaged predictions (single large rectangles with black borders) and the corresponding Pareto front (solid green line) are shown. 
Figures~\ref{fig:optimization_example_higamo_hv} and \ref{fig:optimization_example_existing_methods} show each method's best observed performance, either maximal average HV across samples or maximal HV on average losses, over a tuning grid (Table~\ref{tab:tuning_info} in the appendix).
  
For the strictly convex and linear cases (Figure~\ref{fig:optimization_example_higamo_hv}, top and middle rows), minimizing either of the Dynamic losses~\eqref{eq:final_joint_loss} or \eqref{eq:joint_loss_max_hv_of_mean_losses} yields well-spread outputs across each sample's Pareto front. In case of a linear front (Figure~\ref{fig:opt_ex_mean_losses_line_higamo}), however, a decent spread cannot always be guaranteed because the same HV of average losses with irregular spread can be achieved, for example, by shifting the orange outputs to the left along Sample~1's front and to the right along Sample~2's front.
For the non-convex Pareto fronts, minimizing Dynamic loss~\eqref{eq:joint_loss_max_hv_of_mean_losses} does not yield maximal HV per sample (small rectangles) as is clear by the uneven spread of the predictions across each sample's Pareto front (Figure~\ref{fig:opt_ex_mean_losses_concave_higamo}). Minimizing Dynamic loss~\eqref{eq:final_joint_loss} does lead to well-distributed predictions per sample across the Pareto front irrespective whether the front is strictly convex, linear or non-convex (Figure~\ref{fig:optimization_example_higamo_hv}, left column).

\subsubsection{The differences between average and per-sample formulations are not specific to HV maximization}
\label{sec:differences not specific to HV}
\begin{figure}
    \centering
    \begin{tabular}{cccc}
    & \multicolumn{3}{c}{\textbf{Training on average losses}}\\
    \cmidrule{2-4}
    & \textbf{Linear scalarization} & \textbf{Pareto MTL}  & \textbf{EPO}  \\
    \multirow[t]{1}{*}{
    \rotatebox[origin=c]{90}{\textbf{Strictly convex}}
    }
    &
\begin{subfigure}{0.30\textwidth}
    \centering
    \includegraphics[width=\textwidth]{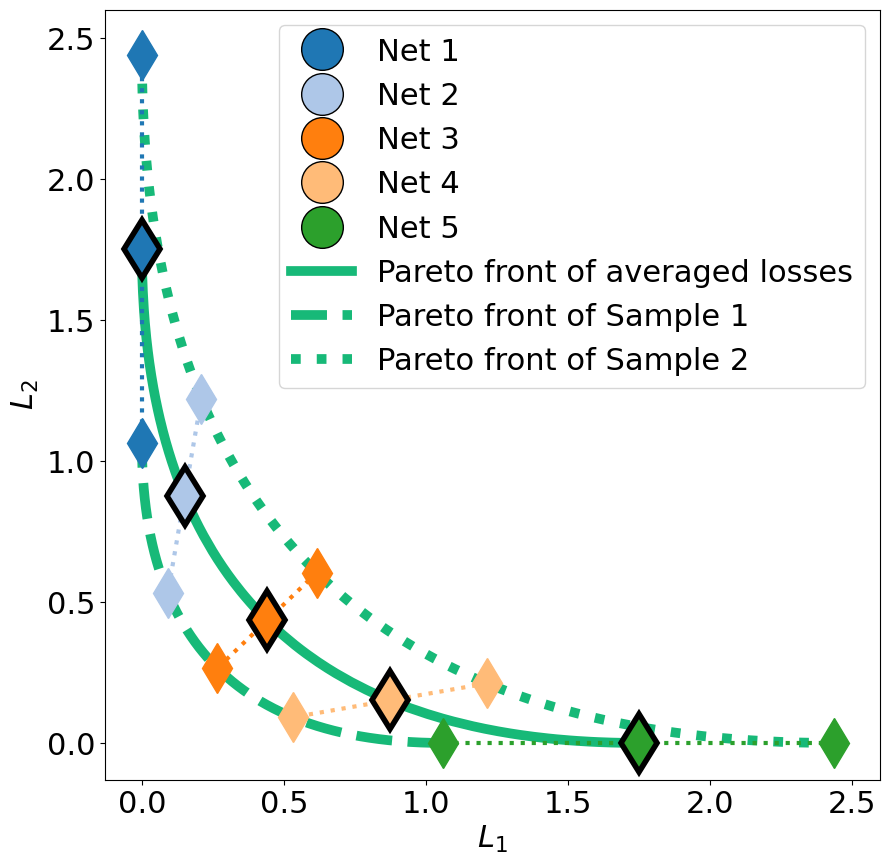}
    \caption{}
    \label{fig:opt_ex_mean_losses_convex_linscal}
\end{subfigure}  
    &
\begin{subfigure}{0.30\textwidth}
    \centering
    \includegraphics[width=\textwidth]{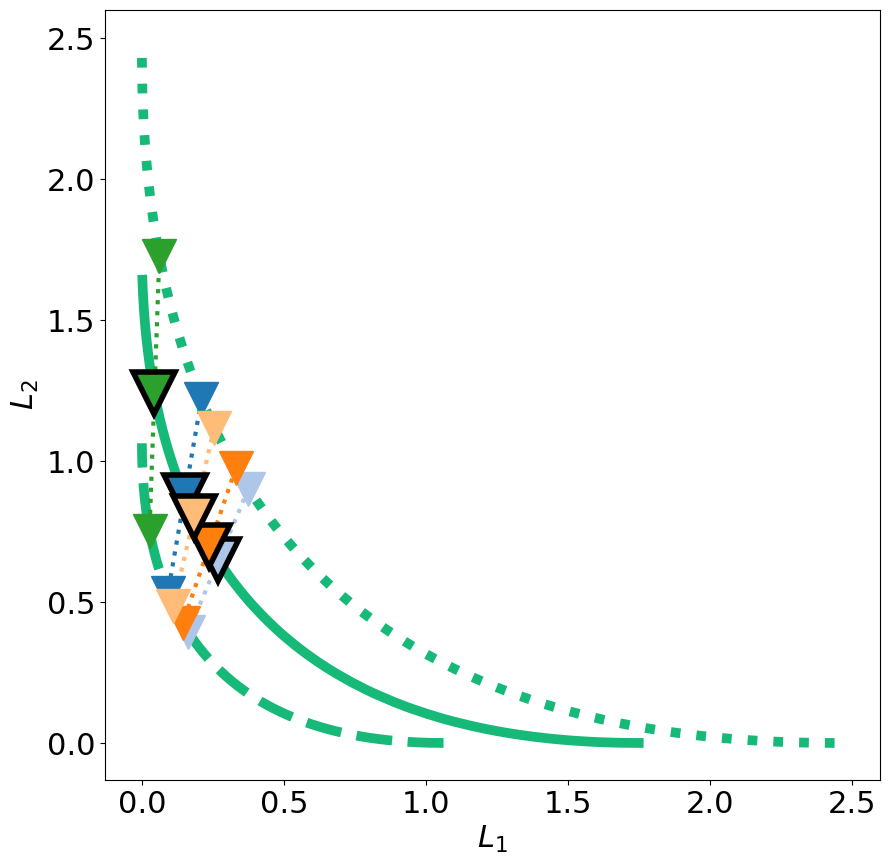}
    \caption{}
    \label{fig:opt_ex_mean_losses_convex_paretomtl}
\end{subfigure}
    &
\begin{subfigure}{0.30\textwidth}
    \centering
    \includegraphics[width=\textwidth]{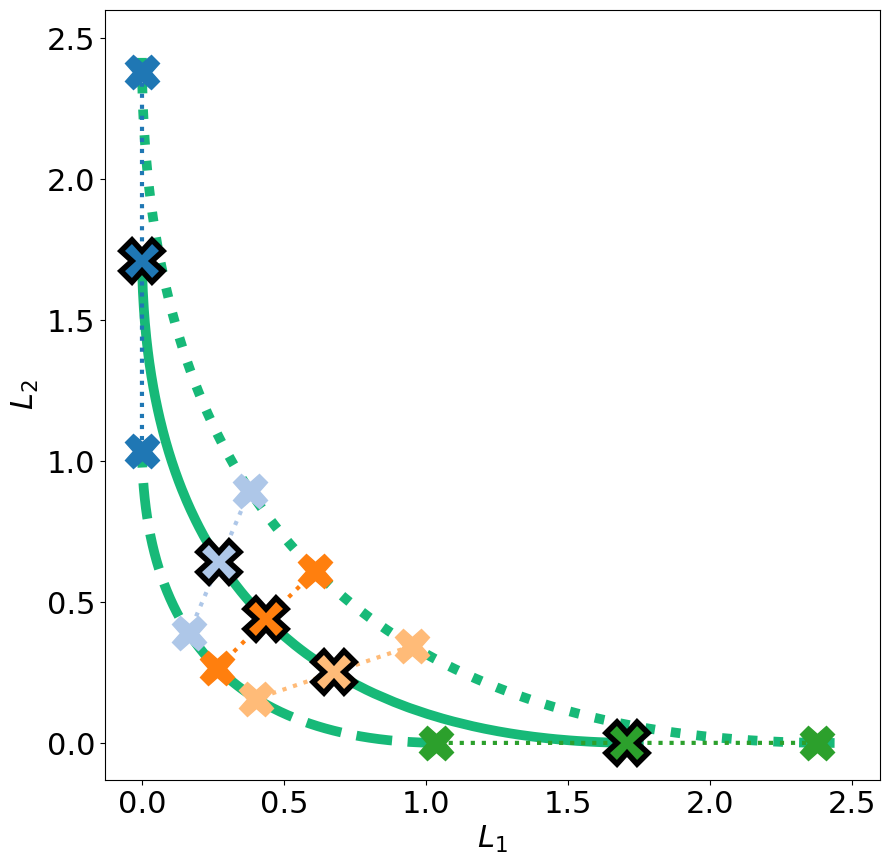}
    \caption{}
    \label{fig:opt_ex_mean_losses_convex_epo}
\end{subfigure}
    \\   
    \multirow[t]{1}{*}{
    \rotatebox[origin=c]{90}{\textbf{Linear} }
    }
    &
\begin{subfigure}{0.30\textwidth}
    \centering
    \includegraphics[width=\textwidth]{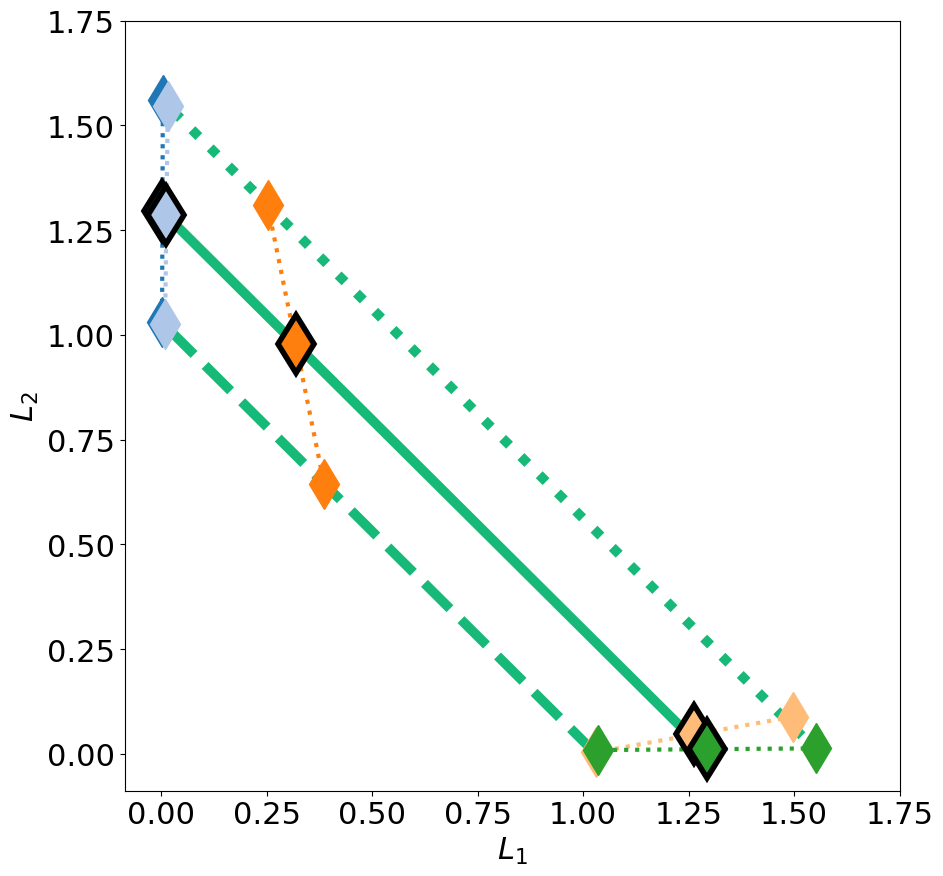}
    \caption{}
    \label{fig:opt_ex_mean_losses_line_linscal}
\end{subfigure}
    &
\begin{subfigure}{0.30\textwidth}
    \centering
    \includegraphics[width=\textwidth]{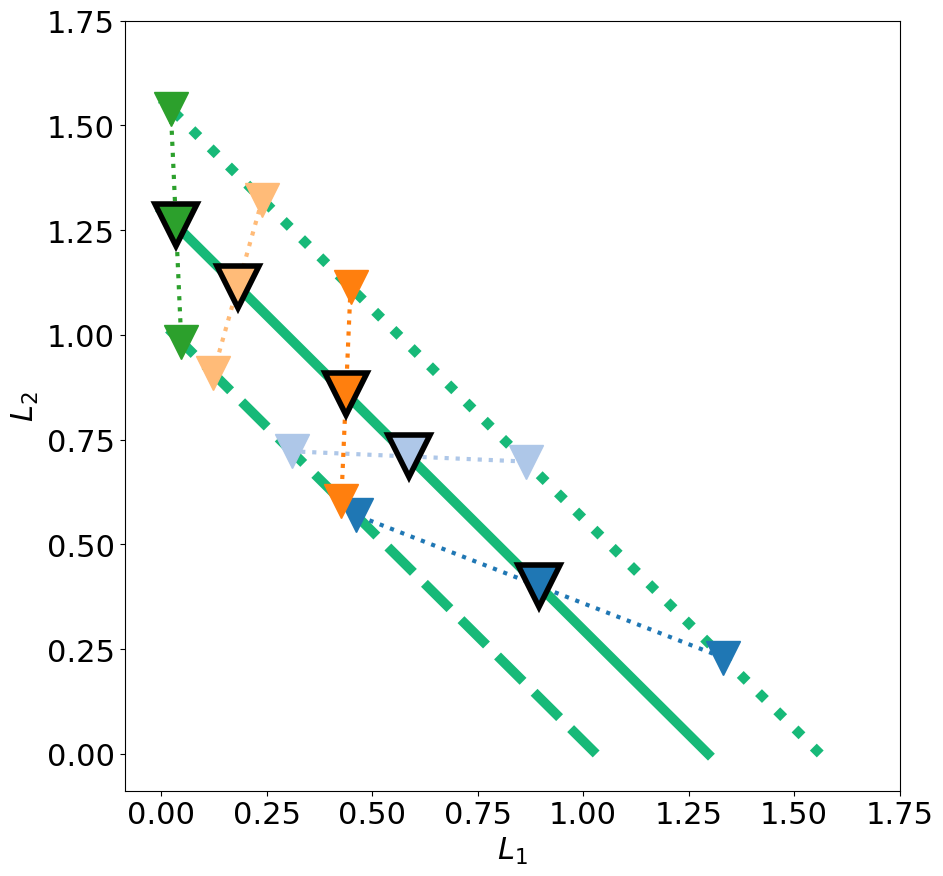}
    \caption{}
    \label{fig:opt_ex_mean_losses_line_paretomtl}
\end{subfigure}
    &
\begin{subfigure}{0.30\textwidth}
    \centering
    \includegraphics[width=\textwidth]{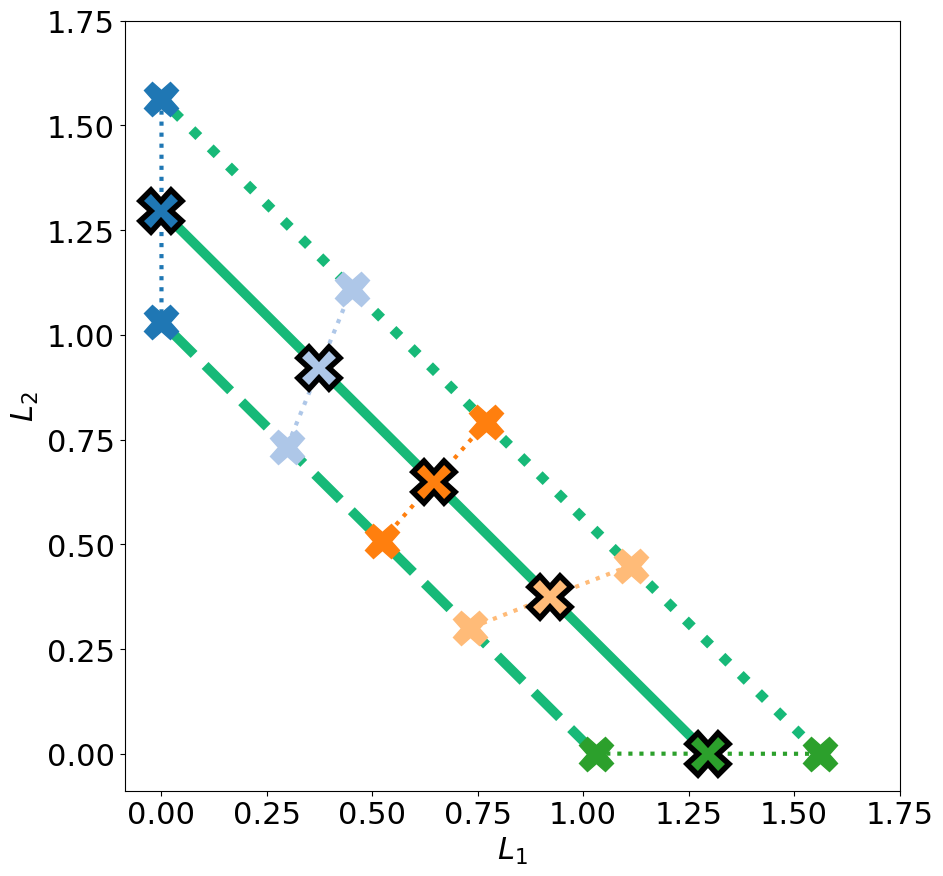}
    \caption{}
    \label{fig:opt_ex_mean_losses_line_epo}
\end{subfigure}
\\
    \multirow[t]{1}{*}{
    \rotatebox[origin=c]{90}{\textbf{Non-convex}} 
    }
    &
\begin{subfigure}{0.30\textwidth}
    \centering
    \includegraphics[width=\textwidth]{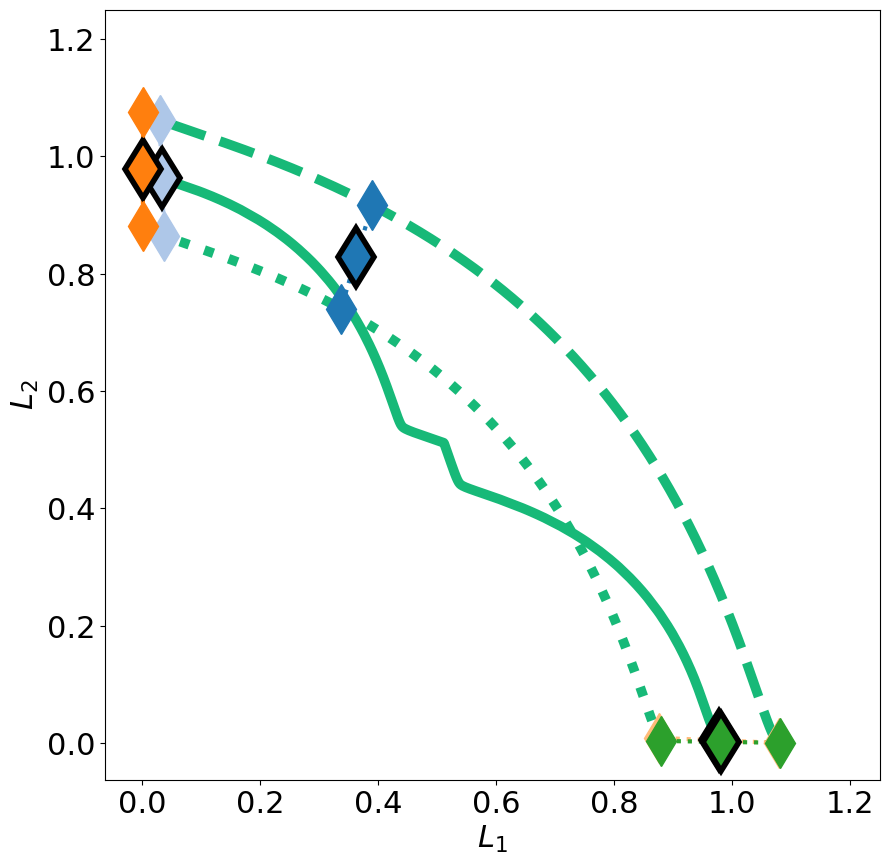}
    \caption{}
    \label{fig:opt_ex_mean_losses_concave_linscal}
\end{subfigure}  
    &
\begin{subfigure}{0.30\textwidth}
    \centering
    \includegraphics[width=\textwidth]{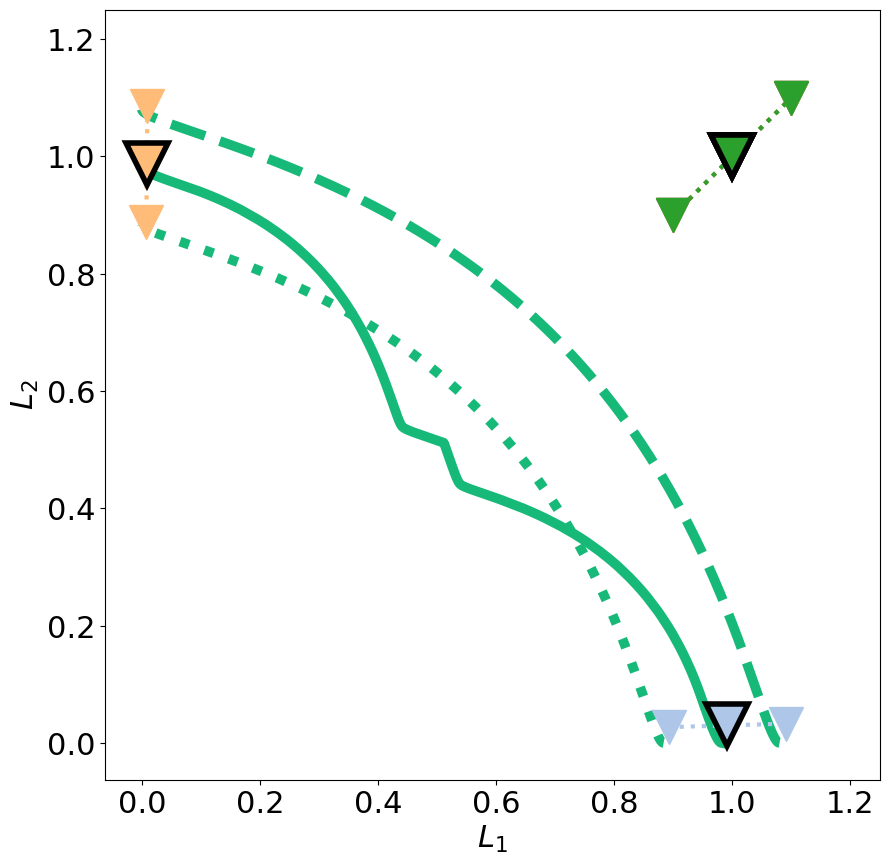}
    \caption{}
    \label{fig:opt_ex_mean_losses_concave_paretomtl}
\end{subfigure}
    &
\begin{subfigure}{0.30\textwidth}
    \centering
    \includegraphics[width=\textwidth]{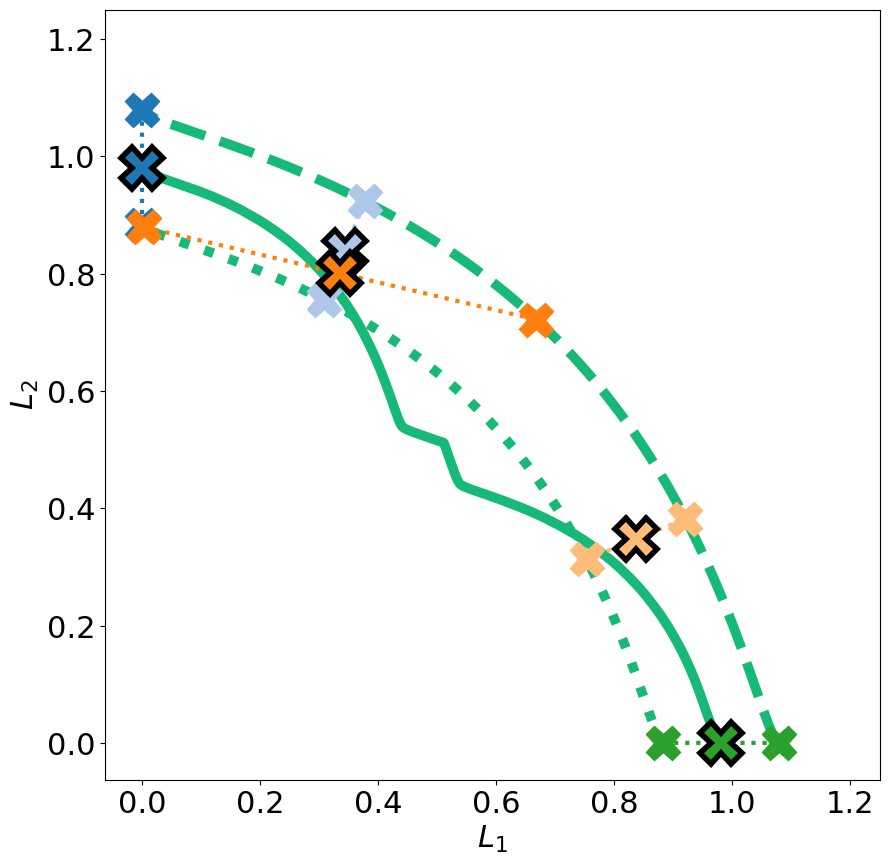}
    \caption{}
    \label{fig:opt_ex_mean_losses_concave_epo}
\end{subfigure}
 \\
    \end{tabular}    
    \centering
    \caption{Optimization in average loss space for a two sample problem with
    Pareto front shapes (rows) using Linear scalarization (left), Pareto MTL (center), and EPO (right).
    }
    \label{fig:optimization_example_existing_methods}
\end{figure}

We investigated the abovementioned cases with three other existing approaches: a baseline linear scalarization approach, i.e., linear combination of losses with fixed weights, and two state-of-the-art approaches PMTL and EPO using dynamic loss formulations to achieve a priori specified user-preferences on the Pareto front. All three methods attempt to learn an approximation of the Pareto front of average losses. Corresponding results are shown in Figure~\ref{fig:optimization_example_existing_methods}. It can be seen that, for linear and non-convex Pareto fronts, Pareto MTL and Linear scalarization are not guaranteed to achieve the desired user-preferences on each sample's front when training models for the chosen user-preferences on the Pareto front of average losses. EPO returns outputs per sample following the chosen user-preferences in the strictly convex and linear cases, but also fails in the non-convex setting.
These findings thus corroborate that training for predictions with specific trade-offs/user-preferences on average losses does not translate to predictions for individual samples for all Pareto front curvatures.

\subsubsection{Training for trade-offs on the average loss Pareto front need not translate to predictions on each sample's front with the same trade-off}
The above results show that, in the strictly convex case, trade-offs on the average Pareto front either given by HV gradients in our approach, weights from dynamic loss formulations (EPO, Pareto MTL), or fixed weights (Linear scalarization) correspond to predictions with the same trade-offs on the individual samples' Pareto fronts. To understand why this happens for the chosen strictly convex case but not for the linear and non-convex case, one should consider the relationship between trade-offs on average and the individual samples' fronts.
In this strictly convex case, almost all given trade-offs on the strictly convex average loss front can be realized by averaging the \emph{same} trade-offs on each sample's strictly convex front. The blue solution is an exception as it is slightly above the average loss front. 
For the linear case (Figure~\ref{fig:opt_ex_mean_losses_line_higamo}) however, \emph{many} pairs of trade-offs on the two samples' fronts can be averaged to yield the given trade-off on the average loss front. 
Similarly, in the non-convex case, predictions with even \emph{opposing} trade-offs on concave sections of the two samples' Pareto fronts can be averaged to yield the desired trade-off on the Pareto front of average losses. Consider, for example, the two small light blue markers at the extremes of each sample's front averaging to a almost 1:1 trade-off on the average loss front in Figure~\ref{fig:opt_ex_mean_losses_concave_higamo}. At this example it can be observed that optimizing for a specific trade-off on the \emph{average loss} front need not result in that specific trade-off on \emph{each sample's} front. 

\subsubsection{Strict convexity of individual samples' Pareto fronts does not guarantee the average loss formulation to yield optimal per-sample Pareto fronts}
\label{sec:counter example strict convexity}
\begin{figure}
    \centering
    \begin{tabular}{ccc}
        & \textbf{Training per sample} & \textbf{Training on average losses} \\
    & $\max \tfrac{1}{|S|}\sum_{k=1}^{|S|} \mathrm{HV}\left(\mathfrak{L}(\Theta,s_{k})\right)$ & $\max \mathrm{HV}\left(\overline{\mathfrak{L}(\Theta,S)}\right)$  \\
    & (Dynamic~loss~\eqref{eq:final_joint_loss}) & (Dynamic~loss~\eqref{eq:joint_loss_max_hv_of_mean_losses})  \\
    \multirow[t]{1}{*}{
    \rotatebox[origin=c]{90}{\textbf{Strictly convex}}}
    &
\begin{subfigure}{0.30\textwidth}
    \centering
    \includegraphics[width=\textwidth]{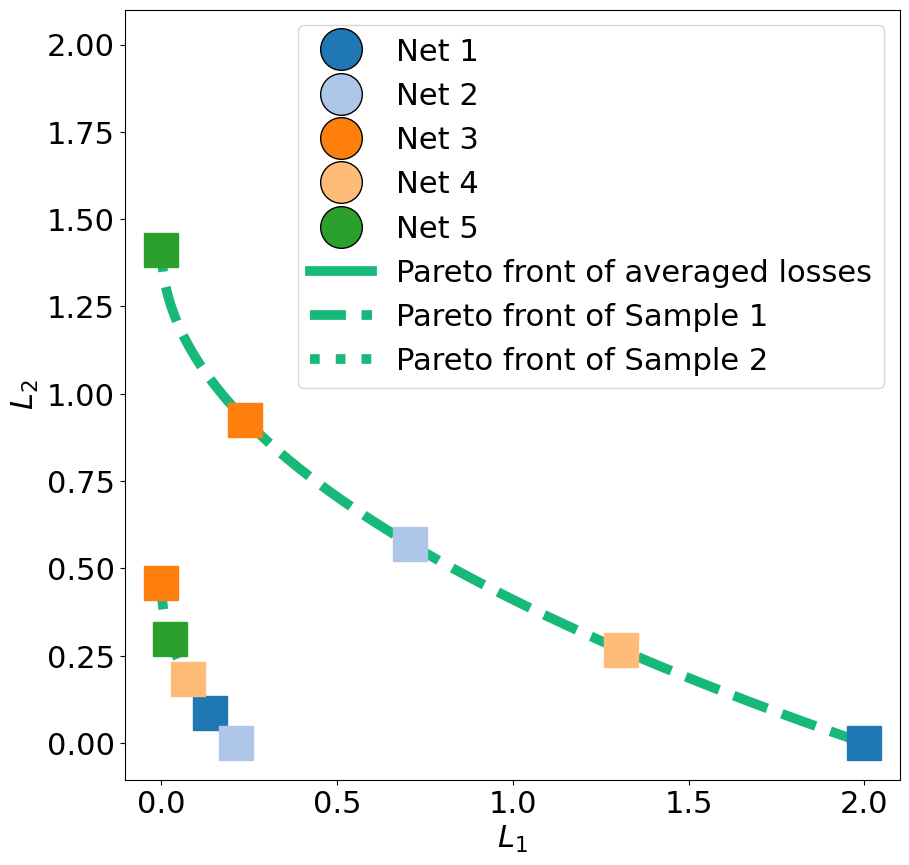}
    \caption{}
    \label{fig:opt_ex_loss_per_sample_convex_counter_example_higamo_hv}
\end{subfigure}
    &
\begin{subfigure}{0.30\textwidth}
    \centering
    \includegraphics[width=\textwidth]{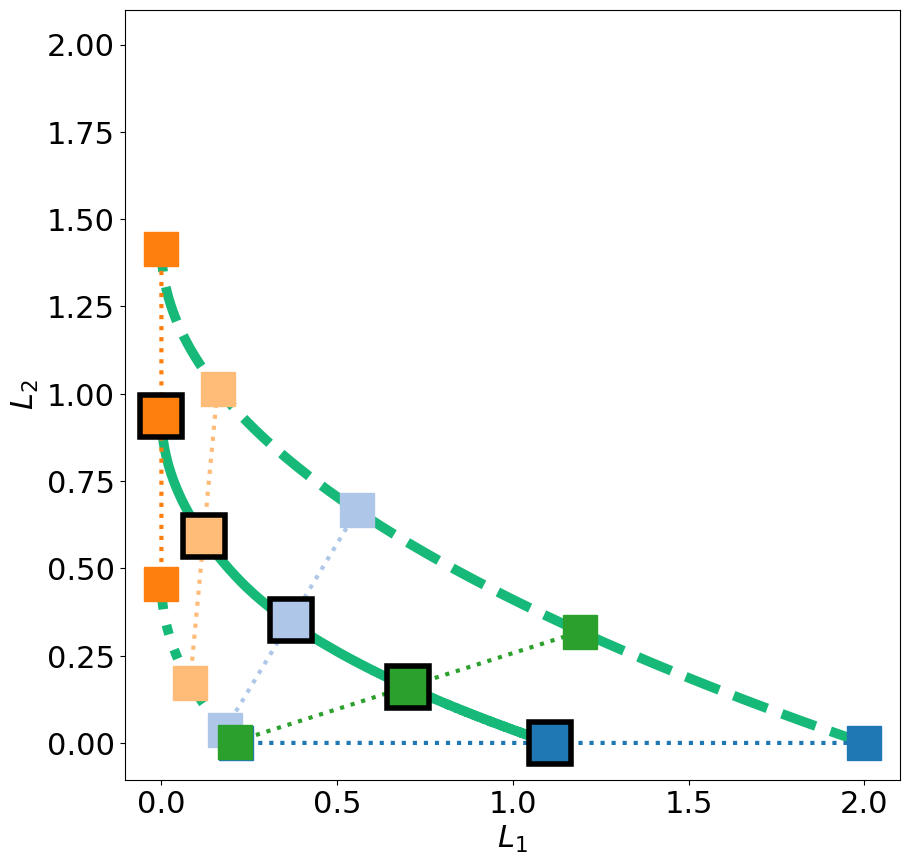}
    \caption{}
    \label{fig:opt_ex_mean_losses_convex_counter_example_higamo_hv}
\end{subfigure}
    \end{tabular}
    \caption{An example of a learning problem with strictly convex Pareto fronts in which HV maximization of average losses does not result in well-distributed outputs on both samples' fronts. HV maximization of each sample's losses (left) and of average losses (right).}
    \label{fig:optimization_example_strictly_convex_counter_example_higamo_hv}
\end{figure}

For HV maximization in the above strictly convex case (Figure~\ref{fig:opt_ex_mean_losses_convex_higamo}), the trade-offs chosen by HV maximization on the average front also appear optimal for HV maximization per sample. This observation is not true in general as illustrated in the following example. Figure~\ref{fig:optimization_example_strictly_convex_counter_example_higamo_hv} shows the result of training five neural networks for the strictly convex case described above but using the following loss functions:
\begin{equation*}
    L_{1} = \left\lVert z_{k} - x_{k}^{1} \right\rVert_{2}^{2}  \quad L_{2} = \left\lVert z_{k} - x_{k}^{2}\right\rVert_{2}^{1.01}
\end{equation*}
with $x_{1}^{1} = [0,0]$, $x_{1}^{2} = [1,1]$, $x_{2}^{1} = [0.05,0.4]$, $x_{2}^{2} = [0.5,0.5]$. Both samples' fronts are strictly convex but asymmetric in the 45$^{\circ}$-line $L_{1}=L_{2}$ and with differing curvature.
It becomes apparent that the Pareto front predictions of one sample are not well-distributed when using Dynamic loss \eqref{eq:joint_loss_max_hv_of_mean_losses} (Figure~\ref{fig:opt_ex_mean_losses_convex_counter_example_higamo_hv}). Dynamic loss \eqref{eq:final_joint_loss}, however, still generates well-distributed predictions per front (Figure~\ref{fig:opt_ex_loss_per_sample_convex_counter_example_higamo_hv}). This observation indicates that strict convexity of the samples' Pareto fronts alone does not guarantee that HV maximizing trade-offs on the front of average losses are also optimal for each sample. 

\subsubsection{Per-sample HV maximization does not guarantee the same output ordering across samples}
Apart from the increased computational burden, another potential disadvantage of using Dynamic loss~\eqref{eq:final_joint_loss} is that a given network does not guarantee the same ordering of trade-offs for different samples. For example, in the strictly convex case of Figure~\ref{fig:opt_ex_loss_per_sample_convex_higamo}, the colored squares on Sample 1's front are beige, light blue, green, blue, and orange from left to right, while Sample 2's front reads beige, orange, blue, green, and light blue. The ordering can only be known during inference if losses can be computed (so not in, e.g., regressions (Section~\ref{sec:sin_cos}), segmentations (Section~\ref{sec:mo_segmentation}), or classification tasks where labels are not known during inference). This condition would pose a limitation in scenarios where a decision-maker requires the ordering to make a decision. A prediction's position on the approximated Pareto front would remain unknown and a choice between predictions would have to be made without knowing what trade-off each prediction represents. Using Dynamic loss~\eqref{eq:joint_loss_max_hv_of_mean_losses} for training would permit to estimate the ordering and trade-offs from data where ground truth labels are available, e.g., training data. In some real-life scenarios, however, losses can be computed without labels (e.g., style transfer, Section~\ref{sec:style_transfer}) or decision-makers rely more on the predictions themselves, e.g., organ segmentations (Section~\ref{sec:mo_segmentation}) or generated images (Section~\ref{sec:style_transfer}), rather than their estimated ordering or trade-offs. Therefore, the practical disadvantage of unordered outputs for per-sample HV maximization is minor.

In conclusion, the above illustrative examples and arguments highlight why optimizing the trade-offs for each sample is preferable to average loss formulations and crucial to ensure approximation of a well-distributed Pareto front for each sample.

\subsection{Implementation}
We implemented the HV maximization of losses from multiple networks, as defined in Equation~\eqref{eq:final_joint_loss}, in Python\footnote{Code is available at \url{https://github.com/timodeist/multi_objective_learning}}. The neural networks were implemented using the PyTorch framework \citep{pytorch}. We use \cite{fonseca2006improved}'s HV computation reimplemented by Simon Wessing, available from \cite{higamo_code}. The HV gradients
$\ddtfrac{\mathrm{HV}\left(\mathfrak{L}(\Theta_{q(i)},s_{k}) \right)}{L_{j}(\theta_{i},s_{k})}$
 are computed following the algorithm by \cite{emmerich2014time}. Networks with identical losses are assigned the same HV gradients. For non-dominated networks with one or more identical losses (which can occur in training with three or more losses), the left- and right-sided limits of the HV function derivatives are not the same \citep{emmerich2014time} and they are set to zero. Non-dominated sorting is implemented based on \cite{deb2002fast}.
We experimentally tested our approach for two and three objectives, but the algorithms for HV and HV gradient computations also extend to more objectives.

\subsection{Time Complexity}
The published time complexities of different steps in calculating HV maximizing gradients for $n$ losses and $p$ solutions are as follows: $\mathcal{O}(np^{2})$ for non-dominated sorting \citep{deb2002fast}, $\mathcal{O}(p^{(n-2)}\log{p}))$ for HV computation of $p$ non-dominated solutions if $n>2$, $\mathcal{O}(p)$ for HV calculation for $n=2$ after sorting in one loss \citep{fonseca2006improved}, $\mathcal{O}(p\log(p))$ for calculating HV gradients 
$\dd{\mathrm{HV}\left(\mathfrak{L}(\Theta_{q(i)},s_{k}) \right)}{L_{j}(\theta_{i},s_{k})}$
 for two and three losses, and $\mathcal{O}(p^{2})$ for HV gradient calculation of four losses \citep{emmerich2014time}. Note that the latter two complexities assume specialized non-dominated sorting and HV computation subroutines that we did not implement. Overall, for moderate $p$ values and $n\leq4$, this means only little additional computational load compared to computing loss gradients for neural network training, which gives an HV maximization-based approach an edge over other competitive approaches in this direction.
\section{Experiments}
\label{sec:experiments}
We now present the application of our method (Dynamic~loss~\eqref{eq:final_joint_loss}) in experiments with different MO problems: a simple MO regression example, a multi-observer medical image segmentation, and a neural style transfer optimization problem.  

We compared the performance of our approach with \textbf{linear scalarization} of average losses and two state-of-the-art approaches called \textbf{Pareto MTL} \citep{lin2019paretomtl}, and \textbf{EPO} \citep{mahapatra2020multi}. Both Pareto MTL and EPO try to find Pareto optimal solutions on the Pareto front of average losses for a given trade-off vector using dynamic loss functions. For a consistent comparison, we used the trade-offs used in the original experiments of EPO for Pareto MTL, EPO, and as fixed weights in linear scalarization.

Experiments were run on systems using Intel(R) Xeon(R) Silver 4110 CPU @ 2.10GHz with NVIDIA GeForce RTX 2080Ti, or Intel(R) Core(R) i5-3570K @ 3.40Ghz with NVIDIA GeForce GTX 1060 6GB. The training was done using the Adam optimizer \citep{kingma2014adam}. The learning rate and $\beta_{1}$ of Adam were tuned for each approach based on the maximal HV of validation loss vectors in the last iteration. Details of the hyperparameter tuning experiments are provided in Appendix~\ref{suppl:tuning}.

\subsection{MO Regression}
\label{sec:sin_cos}
To illustrate our proposed approach, we begin with an artificial MO learning example. Consider two conflicting objectives: given a sample $x_k$ from input variable $X\in [0,2\pi]$, predict the corresponding output $z_{k}$ that matches $y_{k}^{(j)}$ from target variable $Y_{j}$, where $X$ and $Y_{j}$ are related as follows:
\begin{equation*}
  Y_{1}=\cos(X),\quad Y_{2}=\sin(X)
\end{equation*}
The corresponding loss functions are $L_{j}=\mathrm{MSE}_{j}=\tfrac{1}{|S|}\sum_{k=1}^{|S|}(y_{k}^{(j)}-z_{k})^{2}$. We generated 200 samples of input and target variables for training and validation each. We trained five neural networks for 20000 iterations each with two fully connected linear layers of 100 neurons followed by ReLU nonlinearities. The reference point was set to $(20,20)$.

\begin{figure*}[ht!]
\begin{subfigure}{0.32\textwidth}
    \centering
    \includegraphics[width=0.80\textwidth]{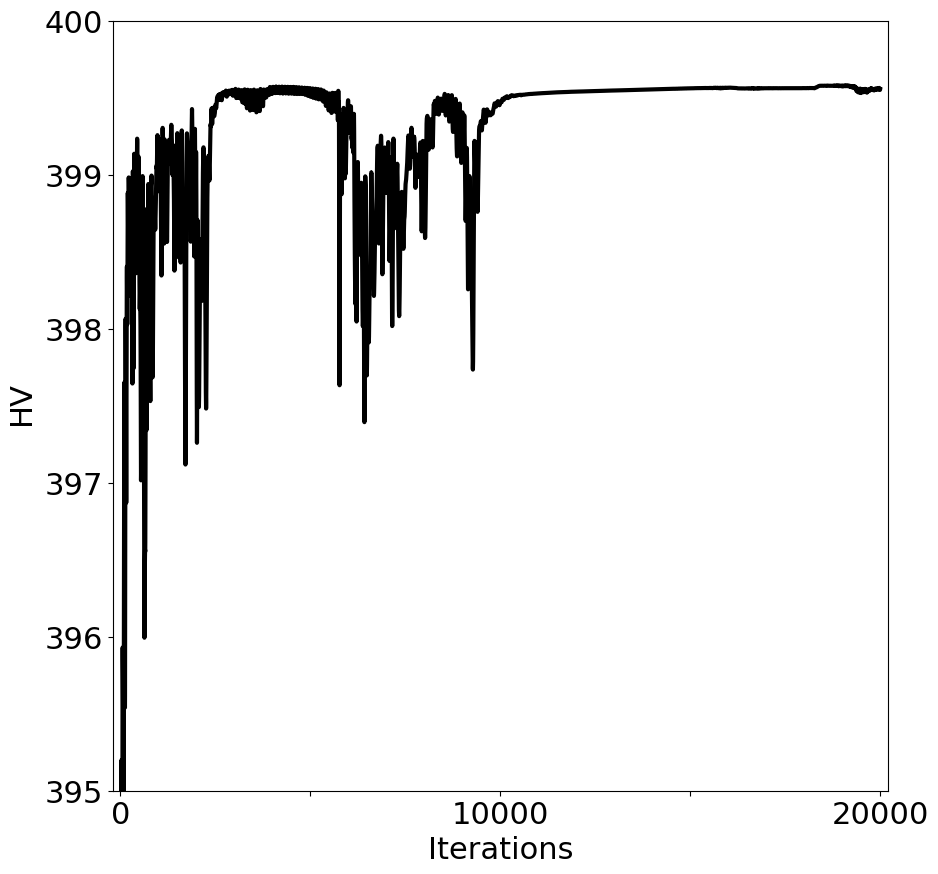}
    \caption{}
    \label{fig:sincos_2d_losses}
\end{subfigure}
\begin{subfigure}{0.32\textwidth}
    \centering
    \includegraphics[width=\textwidth]{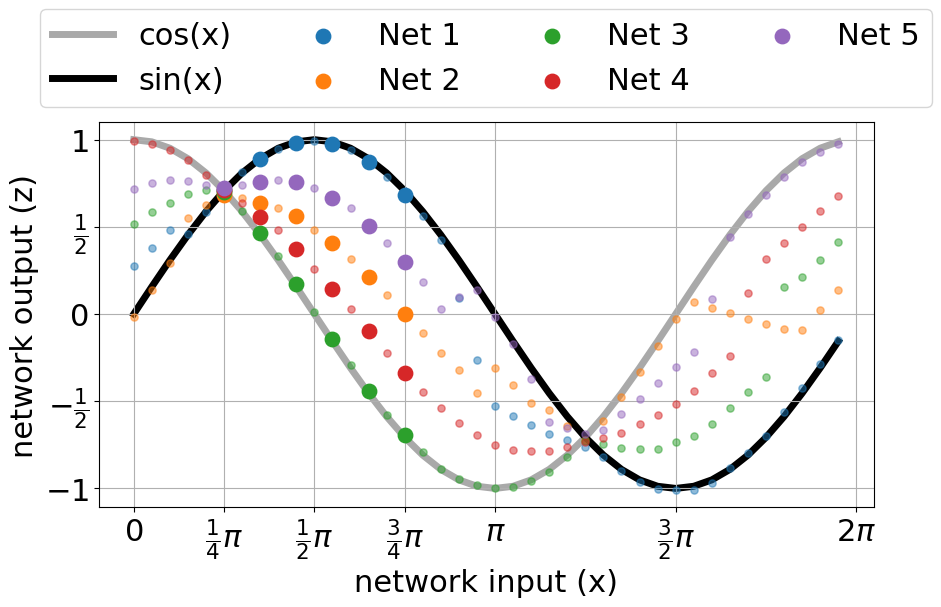}
    \caption{}
    \label{fig:sincos_2d_predictions}
\end{subfigure}
\begin{subfigure}{0.32\textwidth}
    \centering
    \includegraphics[width=0.80\textwidth]{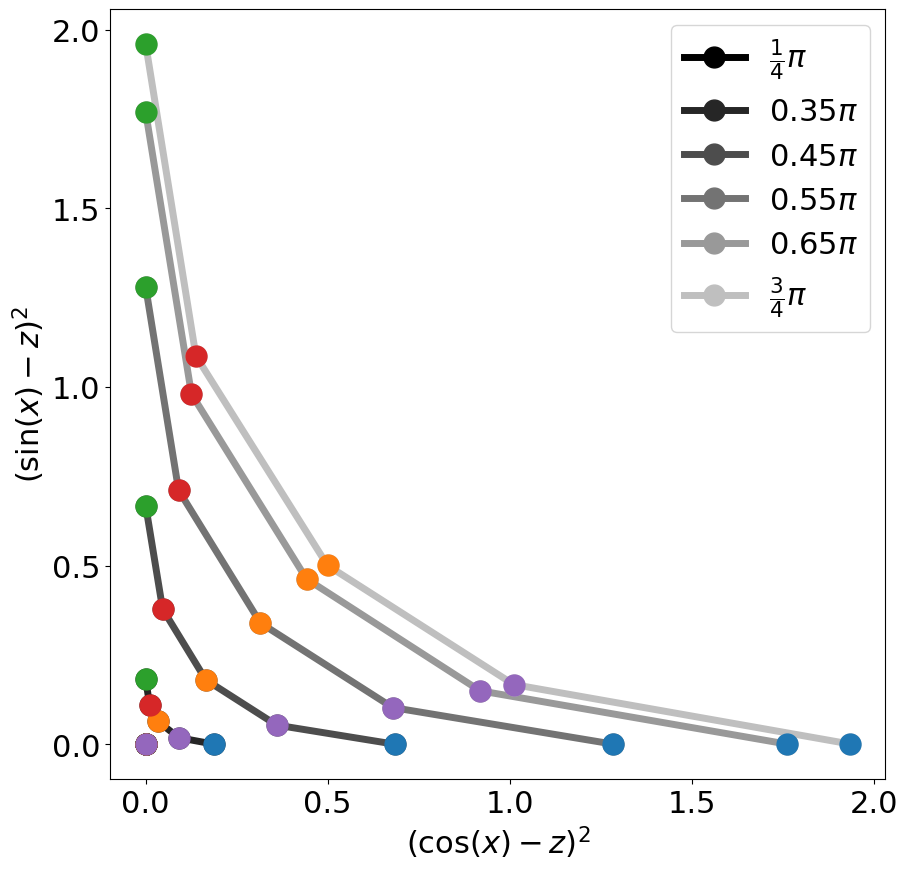}
    \caption{}
    \label{fig:sincos_2d_os}
\end{subfigure}
    \centering
    \caption{MO regression on two losses. (a) HV values for a set of networks over training iterations. (b) Network outputs for $X\in[0,2\pi]$. (c) Generated Pareto front predictions for a selection of six samples from $[\tfrac{1}{4}\pi,\tfrac{3}{4}\pi]$ in loss space.}
    \label{fig:sincos_2d}
\end{figure*}

Figure~\ref{fig:sincos_2d_losses} shows the HV over training iterations for the set of networks which stabilizes visibly. Figure~\ref{fig:sincos_2d_predictions} shows predictions (y-axis) for validation samples evenly sampled from $[0,2\pi]$ (x-axis). These predictions by five neural networks constitute Pareto front approximations for each sampled $x_k$, and correspond to precise predictions for $\cos(X)$ and $\sin(X)$, and trade-offs between both target functions. As discussed in Section~\ref{sec:batch_vs_persample}, a network may generate predictions with changing trade-offs for different samples. This is clearly demonstrated by Networks~2-5 in Figure~\ref{fig:sincos_2d_predictions} for $x\in [\tfrac{3/2}\pi,2\pi]$ which change their trade-offs.

Figure~\ref{fig:sincos_2d_os} shows these Pareto front predictions in loss space (only a selection of outputs is shown to simplify visualization). It becomes clear from Figures~\ref{fig:sincos_2d_predictions}~\&~\ref{fig:sincos_2d_os} that each $x_k$ has a differently sized Pareto front which the networks are able to predict. The Pareto fronts for samples corresponding to $x =\tfrac{1}{4}\pi$ (and $x=\tfrac{5}{4}\pi$) reduce to a single point because $\cos(X)$ and $\sin(X)$ are equal. An illustration of MO learning on three losses is provided in Appendix~\ref{suppl:sin_cos}.

\begin{figure}[h]
\begin{tabular}{ccccc}
    & \textbf{\footnotesize Linear scalarization} & \textbf{\footnotesize Pareto MTL} & \textbf{\footnotesize EPO} & \textbf{\footnotesize HV maximization}\\
    \midrule
    \rotatebox[origin=c]{90}{\textbf{\footnotesize MSE \& MSE}} &
    \begin{subfigure}{0.2\textwidth}
    \centering
    \includegraphics[width=\textwidth]{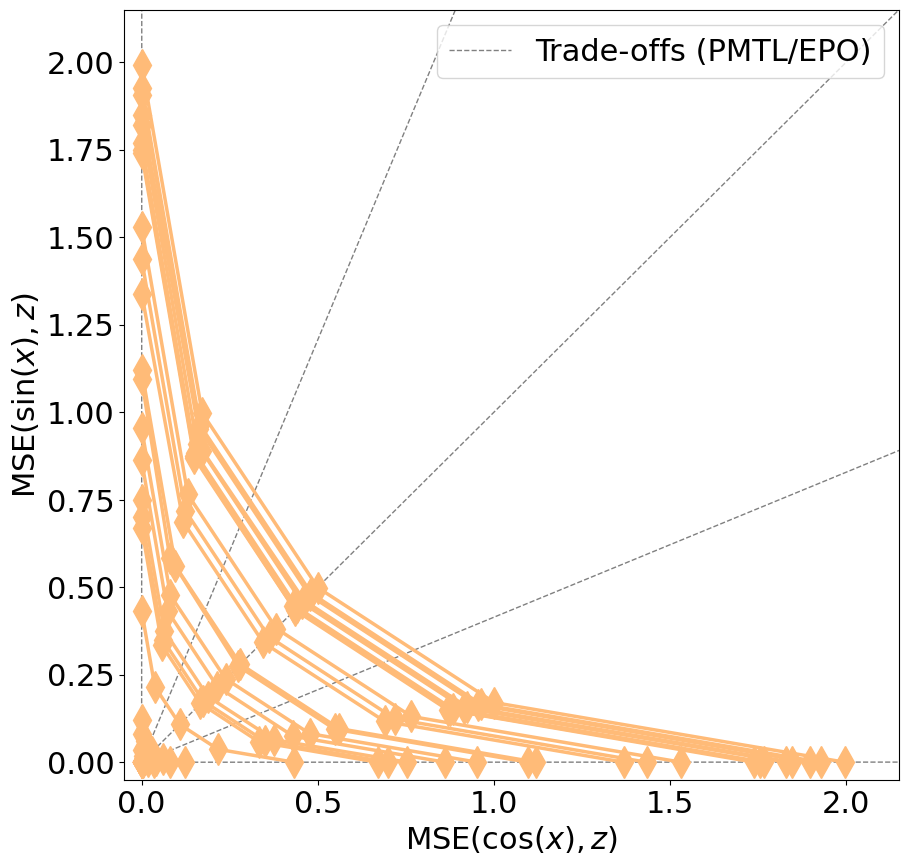}
    \caption{}
    \label{fig:comparison_ls_sin_cos}
    \end{subfigure}
    & 
    \begin{subfigure}{0.2\textwidth}
    \centering
    \includegraphics[width=\textwidth]{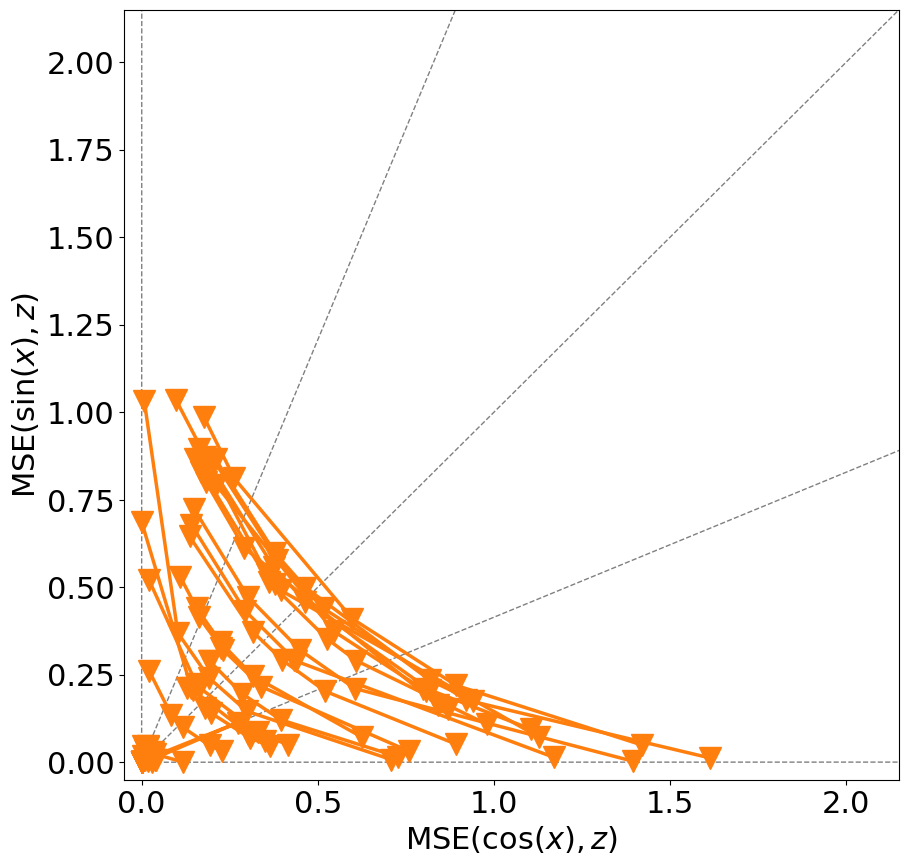}
    \caption{}
    \label{fig:comparison_pmtl_sin_cos}
    \end{subfigure}
    &
    \begin{subfigure}{0.2\textwidth}
    \centering
    \includegraphics[width=\textwidth]{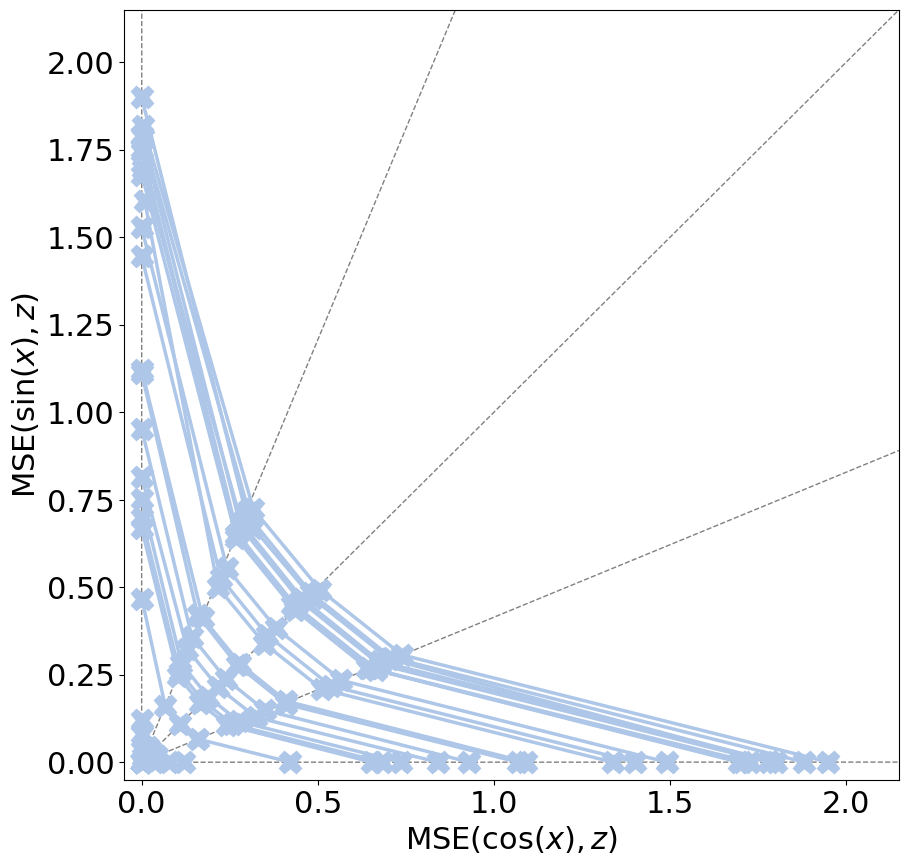}
    \caption{}
    \label{fig:comparison_epo_sin_cos}
    \end{subfigure}
    &
    \begin{subfigure}{0.2\textwidth}
    \centering
    \includegraphics[width=\textwidth]{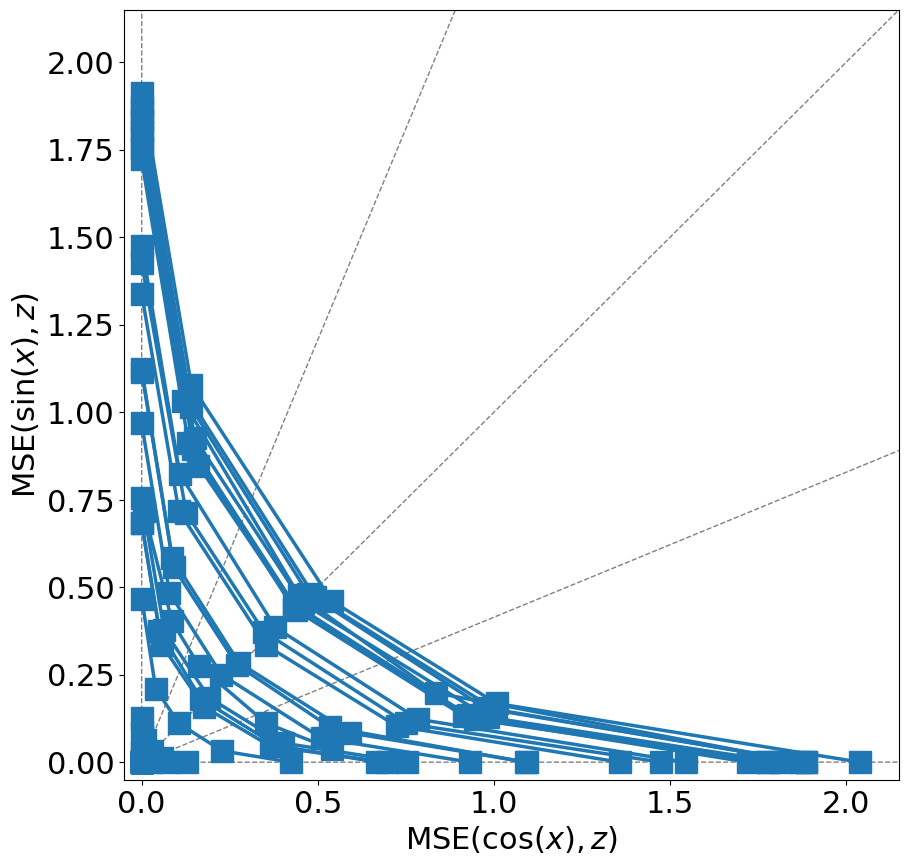}
    \caption{}
    \label{fig:comparison_higamo_hv_sin_cos}
    \end{subfigure}\\
    
    \rotatebox[origin=c]{90}{\textbf{\footnotesize MSE \& L1-Norm}}
    & 
    \begin{subfigure}{0.2\textwidth}
    \centering
    \includegraphics[width=\textwidth]{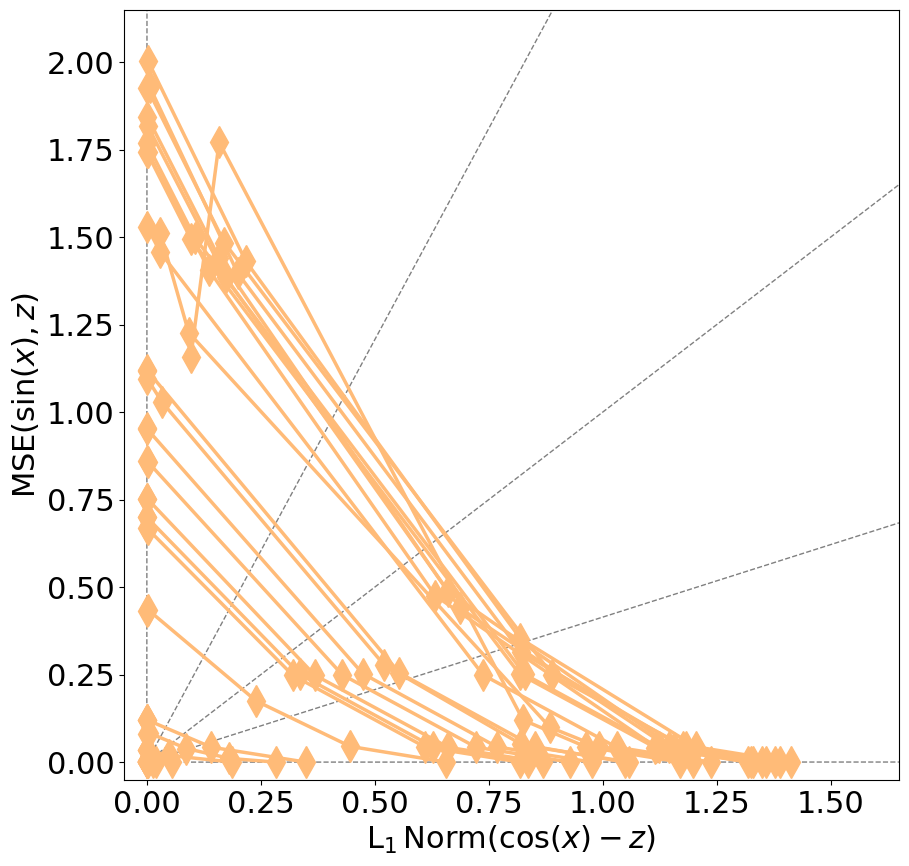}
    \caption{}
    \label{fig:comparison_ls_sin_cos_l1}
    \end{subfigure}
    &
    \begin{subfigure}{0.2\textwidth}
    \centering
    \includegraphics[width=\textwidth]{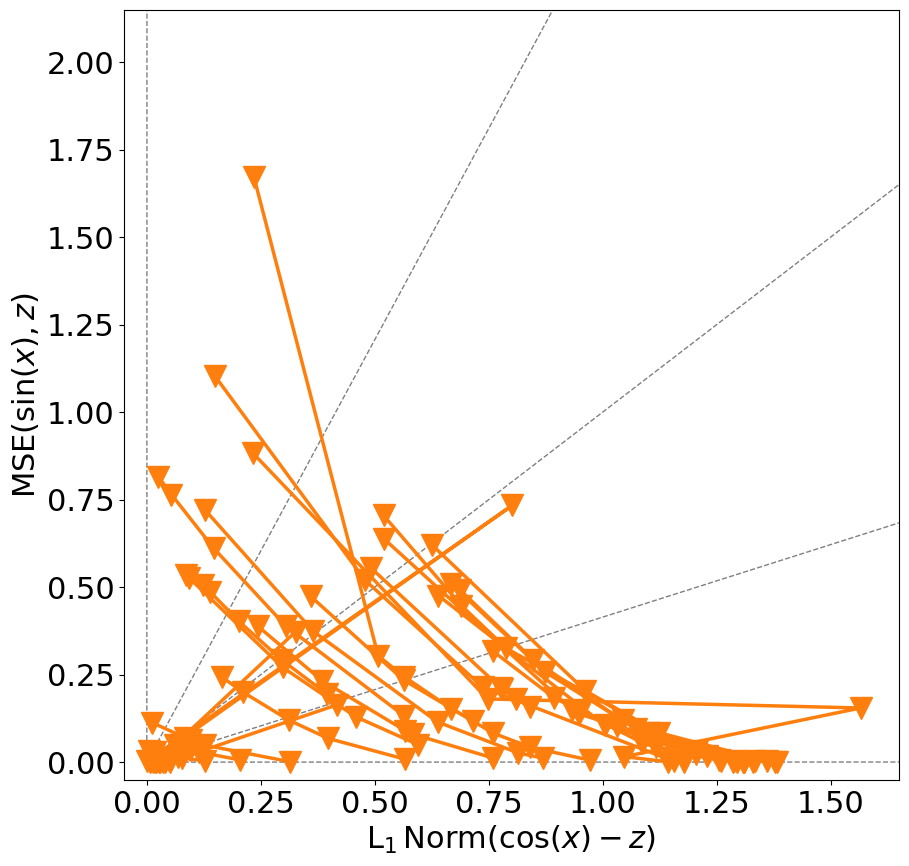}
    \caption{}
    \label{fig:comparison_pmtl_sin_cos_l1}
    \end{subfigure}
    &
    \begin{subfigure}{0.2\textwidth}
    \centering
    \includegraphics[width=\textwidth]{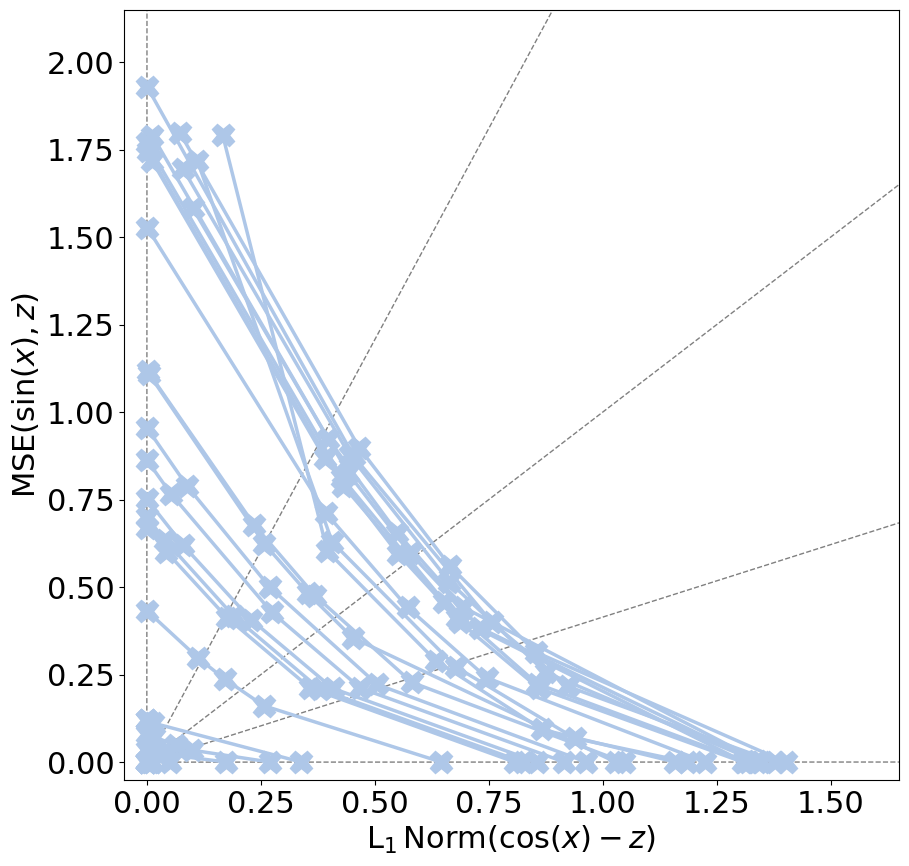}
    \caption{}
    \label{fig:comparison_epo_sin_cos_l1}
    \end{subfigure}
    &
    \begin{subfigure}{0.2\textwidth}
    \centering
    \includegraphics[width=\textwidth]{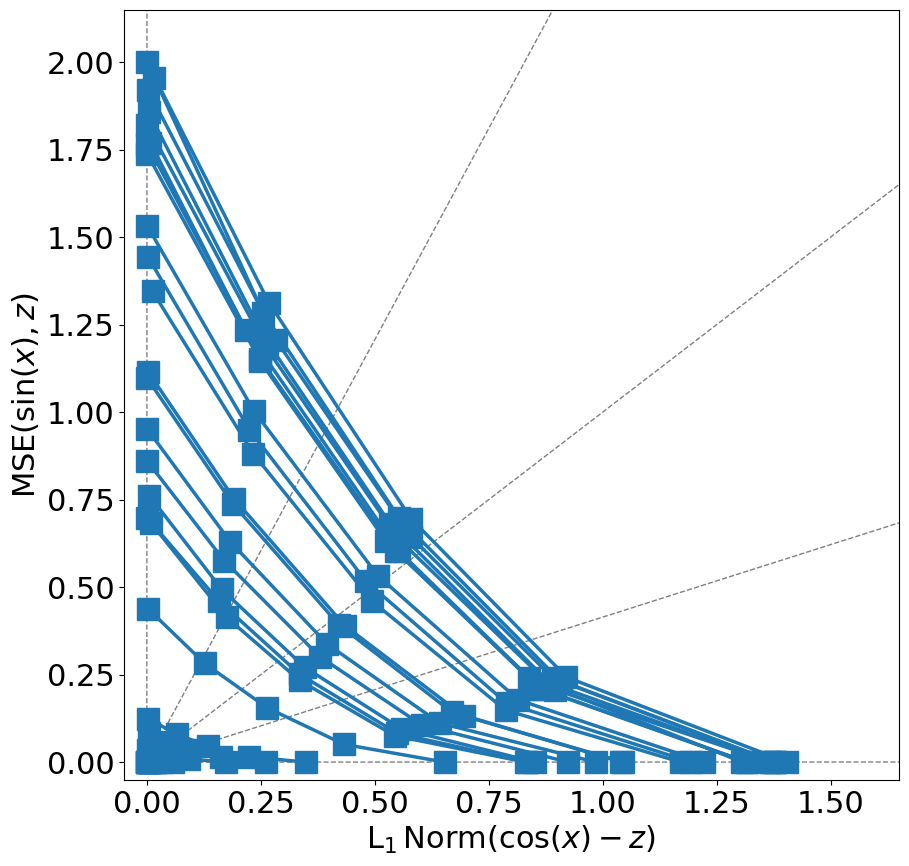}
    \caption{}
    \label{fig:comparison_higamo_hv_sin_cos_l1}
    \end{subfigure}\\
    
    \rotatebox[origin=c]{90}{\textbf{\footnotesize MSE \& scaled MSE}}
    &
    \begin{subfigure}{0.2\textwidth}
    \centering
    \includegraphics[width=\textwidth]{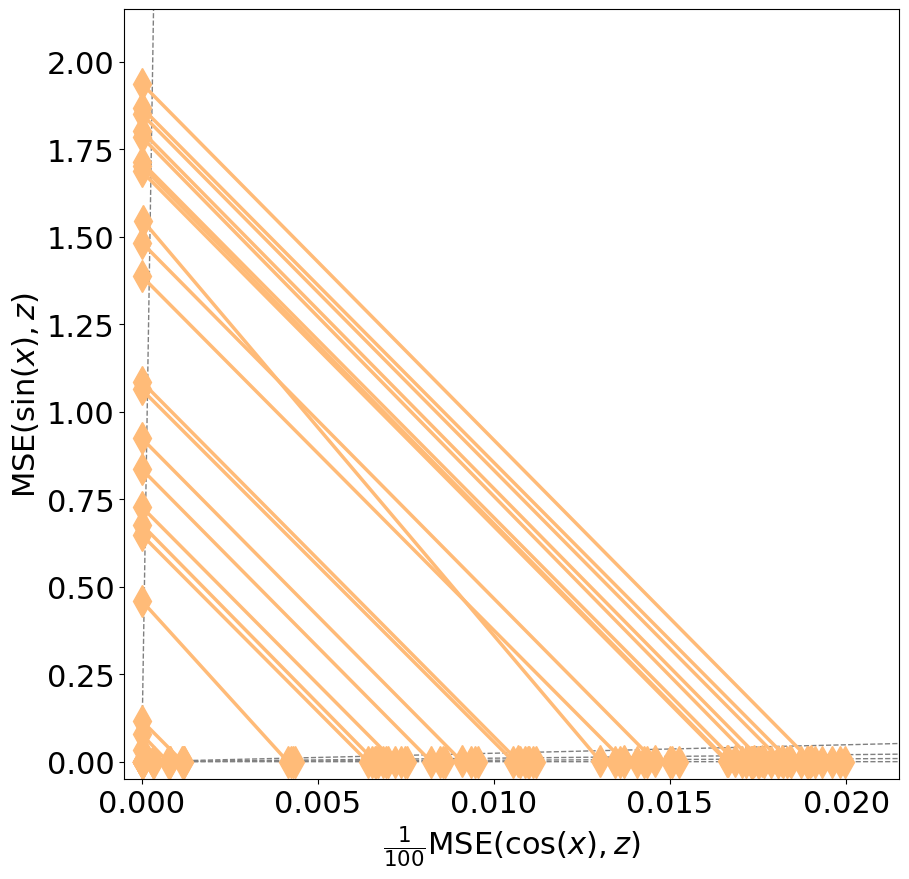}
    \caption{}
    \label{fig:comparison_ls_sin_cos_scaledmse}
    \end{subfigure}
     &
    \begin{subfigure}{0.2\textwidth}
    \centering
    \includegraphics[width=\textwidth]{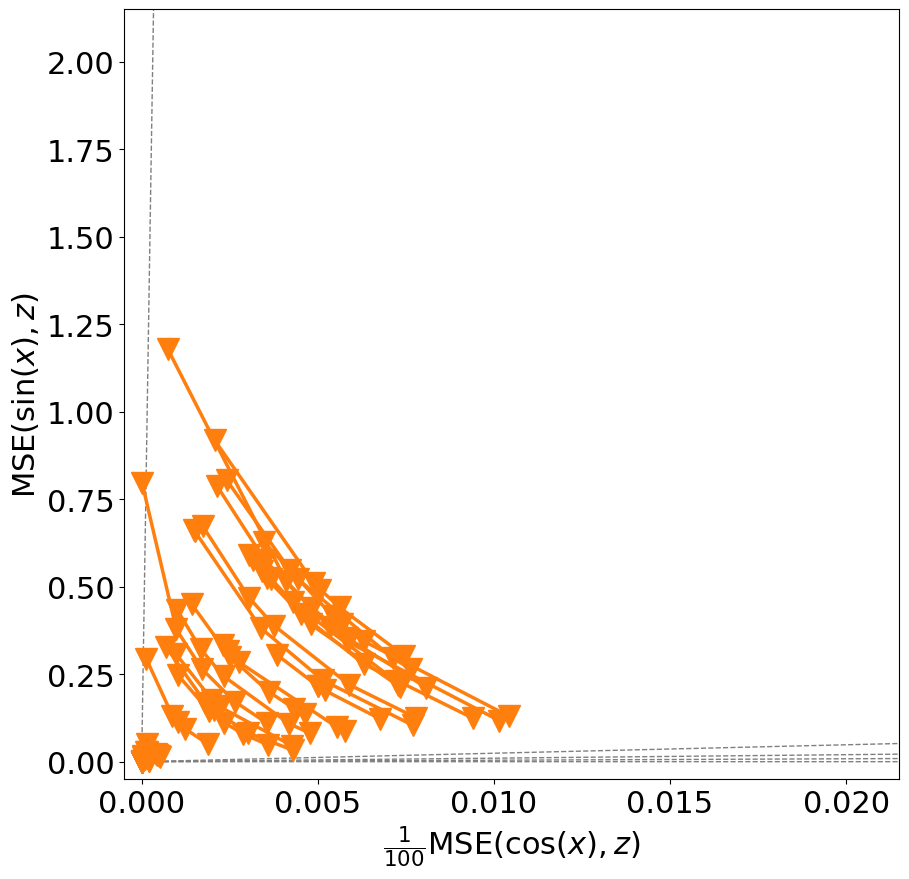}
    \caption{}
    \label{fig:comparison_pmtl_sin_cos_scaledmse}
    \end{subfigure}
    & 
    \begin{subfigure}{0.2\textwidth}
    \centering
    \includegraphics[width=\textwidth]{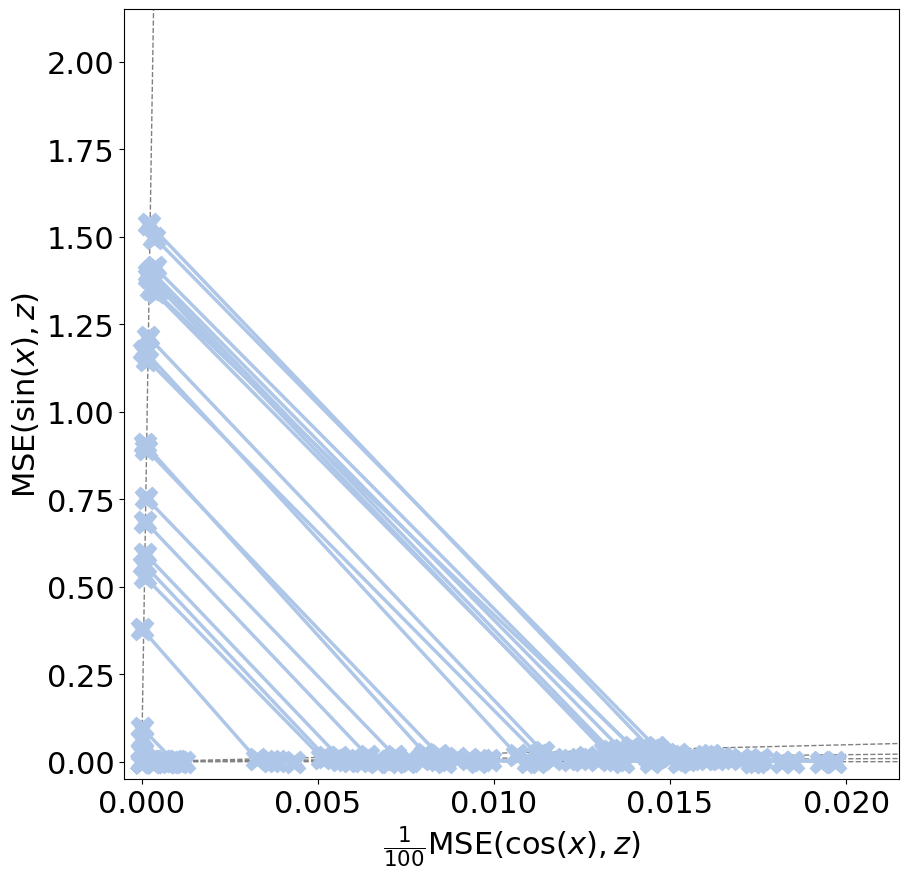}
    \caption{}
    \label{fig:comparison_epo_sin_cos_scaledmse}
    \end{subfigure}
    & 
    \begin{subfigure}{0.2\textwidth}
    \centering
    \includegraphics[width=\textwidth]{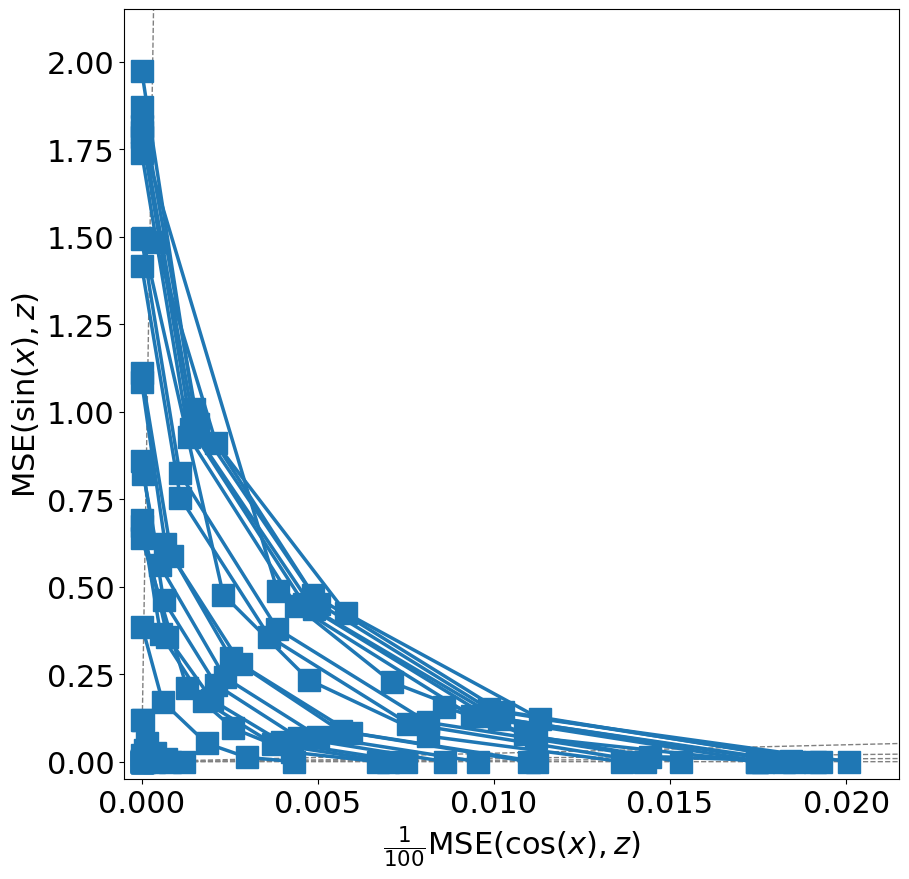}
    \caption{}
    \label{fig:comparison_higamo_hv_sin_cos_scaledmse}
    \end{subfigure}
\end{tabular}
    \centering
    \caption{Pareto front approximations by sets of five neural networks on a random subset of validation samples trained using four approaches. Three different pairs of loss functions are used: (a)-(d) MSE and MSE, (e)-(h) MSE and L1-Norm, and (i)-(l) MSE and scaled MSE.}
    \label{fig:comparison_sin_cos_all}
\end{figure}

\begin{table}[]
\renewcommand{\arraystretch}{1.5}
\caption{\small The mean HV of the approximated Pareto fronts for 200 validation samples for three pairs of loss functions (columnwise): MSE \& MSE, MSE \& L1-Norm, and MSE \& scaled MSE. Median (inter-quartile range) values of the mean HV for 25 runs are reported. The maximal median HV in each column is \textbf{highlighted}. Statistical significance of one-sided Wilcoxon signed rank tests after correction for multiple comparison is indicated for comparisons: LS vs HV maximization (\textsuperscript{$\ast$}), PMTL vs HV maximization (\textsuperscript{$\dagger$}), and EPO vs HV maximization (\textsuperscript{$\ddagger$}). The relevance of small increases in HV close to the maximum (400) is described in Section~\ref{sec:sin_cos_asymmetric}.
}
\label{tab:HV_sin_cos}
\begin{tabular}{m{2cm}m{3.8cm}m{3.8cm}m{3.8cm}}
\toprule
\textbf{} & \textbf{MSE \& MSE} & \textbf{MSE \& L1-Norm} & \textbf{MSE \& scaled MSE}\\
\midrule
\small Linear scalarization (LS) & \footnotesize \textbf{399.5929\textsuperscript{$\ast$} \newline(399.5776 -- 399.6018)}  & \footnotesize 399.2909 \newline(399.2738 -- 399.3045)  & \footnotesize 399.9859 \newline(399.9857 -- 399.9864)\\ 

\small Pareto MTL (PMTL) & \footnotesize 397.1356 \newline(396.3212 -- 397.6288)  & \footnotesize 392.2956 \newline(392.0377 -- 393.4942)  & \footnotesize 398.3159 \newline(397.4799 -- 398.6699)\\ 

\small EPO & \footnotesize 399.5135 \newline(399.5051 -- 399.5348) & \footnotesize 399.0884 \newline(398.998 -- 399.1743)  & \footnotesize 399.9885 \newline(399.9883 -- 399.9889)\\ 

\small HV maximization & \footnotesize 399.5823\textsuperscript{$\dagger \ddagger$} \newline(399.5619 -- 399.6005) & \footnotesize \textbf{399.3795\textsuperscript{$\ast \dagger \ddagger$} \newline(399.3481 -- 399.4039)}  & \footnotesize \textbf{399.9954\textsuperscript{$\ast \dagger \ddagger$} \newline(399.9927 -- 399.9957)}\\ 
\bottomrule
\end{tabular}
\end{table}

\subsubsection{Asymmetric Pareto Fronts}
\label{sec:sin_cos_asymmetric}
Conflicting loss functions may behave or scale differently, giving rise to asymmetric Pareto fronts (asymmetric in the line where losses are equal, i.e., $L_{1}=L_{2}$). We investigated the ability of different MO learning approaches to predict different types of Pareto fronts using combinations of basic loss functions: the symmetric case with two MSE losses as in Figure \ref{fig:sincos_2d}, and two asymmetric cases each with MSE as one loss and L1-norm or MSE scaled by $\tfrac{1}{100}$ as the second loss. Figure~\ref{fig:comparison_sin_cos_all} shows Pareto front approximations for all three cases. The mean HV over 200 validation samples is computed for all approaches and Table~\ref{tab:HV_sin_cos} displays the median and inter-quartile ranges (IQR) over 25 runs. 
The magnitude of the HV is largely determined by the position of the reference point. For $r=(20,20)$ the maximal HV equals 400 minus the area bounded by the utopian point $(0,0)$ and a sample's Pareto front. Even poor approximations of a sample's Pareto front can yield a HV $\geq390$. For these reasons, HVs in Table~\ref{tab:HV_sin_cos} appear large and minuscule differences between HVs are relevant.

Figures~\ref{fig:comparison_ls_sin_cos}~\&~\ref{fig:comparison_epo_sin_cos} show that fixed linear scalarizations and EPO produce networks generating well-distributed outputs with low losses that predict a sample's symmetric Pareto front for two conflicting MSE losses. As these a priori selected trade-offs appear to span this Pareto front shape well, linear scalarization's training based on fixed loss weights is more efficient than training on a dynamic loss as used by HV maximization (Figure~\ref{fig:comparison_higamo_hv_sin_cos}) which is affected by stochastic effects. This increased efficiency of training using fixed weights that are suitable for symmetric MSE losses presumably results in a slightly higher HV (Table~\ref{tab:HV_sin_cos}) for linear scalarization. 
The positions on the front approximated by linear scalarization seem to be far from the pre-specified trade-offs (gray lines). This is expected because, by definition of linear scalarization, the solutions should lie on the approximated Pareto front where the tangent is perpendicular to the search direction specified by the trade-offs. For Pareto MTL, networks are clustered closer towards the center of the approximated Pareto front. Figure~\ref{fig:comparison_sin_cos_all} indicates that methods relying on pre-specified trade-offs can be sufficient when Pareto fronts are symmetric, which is also supported by an MO segmentation experiment on medical imaging data (Section~\ref{sec:mo_segmentation}).

Optimizing MSE and L1-Norm (Figures~\ref{fig:comparison_ls_sin_cos_l1}-\ref{fig:comparison_higamo_hv_sin_cos_l1}) results in an asymmetric Pareto front approximation. The predictions by our HV maximization-based approach remain well distributed across the fronts. EPO also still provides a decent spread albeit less uniform across samples whereas linear scalarization and Pareto MTL tend to both or mostly the lower extrema, respectively.

The difficulty of manually pre-specifying the trade-offs without knowledge of the Pareto front becomes more evident when optimizing losses with highly different scales (Figures~\ref{fig:comparison_ls_sin_cos_scaledmse}-\ref{fig:comparison_higamo_hv_sin_cos_scaledmse}). The pre-specified trade-offs do not evenly cover the Pareto fronts and consequently the networks trained by EPO do not cover the Pareto front evenly despite following the pre-specified trade-offs. Further, the networks optimized by Pareto MTL cover only the upper part of the fronts. Networks trained with fixed linear scalarizations tend towards both extrema. Our approach, on the other hand, trains networks that follow well-distributed trade-offs on the Pareto front.One might suggest that losses that exhibit different scales as in Figures~\ref{fig:comparison_ls_sin_cos_scaledmse}-\ref{fig:comparison_higamo_hv_sin_cos_scaledmse} can be `fixed' by manual rescaling. Results presented in Appendix~\ref{suppl:loss_rescale} show that rescaling losses based on scale differences observed at initialization does not sufficiently improve methods based on pre-specified trade-offs (Pareto MTL, EPO) or fixed linear scalarizations.

These experiments clearly demonstrate the advantage of MO learning by HV maximization over existing strategies when the Pareto front is asymmetric. 

\begin{figure}[h!]
    \centering
    \includegraphics[width=0.7\textwidth]{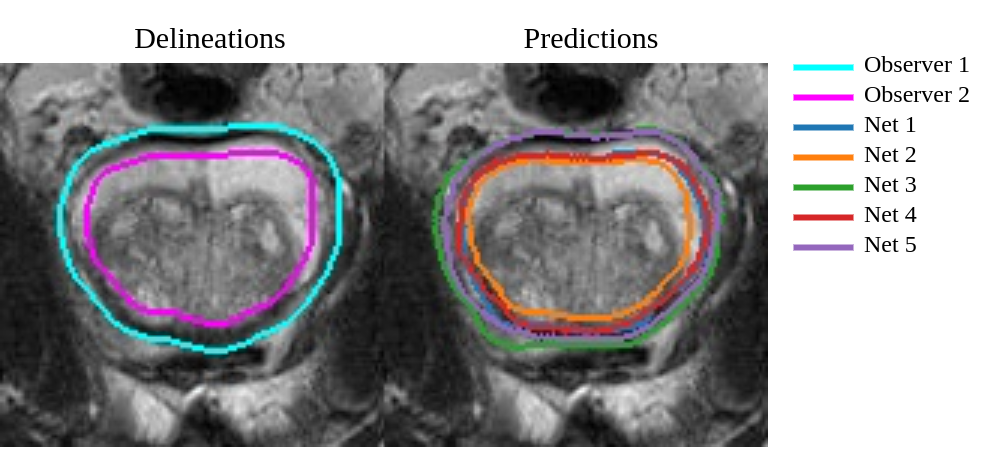}
    \caption{An example of one approximated Pareto set for Multi-observer segmentation consisting of predicted segmentation contours from five neural networks trained with HV maximization.}
    \label{fig:mo seg predictions}
\end{figure}

\subsection{Multi-Observer Medical Image Segmentation}
\label{sec:mo_segmentation}
Multi-observer medical image segmentation pertains to learning automatic segmentation based on delineations provided by multiple expert observers, which may be conflicting due to inter-observer variability \cite{villeirs2005interobserver, white2009inter}. The key motivation behind learning automatic medical image segmentation multi-objectively rather than learning from average of the two conflicting expert delineations, is to be able to present multiple automatically segmented contours covering the entire range of conflict between expert delineations to the clinicians. It is expected that this will increase the chance of one of the automatic contours being accepted without modification, and therefore would increase the clinical adaptability of this automatic segmentation approach. 

We applied our MO learning approach to the multi-observer medical image segmentation scenario mentioned in \cite{ArkadiySPIE2020}. The dataset \citep{simpson2019large} contains Magnetic Resonance Imaging (MRI) scans of prostate regions of 32 patients. The original single observer delineations are systematically perturbed to simulate different styles of delineation. We generate a bi-observer learning scenario from this dataset (Figure~\ref{fig:mo seg predictions}), where the two observer delineations disagree in the extent of the prostate region. 
We trained five neural networks for 10000 iterations to minimize soft Dice losses with the delineations provided by the two observers. The famous UNet \citep{UnetMiccai} architecture was used for the neural networks. The reference point was set to $(20, 20)$.
\begin{figure*}[h!]
\begin{subfigure}{0.4\textwidth}
    \centering
    \includegraphics[width=\textwidth]{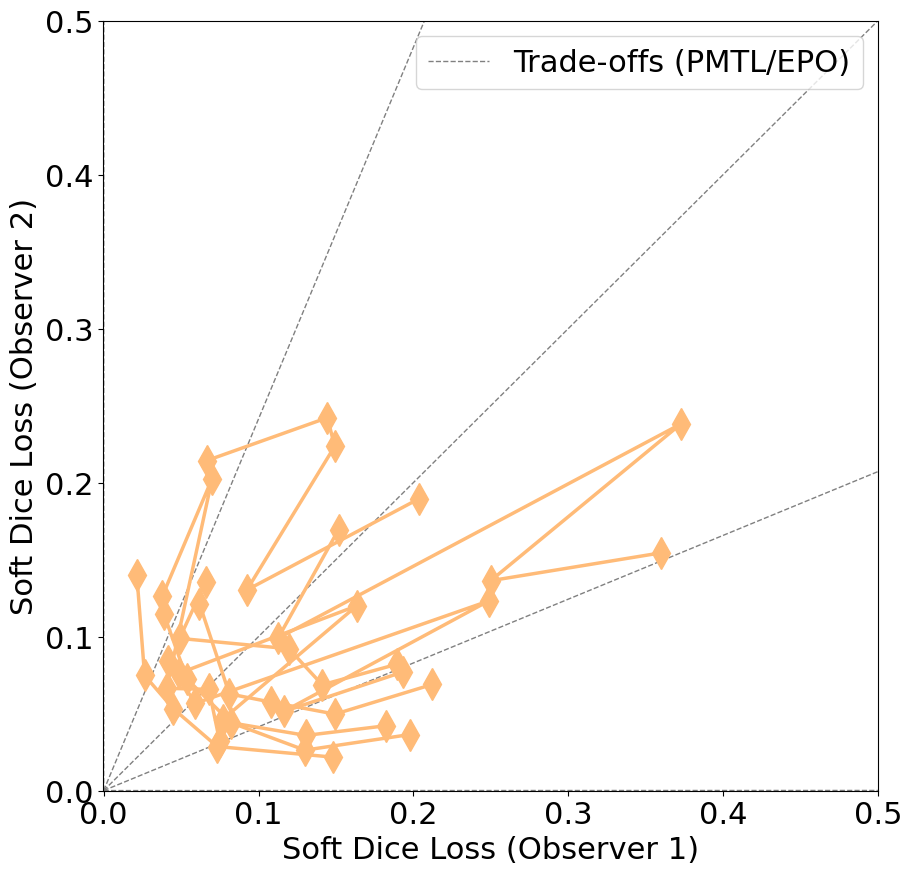}
    \caption{Linear scalarization}
    \label{fig:mo_seg_lin_scal}
\end{subfigure}
\begin{subfigure}{0.4\textwidth}
    \centering
    \includegraphics[width=\textwidth]{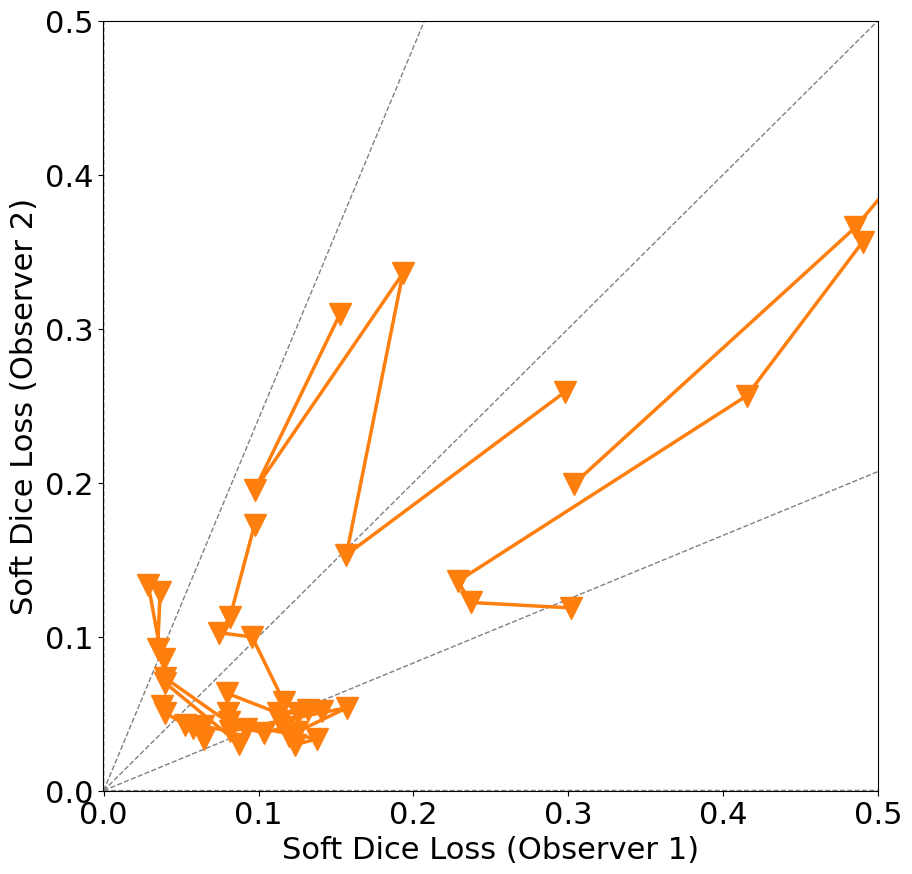}
    \caption{Pareto MTL}
    \label{fig:mo_seg_pareto_mtl}
\end{subfigure}
\begin{subfigure}{0.4\textwidth}
    \centering
    \includegraphics[width=\textwidth]{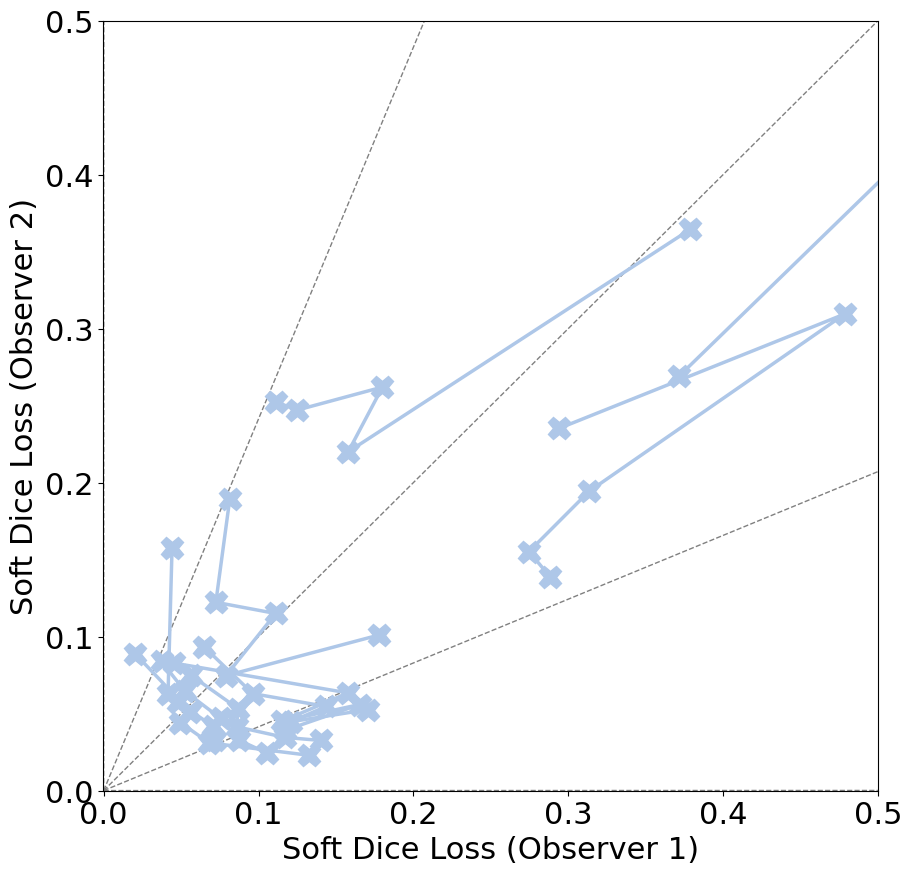}
    \caption{EPO}
    \label{fig:mo_seg_epo}
\end{subfigure}
\begin{subfigure}{0.4\textwidth}
    \centering
    \includegraphics[width=\textwidth]{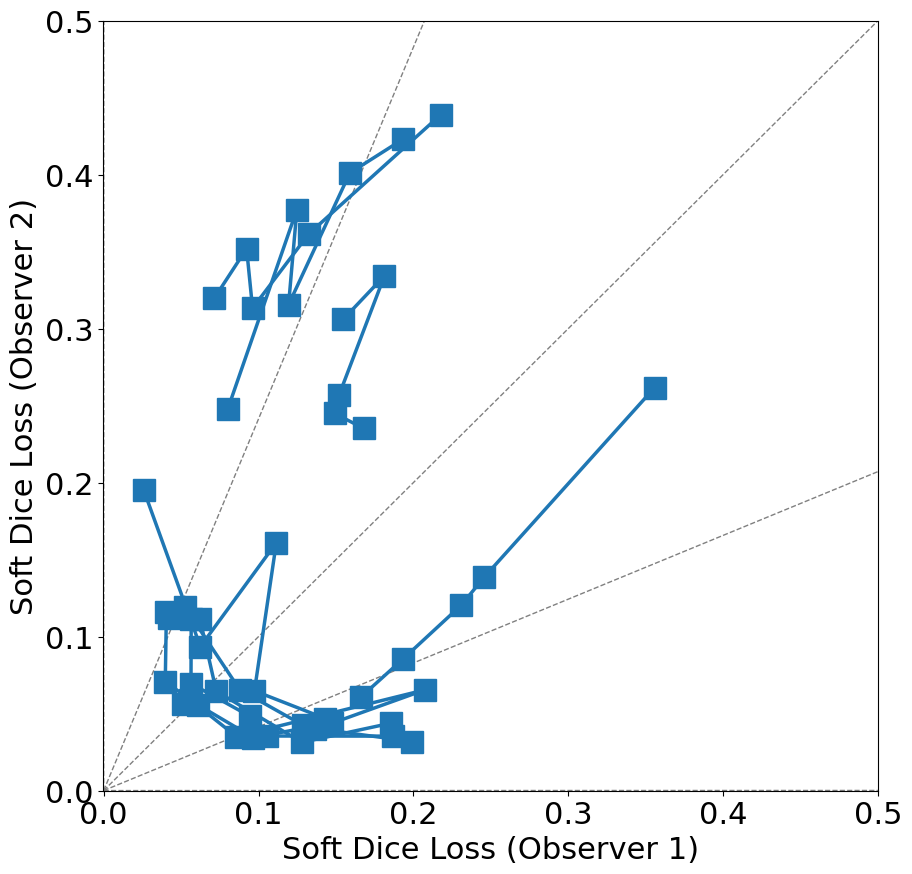}
    \caption{HV maximization}
    \label{fig:mo_seg_higamo_hv}
\end{subfigure}
    \centering
    \caption{Pareto front approximations for 10 randomly selected validation samples for multi-observer medical image segmentation by different approaches.}
    \label{fig:mo seg results}
\end{figure*}

Figure~\ref{fig:mo seg predictions} shows one example of multi-objective decision-making in our multi-observer medical image segmentation scenario. The delineations from Observer 2 consistently have a prostate region that is under-segmented by 10 pixels as compared to Observer 1. The predictions from two out of five neural networks follow one delineation style each, the rest of the predictions partially match both of the delineation styles, thus allowing the decision-maker to choose one segmentation from five possibilities between two extreme styles of delineation.

The Pareto front approximations (represented by soft Dice loss) from different approaches for ten randomly selected validation samples are shown in Figure~\ref{fig:mo seg results}. It can be seen for all the approaches that, while the Pareto front approximations seem to be optimal for some validation samples, it is not true for all the samples. This highlights the unavoidable generalization gap on the validation samples and the shortcoming of the existing as well as our proposed approach in meeting the goal of MO decision-making per sample during inference.

The hypervolume values of the Pareto front approximation on validation samples from 50 Monte Carlo cross-validation runs with a 80:20 split are reported in Table \ref{tab:HV_mo_segmentation}. A t-test was chosen because the distribution of mean HVs was approximately normal. The network initializations were different in each run. The hypervolume was observed to be maximal for our proposed HV maximization-based approach indicating that the Pareto fronts approximations on validation samples were closer to the Pareto front when using the HV maximization approach. The differences in hypervolume values were statistically significant between  Pareto MTL and HV maximization ($t(49)=-8.350, p=5.59e^{-11}$), and EPO and HV maximization ($t(49)=-3.066, p=0.004$), but not between linear scalarization and HV maximization ($t(49)=-0.915, p=0.365$). It may be attributed to the fact that the Pareto front is strictly convex and symmetrical for all validation samples. On the one hand, symmetric Pareto fronts are well-suited for linear scalarization (in line with results shown in Section~\ref{sec:sin_cos_asymmetric}), and on the other hand, the noise in HV maximizing gradients due to batchwise training might have limited the performance of HV maximization approach. 

\begin{table}[]
\renewcommand{\arraystretch}{1.5}
\centering
\caption{\small{Mean $\pm$ standard deviation hypervolume of the approximated Pareto fronts by different approaches for validation samples from 50 Monte Carlo cross-validation runs. The maximal mean hypervolume is \textbf{highlighted}. Statistical significance using one-sided paired t-tests is indicated for post-hoc comparisons: LS vs HV maximization (\textsuperscript{$\ast$}), PMTL vs HV maximization (\textsuperscript{$\dagger$}), and EPO vs HV maximization (\textsuperscript{$\ddagger$}). The relevance of small increases in HV close to the maximum (400) is explained in Section~\ref{sec:sin_cos_asymmetric}.
}}
\label{tab:HV_mo_segmentation}
\begin{tabular}{rl}
\toprule
\textbf{} & \textbf{Hypervolume}\\
\midrule
\small Linear scalarization (LS) & \footnotesize 396.6268 $\pm$ 0.6108\\
\small Pareto MTL (PMTL) & \footnotesize 396.2791 $\pm$ 0.6128\\
\small EPO & \footnotesize 396.4999 $\pm$ 0.6700\\
\small HV maximization & \footnotesize \textbf{396.6778 $\pm$ 0.5777\textsuperscript{$\dagger \ddagger$}}\\
\bottomrule
\end{tabular}
\end{table}

\subsection{Neural Style Transfer}
\label{sec:style_transfer}
We further apply our approach to the problem of style transfer, i.e., the transfer of the artistic style of an image onto a target image while preserving its semantic content. Users likely cannot provide their preferred trade-off between style and content without seeing the resulting images. Providing an approximation of Pareto front is thus a useful tool in aiding decision-making.

Contrary to MO regression (Section~\ref{sec:sin_cos}) and MO segmentation (Section~\ref{sec:mo_segmentation}), this is not a MO learning but a MO optimization problem using neural networks with differently shaped Pareto fronts per sample. We selected the problem definition by \cite{gatys2016image}, where pixels of an image are optimized to minimize a weighted combination of content loss (semantic similarity with the target image) and style loss (artistic similarity with the style image). The content loss and the style loss are computed from features of a pretrained VGG network \citep{Simonyan15}. We reused and adjusted Pytorch's neural style transfer implementation \citep{pytorch_style_tutorial}. The reference point was chosen as (100, 10000) based on preliminary runs. 
The results presented below are based on 25 image pairs, obtained by combining 5 content and 10 style images. All images or their references are available in Appendix~\ref{suppl:image_details}, most were collected from WikiArt \citep{wikiart} and are in the public domain or available under fair use.

\begin{figure*}[h!]
\begin{subfigure}{0.45\textwidth}
    \centering
    \includegraphics[width=\textwidth]{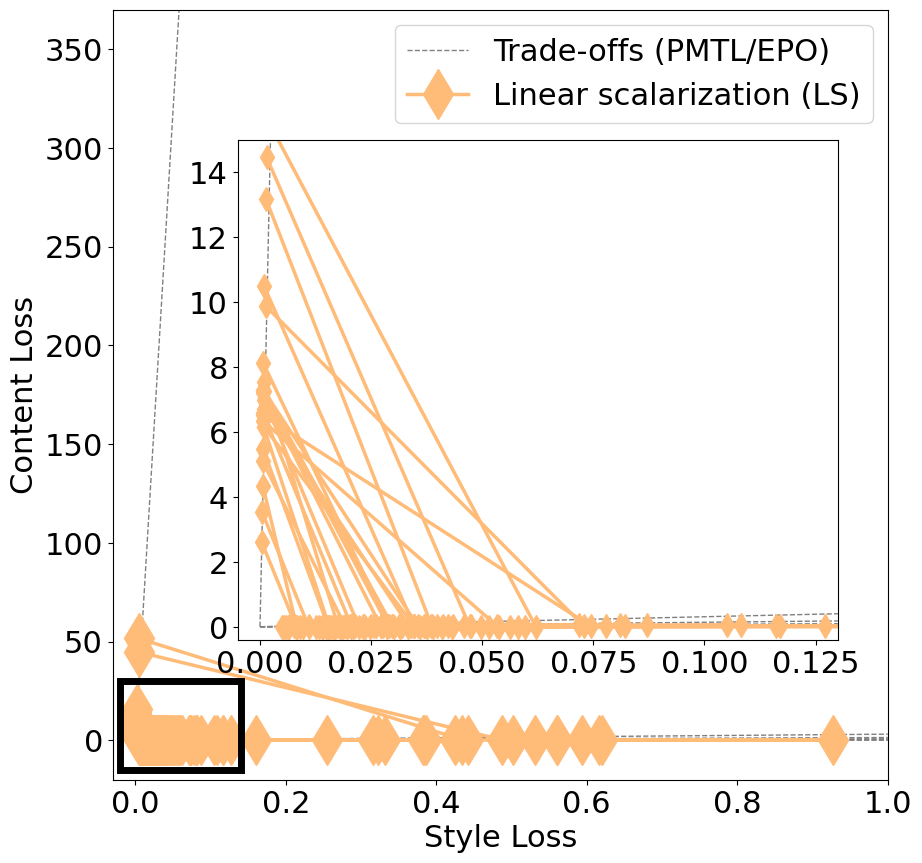}
    \caption{}
    \label{fig:2d_style_transfer_comparison_multi_fig_lin_scal}
\end{subfigure}
\begin{subfigure}{0.45\textwidth}
    \centering
    \includegraphics[width=\textwidth]{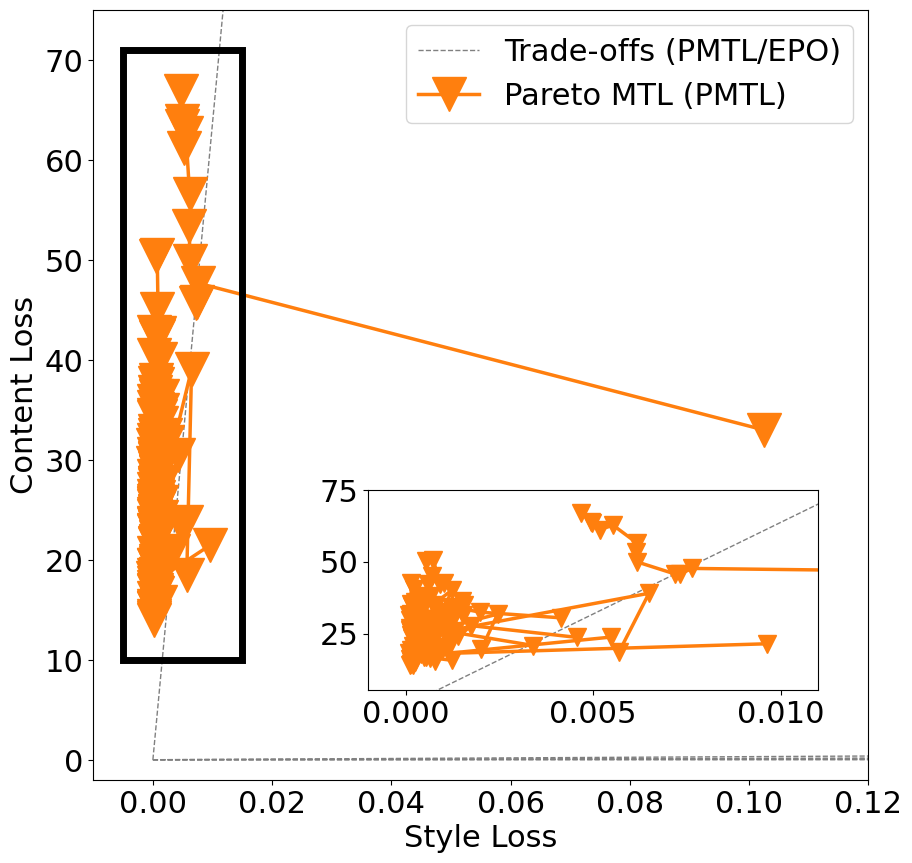}
    \caption{}
    \label{fig:2d_style_transfer_comparison_multi_fig_pareto_mtl}
\end{subfigure}
\begin{subfigure}{0.45\textwidth}
    \centering
    \includegraphics[width=\textwidth]{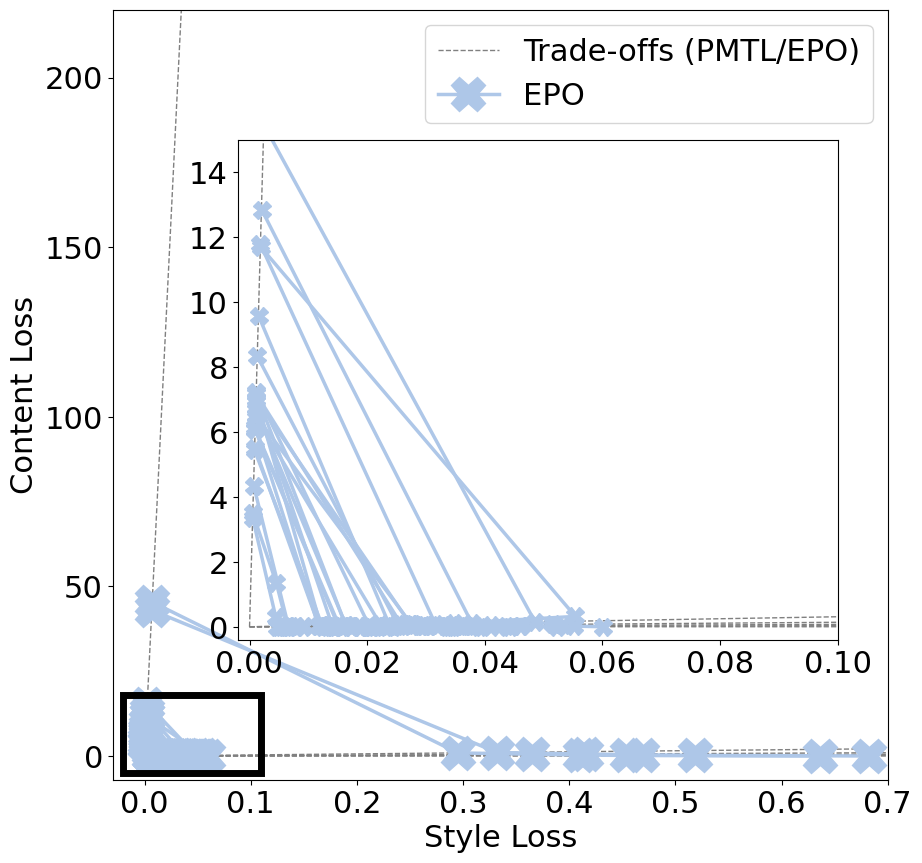}
    \caption{}
    \label{fig:2d_style_transfer_comparison_multi_fig_epo}
\end{subfigure}
\begin{subfigure}{0.45\textwidth}
    \centering
    \includegraphics[width=\textwidth]{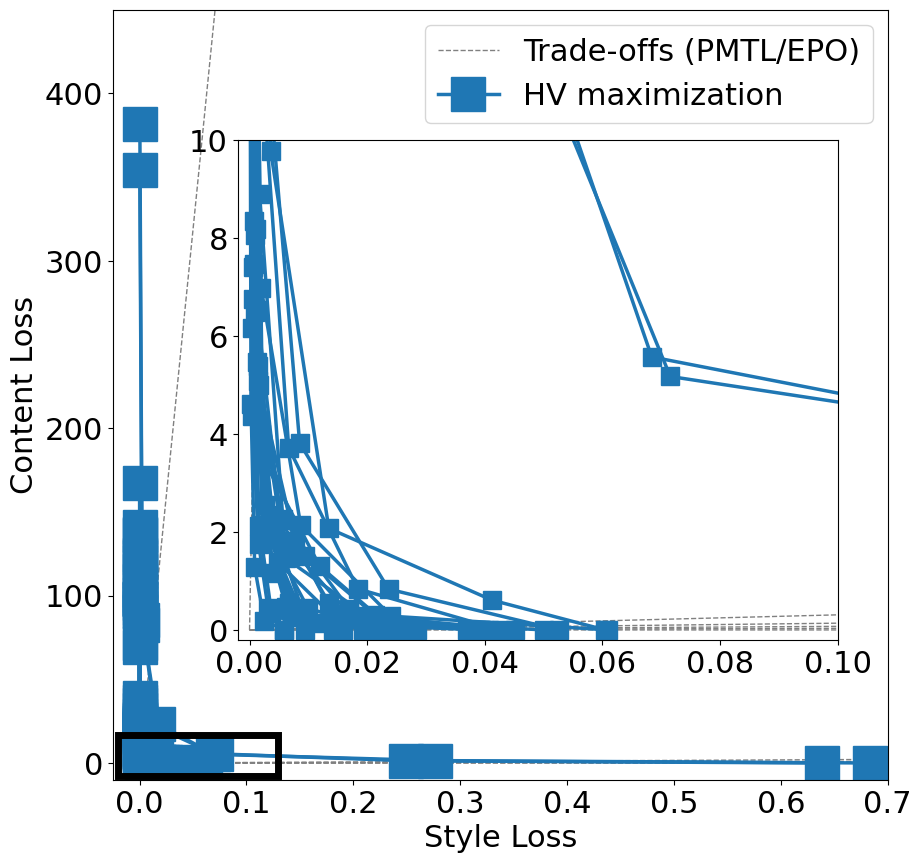}
    \caption{}
    \label{fig:2d_style_transfer_comparison_multi_fig_higamo_hv}
\end{subfigure}
    \centering
    \caption{Pareto front estimates in loss space by different approaches for neural style transfer using four approaches: (a) Linear scalarization (b) Pareto MTL, (c) EPO, and (d) HV maximization. Sections within the black frames are magnified.}
    \label{fig:2d_style_transfer_comparison_multi_fig}
\end{figure*}
Figure~\ref{fig:2d_style_transfer_comparison_multi_fig} shows the obtained Pareto front estimates for 25 image sets by each approach. Linear scalarization (a) and EPO (c) determine solutions close to or on the chosen user-preferences which, however, do not diversely cover the range of possible trade-offs. Pareto MTL (b) achieves sets of clustered and partly dominated solutions which do not cover trade-offs with low content loss. On the other hand, HV maximization (d) returns Pareto front estimates that broadly cover trade-offs across image sets without having to specify user preferences, which is also reflected in the significantly larger median HVs reported in Table~\ref{tab:HV_style_transfer}. As noted in Section~\ref{sec:sin_cos_asymmetric}, the magnitude of the reported HVs is due to the choice of reference point and already small increases in the HV can indicate a relevant improvement in Pareto front approximation quality.
These results show that estimates based on pre-specified trade-offs may not span the Pareto front well. HV maximization appears to find better estimates when the shape of the Pareto front is unknown, which enables a posteriori decision-making.  

\begin{figure*}[h!]
\centering
\includegraphics[width=1.00\textwidth]{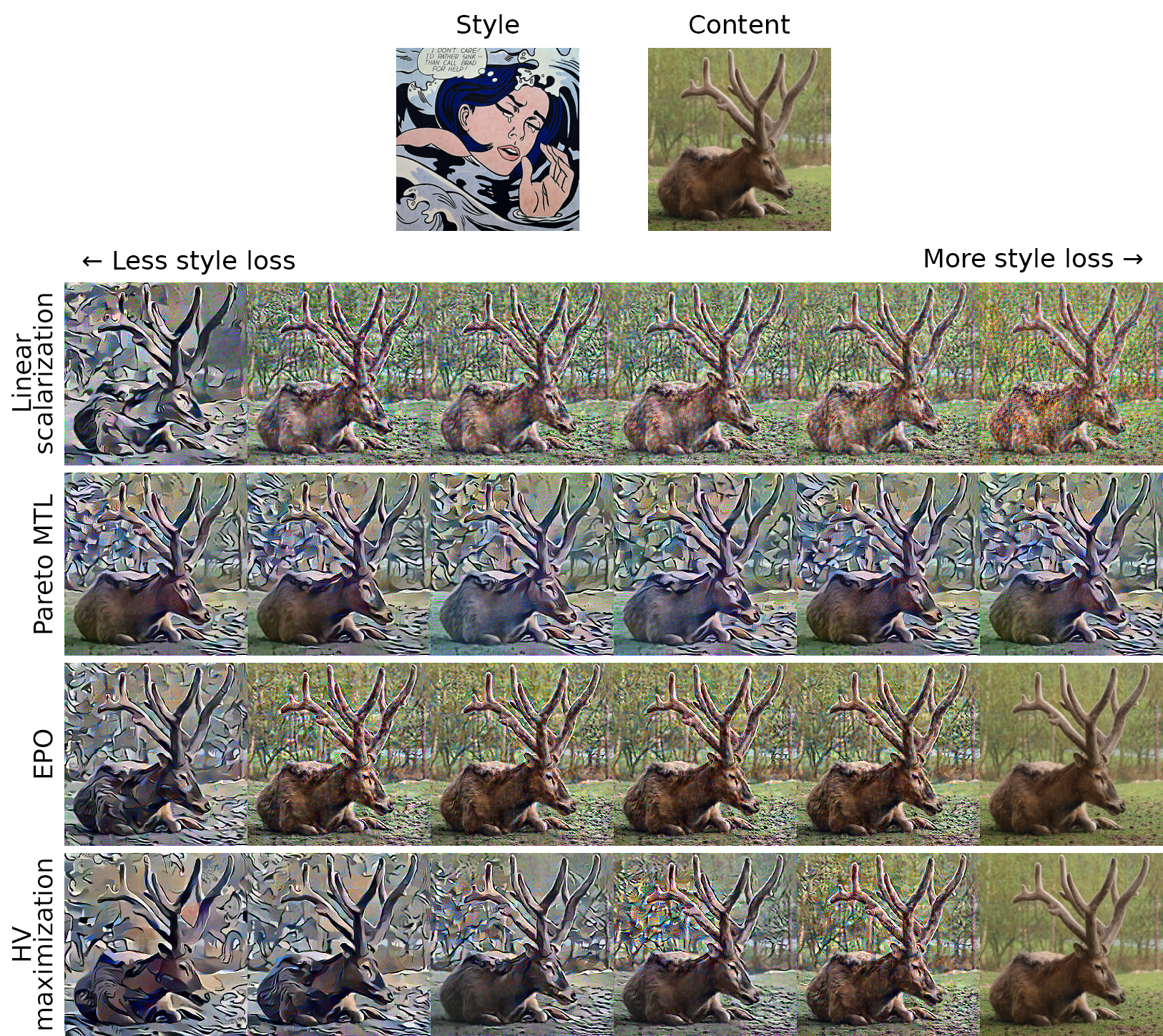}
\caption{Images generated by all four approaches for image set B19.} 
\label{fig:style_transfer_2d_example_corrected}
\end{figure*}

Figure~\ref{fig:style_transfer_2d_example_corrected} shows the images generated by each approach for one of the sets used in Figure~\ref{fig:2d_style_transfer_comparison_multi_fig}. This case (B19, see Appendix~\ref{suppl:image_details} for image set definitions) was manually selected for its aesthetic appeal.\footnote{Generated images for all image sets B1-B25 are available at \url{https://github.com/timodeist/multi_objective_learning}}
The images seen here match observations from Figure~\ref{fig:2d_style_transfer_comparison_multi_fig}, e.g., Pareto MTL's images show little diversity in style and content, many images by linear scalarization of EPO have too little style match (`uninteresting' images), and images by HV maximization show most interesting diversity.

\begin{table}[]
\renewcommand{\arraystretch}{1.5}
\centering
\caption{\small{Median (inter-quartile range) hypervolume of the approximated Pareto fronts by different approaches for 25 image sets of neural style transfer, each time using a different content and style image. The maximal median hypervolume is \textbf{highlighted}. Statistical significance in one-sided Wilcoxon signed rank test after multiple comparisons correction is indicated for comparisons: LS vs HV maximization (\textsuperscript{$\ast$}), PMTL vs HV maximization (\textsuperscript{$\dagger$}), and EPO vs HV maximization (\textsuperscript{$\ddagger$}).
The relevance of small increases in HV close to the maximum ($10^{6}$) is described in Section~\ref{sec:sin_cos_asymmetric}.
}}
\label{tab:HV_style_transfer}
\begin{tabular}{rl}
\toprule
\textbf{} & \textbf{Hypervolume}\\
\midrule
\small Linear scalarization (LS) & \footnotesize 999990.7699 (999988.6580 -- 999992.5850)\\
\small Pareto MTL (PMTL) & \footnotesize 997723.8748 (997583.5152 -- 998155.6837)\\
\small EPO & \footnotesize 999988.4297 (999984.4808 -- 999989.8338)\\
\small HV maximization & \footnotesize \textbf{999999.7069 (999999.4543 -- 999999.8266)\textsuperscript{$\ast \dagger \ddagger$}}\\
\bottomrule
\end{tabular}
\end{table}

\begin{figure}[h!]
\centering
\includegraphics[width=0.8\textwidth]{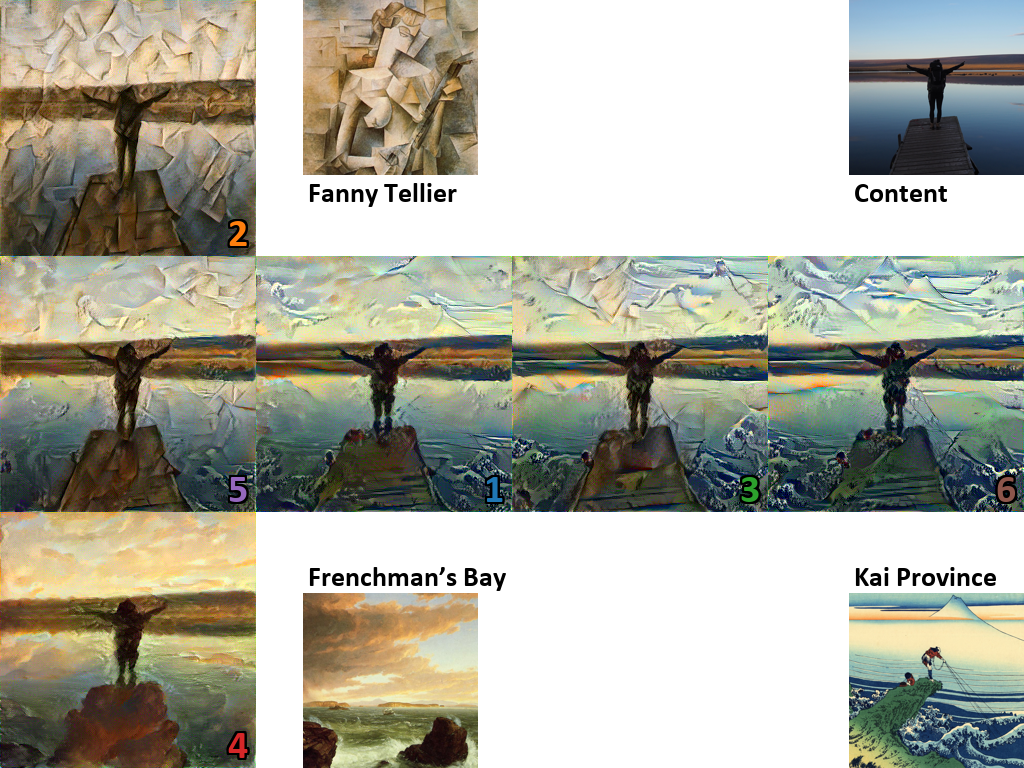} 
\caption{Neural multi-style transfer as an example for HV optimization in three (style) dimensions. The T-shape approximately reflects their style loss ordering. For example, the Frenchman's Bay loss of Image~4 is lower than of Image~5 and Image~2.}
\label{fig:style_transfer_tshape}
\end{figure}

\subsubsection{Neural multi-style transfer}
\begin{figure}[!]
\centering
\includegraphics[width=0.48\textwidth]{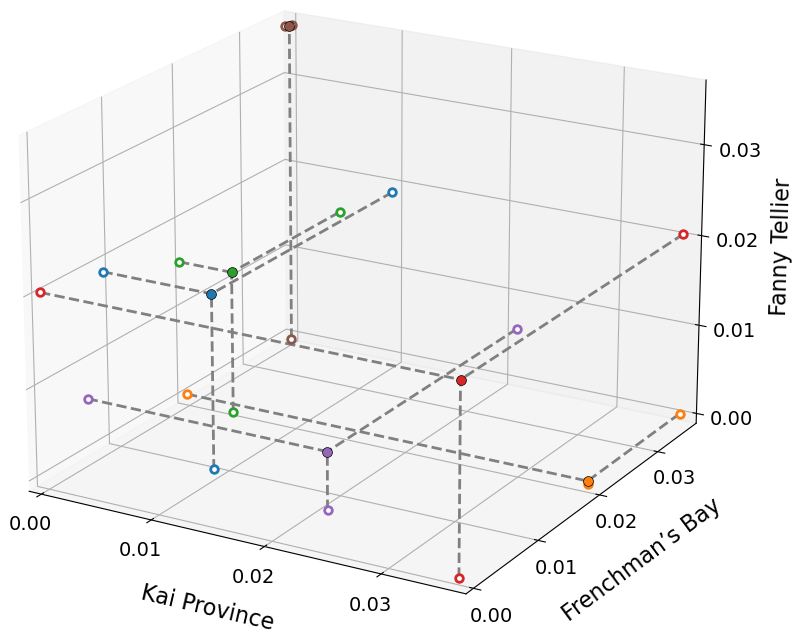}
\caption{Neural multi-style transfer Pareto front approximation in loss space. Each filled circle corresponds to the image with a matching colored number in Figure~\ref{fig:style_transfer_tshape}.}
\label{fig:style_transfer_objective_space}
\end{figure}

Trading-off content against one or more style losses results in many `uninteresting' images with visually irrelevant changes in content loss (Figure~\ref{fig:style_transfer_2d_example_corrected}). To improve neural style transfer by removing these undesirable images and thereby demonstrating the uses of HV maximization over three objectives, we removed the content loss, defined style losses for three different style images, initialized the optimization on the content image, and ensured that sufficient original content is retained by tuning (limiting) the number of optimization iterations.  Figure~\ref{fig:style_transfer_tshape} shows the Pareto front approximation with six images after HV maximization. This example was selected for its aesthetic appeal. Three solutions are close to the distinct artistic styles, and the others are mixes of different styles with trade-offs between the style losses. Viewing the images in loss space (Figure \ref{fig:style_transfer_objective_space}) demonstrates that the images are diverse and clearly dispersed from each other.
\section{Discussion}
\label{sec:discussion}

We adapted the gradient-based HV maximization approach from MO optimization for the goal of training a set of neural networks so that they jointly predict Pareto front approximations for each sample during inference, without prior need for user-specified trade-offs. We further show that training for Pareto front approximations of average losses is, in general, not sufficient to attain well-spread approximations on convex or concave Pareto fronts of individual samples. 

The performance of our approach is demonstrated on two MO learning cases and one neural style transfer optimization case. The experimental results show that, while existing approaches perform similar to HV maximization for problems that exhibit a symmetric Pareto front, a priori specifying the trade-offs fails when the Pareto front is asymmetric and consequently causes worse performance. In contrast, our HV maximization approach still finds well-spread solutions on these asymmetric Pareto fronts. Furthermore, we have shown that the issue of a priori specified trade-offs not being able to cover the entire Pareto front in case of asymmetric Pareto front shapes cannot be solved trivially by rescaling the losses based on their initial magnitude.
It should also be noted that multi-objective machine learning problems in real-world are likely not symmetric. Our HV-based approach therefore is preferable over methods relying on trade-offs specified a priori.

HV maximization does not require specifying $p$  trade-offs a priori (based on the number of predictions, $p$, required on the Pareto front), which essentially are $p(n-1)$ hyperparameters of the learning process for $n$ losses. Choosing these trade-offs well requires knowledge of the Pareto front shapes which is often not known a priori. HV maximization, however, introduces the $n$-dimensional reference point $r$ and thus $n$ additional hyperparameters. Luckily, choosing a reference point such that the entire Pareto front is approximated by well-spread predictions is not complex. It often suffices to use losses of randomly initialized networks rescaled by a factor $\geq$1 as the reference point. For the special case that only a specific section of the Pareto front is relevant and this information is known a priori, the reference point can be chosen so that the Pareto front approximation only spans the chosen section.

The HV-based training for set of neural networks can, in theory, be applied to any number of networks, $p$, and loss functions, $n$. In practice, the time complexity of exact HV (exponential in $n$, \cite{fonseca2006improved}) and HV gradient (quadratic in $p$ with $n\leq 4$, \cite{emmerich2014time}) computations is limiting but may be overcome by algorithmic improvements using, e.g., HV approximations. Further, in our current implementation, a separate network is trained corresponding to each prediction. This increases computational load linearly if more predictions on the Pareto front are desired. We chose for this setup for the sake of simplicity in experimentation and clarity when demonstrating our proof-of-concept. It is expected that the HV maximization formulation would work similarly if the parameters of some of the neural network layers are shared, which would decrease computational load.

When training networks on multiple losses using the preferred per-sample Dynamic loss~\eqref{eq:final_joint_loss}, a network is not restricted to generate predictions in a specific order of trade-offs for all samples. Instead, predictions for different samples can follow any order of networks on the samples' Pareto fronts as observed in Figure~\ref{fig:sincos_2d_predictions}. Limiting this flexibility might simplify the learning problem and speed up convergence without significantly decreasing prediction quality.

In conclusion, the present work describes a method for and advantages of learning-based a posteriori MO decision-making based on MO training of neural networks using HV maximization. We provided a detailed analysis of MO learning for different Pareto front shapes using an artificial MO regression problem. Additionally, the MO segmentation (Section~\ref{sec:mo_segmentation}) and MO style transfer problem (Section~\ref{sec:style_transfer}) yielded encouraging results that emphasize the real-world relevance of the proposed method. Future work should further investigate the usability of HV maximization in more complex learning problems on real-world data.

\acks{ We would like to thank dr. Marco Virgolin from Chalmers University of Technology for his valuable contributions and discussions on concept and code.
 The research is part of the research programme, Open Technology Programme with project number 15586, which is financed by the Dutch Research Council (NWO), Elekta, and Xomnia. Further, the work is co-funded by the public-private partnership allowance for top consortia for knowledge and innovation (TKIs) from the Ministry of Economic Affairs.}


\newpage

\vskip 0.2in
\bibliography{hv_nn_bibliography}

\newpage

\appendix
\renewcommand\thefigure{\thesection \arabic{figure}}
\renewcommand\thetable{\thesection \arabic{table}}
\section{MO Regression}
\label{suppl:sin_cos}
\setcounter{figure}{0}
\setcounter{table}{0}
The MO regression case with two losses in Section~4.1 is now extended by the following target variable
\begin{equation*}
Y_{3}=\sin(X+\pi)
\end{equation*}
with corresponding loss function $L_{3} = (y^{(3)}_{k}-z_{k})^{2}$. Results, shown in Figures~\ref{fig:sincos_3d_losses}-\ref{fig:sincos_3d_os}, show similar behavior as for two losses. A noteworthy exception is that Pareto fronts never reduce to a single point (Figure~\ref{fig:sincos_3d_os}) because $\sin(X)$, $\cos(X)$, and $\sin(X+\pi)$ never coincide (Figure~\ref{fig:sincos_3d_predictions})

\begin{figure*}[h!]
\begin{subfigure}{0.32\textwidth}
    \centering
    \includegraphics[width=\textwidth]{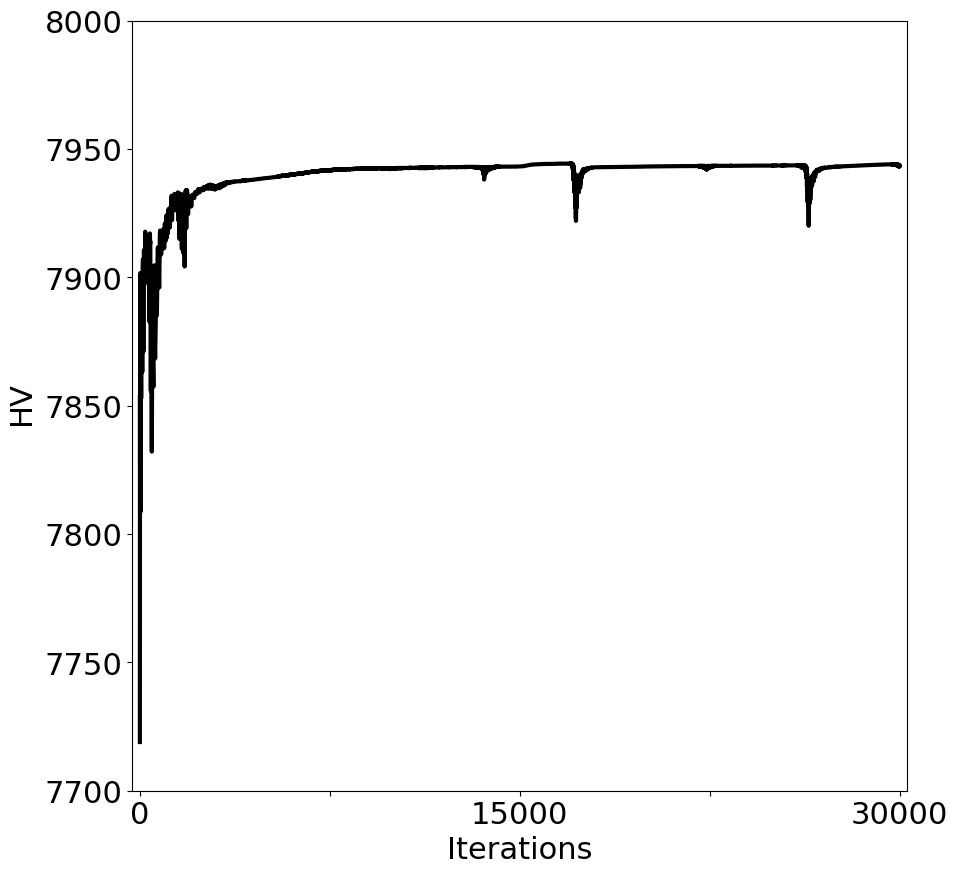}
    \caption{}
    \label{fig:sincos_3d_losses}
\end{subfigure}
\begin{subfigure}{0.32\textwidth}
    \centering
    \includegraphics[width=\textwidth]{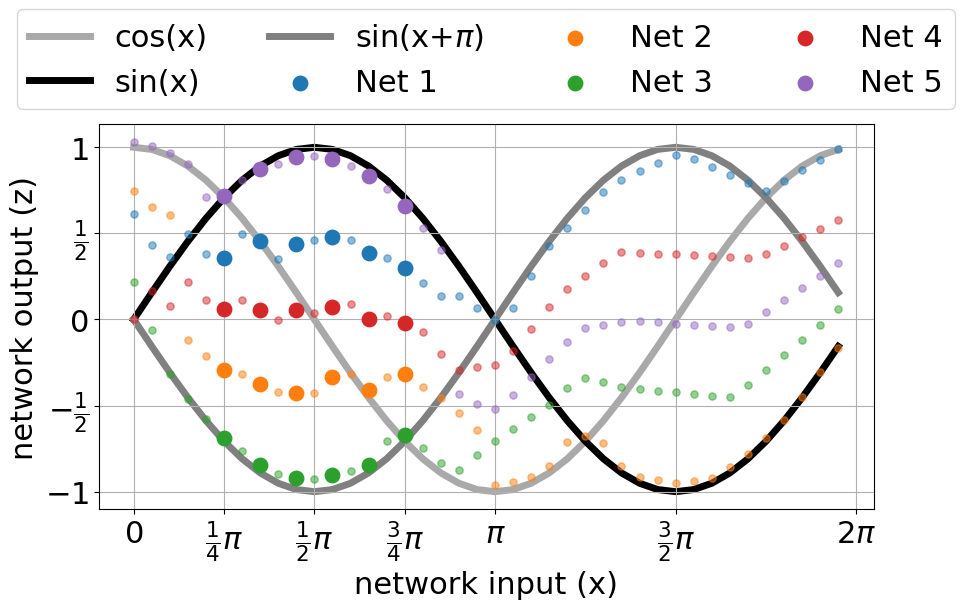}
    \caption{}
    \label{fig:sincos_3d_predictions}
\end{subfigure}
\begin{subfigure}{0.32\textwidth}
    \centering
    \includegraphics[width=\textwidth]{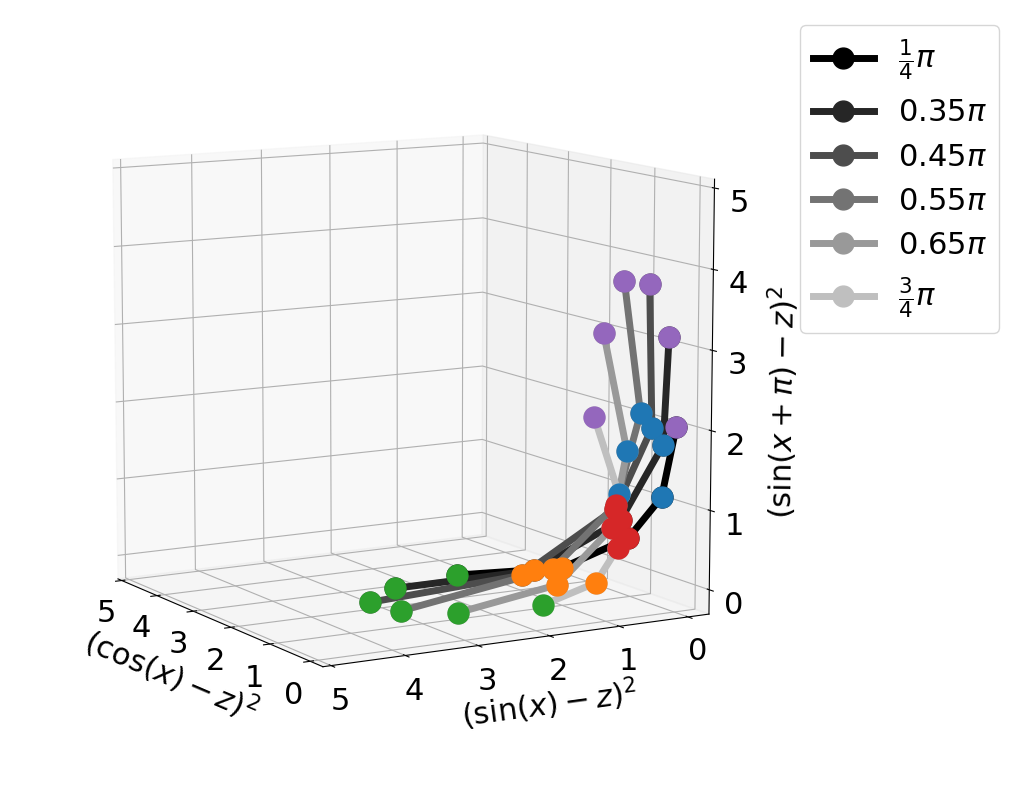}
    \caption{}
    \label{fig:sincos_3d_os}
\end{subfigure}
    \centering
    \caption{Multi-objective regression on three losses. (a) HV for set of networks over training iterations (b) Network outputs for $X\in[0,2\pi]$ (c) Generated Pareto front predictions for six samples selected from $[\tfrac{1}{4}\pi,\tfrac{3}{4}\pi]$ in loss space.}
    \label{fig:sincos_3d}
\end{figure*}

\section{Hyperparameter Tuning}
\label{suppl:tuning}
\setcounter{figure}{0}
\setcounter{table}{0}
Hyperparameters used in MO regression, multi-observer medical image segmentation, and neural style transfer were tuned using grid search. The used grids and chosen settings are shown in Table~\ref{tab:tuning_info}.
Hyperparameter grid tuning for style transfer is performed on five image pairs, each containing one style and one content image. The hyperparameter setting with maximal HV after 10000 optimization iterations was selected. On 3D neural style transfer, only HV maximization was applied. Thus, no tuning results are displayed for the other approaches. The images shown in Figure~\ref{fig:style_transfer_tshape} were optimized for 2500 iterations instead of for 5000 iterations used for the grid search because 2500 is sufficient to yield pleasing results and to retain more content in the image.
\begin{landscape}
\begin{table*}
\caption{Chosen hyperparameter settings and tuning grids. $\gamma$: learning rate. $\beta_{1}$: coefficient for computing moving average of gradients in Adam, $\lambda$: weight decay.}
\label{tab:tuning_info}
\small
\begin{tabular}{cccccccccccccccccc}

& & \multicolumn{6}{c}{\textbf{MO regression}}  & \multicolumn{3}{c}{\textbf{Segmentation}} &\multicolumn{4}{c}{\textbf{Neural style transfer}} & \multicolumn{1}{c}{\textbf{Loss rescaling}}\\
& & \multicolumn{2}{c}{\thead{\textbf{MSE \&}\\ \textbf{MSE}}} & \multicolumn{2}{c}{\thead{\textbf{MSE \&}\\ \textbf{L1-Norm}}} & \multicolumn{2}{c}{\thead{\textbf{MSE \&}\\ \textbf{Scaled MSE}}} & & & & \multicolumn{2}{c}{\textbf{2D}} & \multicolumn{2}{c}{\textbf{3D}} & \multicolumn{1}{c}{\,} \\
\toprule
\multirow{5}{0.2em}{\rotatebox{90}{Best settings}}& & $\gamma$ & $\beta_{1}$ & $\gamma$ & $\beta_{1}$ & $\gamma$ & $\beta_{1}$ & $\gamma$ & $\beta_{1}$ & $\lambda$ & $\gamma$ & $\beta_{1}$ & $\gamma$ & $\beta_{1}$ & $\gamma$\\
& \textbf{LS} & $10^{-4}$ & 0.99 & $10^{-4}$ & 0.99 & $10^{-4}$ & 0.9 & $10^{-4}$ & 0.9 & $10^{-5}$ & $10^{-2}$ & 0.999 & - & - & $10^{-3}$ \\
& \textbf{PMTL} & $10^{-2}$ & 0.9 & $10^{-2}$ & 0.99 & $10^{-4}$ & 0.99 & $10^{-4}$ & 0 & $10^{-5}$ & $10^{-2}$ & 0.5 & - & - & $10^{-4}$\\
& \textbf{EPO} & $10^{-5}$ & 0.9 & $10^{-5}$ & 0.99 & $10^{-5}$ & 0.5 & $10^{-4}$ & 0.9 & $10^{-5}$ & $10^{-3}$ & 0.99 & - & - & $10^{-4}$\\
& \textbf{HV max.} & $10^{-3}$ & 0.5 & $10^{-3}$ & 0.99 & $10^{-3}$ & 0.99 & $10^{-4}$ & 0.5 & $10^{-5}$ & $10^{-3}$ & 0.99 & $10^{-2}$ & 0.99 & $10^{-4}$\\
\midrule
\multirow{4}{0.2em}{\rotatebox{90}{Grid search}}& \textbf{$\gamma$-grid} 
& \multicolumn{6}{c}{$10^{-\{1,2,3,4,5\}}$}
& \multicolumn{3}{c}{$10^{-\{3,4,5\}}$}
& \multicolumn{2}{c}{$10^{-\{1,2,3\}}$}
& \multicolumn{2}{c}{$10^{-\{1,2,3\}}$}
& \multicolumn{1}{c}{$10^{-\{1,2,3,4,5\}}$}\\
& \textbf{$\beta_{1}$-grid} & \multicolumn{6}{c}{$\{0.5,0.9,0.99\}$} & \multicolumn{3}{c}{$\{0,0.5,0.9\}$} & \multicolumn{2}{c}{$\{0.5,0.9,0.99,0.999\}$} & \multicolumn{2}{c}{$\{0.9,0.99,0.999\}$} & \multicolumn{1}{c}{$0.9$} \\
& \textbf{$\lambda$-grid} 
& \multicolumn{6}{c}{0} 
& \multicolumn{3}{c}{$10^{-\{4,5\}}$}
& \multicolumn{2}{c}{0} 
& \multicolumn{2}{c}{0} & 0\\
& \thead{\textbf{Grid search}\\\textbf{iterations}} & \multicolumn{6}{c}{$2\times 10^{4}$} & \multicolumn{3}{c}{$3\times 10^{3}$} & \multicolumn{2}{c}{$10^{4}$} & \multicolumn{2}{c}{$5\times 10^{3}$} & \multicolumn{1}{c}{$10^{4}$}\\ \bottomrule

\end{tabular}

\begin{tabular}{ccccc}
& & \multicolumn{3}{c}{\textbf{Section 3.3}} \\
& & \multicolumn{1}{c}{\thead{\textbf{Strictly}\\\textbf{convex}}} & \multicolumn{1}{c}{\textbf{Linear}} & \multicolumn{1}{c}{\thead{\textbf{Non-}\\\textbf{convex}}}\\
\toprule
\multirow{6}{0.2em}{\rotatebox{90}{Best settings}} & & $\gamma$ & $\gamma$ & $\gamma$ \\
& \textbf{LS}                       & $10^{-2}$ & $10^{-2}$ & $10^{-5}$ \\
& \textbf{PMTL}                     & $10^{-2}$ & $10^{-4}$ & $10^{-1}$ \\
& \textbf{EPO}                      & $10^{-3}$ & $10^{-4}$ & $10^{-2}$ \\
& \textbf{HV max. (per sample)}     & $10^{-2}$ & $10^{-4}$ & $10^{-3}$ \\
& \textbf{HV max. (average losses)} & $10^{-3}$ & $10^{-4}$ & $10^{-3}$ \\
\midrule
\multirow{4}{0.2em}{\rotatebox{90}{Grid search}}& \textbf{$\gamma$-grid} & 
\multicolumn{3}{c}{$10^{-\{1,2,3,4,5\}}$} \\
& \textbf{$\beta_{1}$-grid} & \multicolumn{3}{c}{$0.9$} \\
& \textbf{$\lambda$-grid} & \multicolumn{3}{c}{0} \\
& \thead{\textbf{Grid search}\\\textbf{iterations}} & \multicolumn{3}{c}{$10^{4}$} \\ \bottomrule

\end{tabular}
\end{table*}

\end{landscape}

\begin{landscape}
\section{Source image details for style transfer}
\label{suppl:image_details}
\begin{table}[h!]
\caption{Description of images used for neural style transfer. Image sets are labelled with a prefix and number pair. Sets with prefix A are used for 2D hyperparameter tuning, sets with prefix B are used for 2D result generation, etc. 
}
\label{tab:style_transfer_image_info}
\resizebox{0.90\columnwidth}{!}{
\begin{tabular}{clllcccc}
& & & & \multicolumn{4}{c}{\textbf{Contained in image sets}} \\			
&	\textbf{Title} &	\textbf{Artist} &	\textbf{Link} &	\thead{\textbf{2D}\\ \textbf{tuning}} &	\thead{\textbf{2D}\\ \textbf{results}} &	\thead{\textbf{3D}\\ \textbf{tuning}} &	\thead{\textbf{3D}\\ \textbf{results}} \\
\midrule
\multirow{19}{0.001em}{\rotatebox{90}{\textbf{Style}}} &	The Starry Night &	Vincent van Gogh &	\href{https://www.wikiart.org/en/vincent-van-gogh/the-starry-night-1889}{WikiArt} &	A1 &	 &	C1 &	\\
&	Dwelling in the Fuchun Mountains &	Huang Gongwang &	\href{https://en.wikipedia.org/wiki/File:DwellingInTheFuchun.jpg}{Wikipedia} &	A2 & &		C1 & \\	
&	Supremus 55 &	Kazimir Malevich &	\href{https://commons.wikimedia.org/wiki/File:Supremus_55_(Malevich,_1916).jpg}{Wikimedia} &	A3 & &		C1 &	\\
 &	The Great Wave off Kanagawa	& Katsushika Hokusai &	\href{https://www.wikiart.org/en/katsushika-hokusai/the-great-wave-of-kanagawa-1831}{WikiArt} &	A4 & &		C2 & \\	
 &	The Shipwreck &	William Turner &	\href{https://www.wikiart.org/en/william-turner/shipwreck}{WikiArt} &	A5 & &		C2 & \\	
 &	Figure dans un Fauteuil &	Pablo Picasso &	\href{https://www.wikiart.org/en/pablo-picasso/seated-female-nude-1910}{WikiArt} &	 & &		C2 &	\\
 &	Femme au Chapeau &	Henri Matisse &	\href{https://www.wikiart.org/en/henri-matisse/woman-with-hat-1905}{WikiArt} &	 & &		C3 & \\	
 &	Marilyn Monroe &	Andy Warhol &	\href{https://www.wikiart.org/en/andy-warhol/marilyn-1}{WikiArt} &	 & &		C3 & \\	
 &	The Scream &	Edvard Munch &	\href{https://www.wikiart.org/en/edvard-munch/the-scream-1893}{WikiArt} &	 & &		C3 & \\	
 &	Red Vineyards at Arles &	Vincent van Gogh &	\href{https://commons.wikimedia.org/wiki/File:Red_vineyards.jpg}{Wikimedia} &		 & B1,11,21 & &		D1\\
 &	Lofty Mount Lu &	Shen Zhou &	\href{https://www.wikiart.org/en/shen-zhou/lofty-mount-lu-1467}{WikiArt} & &		B6,16,26 & &		D1 \\
 &	Proun 19D &	El Lissitzky &	\href{https://www.wikiart.org/en/el-lissitzky/proun-19d-1922}{WikiArt} & &		B2,12,22,27 & &		D1 \\
 &	Kajikazawa in Kai Province & Katsushika Hokusai &	\href{https://www.wikiart.org/en/katsushika-hokusai/kajikazawa-in-kai-province}{WikiArt} & &		B7,17 & &		D2\\
 &	View Across Frenchman`s Bay &	Thomas Cole & \href{https://www.wikiart.org/en/thomas-cole/view-across-frenchman-s-bay-from-mount-desert-island-after-a-squall-1845}{WikiArt} & &		B3,13,23 & &		D2\\
 &	Girl with mandolin (Fanny Tellier) &	Pablo Picasso &	\href{https://www.wikiart.org/en/pablo-picasso/girl-with-mandolin-fanny-tellier-1910}{WikiArt} & &		B8,18,28 & &		D2\\
 &	Portrait of Matisse &	Andr\'{e} Derain &	\href{https://www.wikiart.org/en/andre-derain/portrait-of-matisse-1905}{WikiArt} & &		B4,14,24,29 & &		D3\\
 &	Drowning Girl &	Roy Lichtenstein &	\href{https://www.wikiart.org/en/roy-lichtenstein/drowning-girl-1963}{WikiArt} & &		B9,19 & &		D3\\
 &	Anxiety &	Edvard Munch &	\href{https://www.wikiart.org/en/edvard-munch/anxiety-1894}{WikiArt} & &		B5,15,25 & &		D3\\
 & The Wanderer Above the Sea of Fog & Caspar D. Friedrich & \href{https://www.wikiart.org/en/caspar-david-friedrich/the-wanderer-above-the-sea-of-fog}{WikiArt} & & B10,20,30 & & \\ 
\midrule							
\multirow{8}{0.001em}{\rotatebox{90}{\textbf{Content}}} &	Neckarfront in Tübingen &	Andreas Praefcke &	\href{https://commons.wikimedia.org/wiki/File:Tuebingen_Neckarfront.jpg}{Wikimedia} &	A1,2 & &		C1 & \\	
 &	Golden Gate Bridge &	Rich Niewiroski Jr. &	\href{https://commons.wikimedia.org/wiki/File:GoldenGateBridge-001.jpg}{Wikimedia} &	A3,4 & &		C2 & \\	
 &	Lenna &	Dwight Hooker, Playboy &	\href{https://en.wikipedia.org/wiki/File:Lenna_(test_image).png}{Wikipedia} &	A5 & &		C3 & \\	
 &	New York Skyline &	Petr Kratochvil &	\href{https://www.publicdomainpictures.net/en/view-image.php?image=207924\&picture=new-york-skyline}{Publicdomainpictures} & &		B1-5 & &		D1\\
 &	Sitojaure &	Frank Dankers & \href{https://github.com/timodeist/multi_objective_learning}{repository} & &			B6-10 & &		D2 \\
 &	Richard Feynman &	Tamiko Thiel &	\href{https://commons.wikimedia.org/wiki/File:RichardFeynman-PaineMansionWoods1984_copyrightTamikoThiel_bw.jpg}{Wikimedia}  & &		B11-15 & &		D3\\
 
 &	Deer &	Hans Dankers & \href{https://github.com/timodeist/multi_objective_learning}{repository} & &			B16-20 & & \\		
 &	Dolomites &	Frank Dankers & \href{https://github.com/timodeist/multi_objective_learning}{repository} & &			B21-25 & & \\	
\bottomrule
\end{tabular}
}
\end{table}
\FloatBarrier
\end{landscape}

\section{Loss rescaling}
\label{suppl:loss_rescale}
\setcounter{figure}{0}
\setcounter{table}{0}
To show that a simple rescaling of loss functions prior to training is not sufficient to ensure a diverse approximation of the Pareto front given a set of trade-offs or fixed linear scalarizations chosen a priori, we apply such rescaling on the neural style transfer optimization problem of Section~4.3\footnote{The described rescaling method may however work on trivial cases as in Figures~\ref{fig:comparison_ls_sin_cos_scaledmse}-\ref{fig:comparison_higamo_hv_sin_cos_scaledmse} that were downscaled to study and explain the behavior of algorithms on asymmetric fronts.} The content loss is rescaled by a factor $w$ determined on tuning sets. $w$ is determined by averaging ratio of style and content losses at initialization of the optimization,$w = \tfrac{1}{Mp}\sum_{m=1}^{M}\sum_{i=1}^{p}\tfrac{L_{1}(\theta_{1},s_{m})}{L_{2}(\theta_{1},s_{m})}$, for $M=5$ tuning sets (A1-A5, Table~\ref{tab:style_transfer_image_info}). This factor is stable across those five tuning sets with mean 0.0108, standard deviation 0.0027, and range 0.0074-0.0158. The same factor should therefore be usable for other image sets.
Learning rate $\gamma$ has been tuned on tuning sets A1-A5 for each approach. The solver used by EPO has been changed from GLPK to ECOS in this experiment because GLPK failed in some instances.
Figure~\ref{fig:2d_style_transfer_rescaled_comparison_multi_fig} shows results following the setup of Figure~\ref{fig:2d_style_transfer_comparison_multi_fig} for a smaller selection of sets (B1-B5 plus B19 for visual comparison to Figure~\ref{fig:style_transfer_2d_example_corrected}). Linear scalarization and EPO determine extreme solutions either exhibiting low style loss or low content loss. In contrast, Pareto MTL returns more diverse solutions and HV maximization clearly succeeds at generating diverse solution for each image set. Figure~\ref{fig:style_transfer_rescaled_example} shows the resulting images for image set B19. In comparison with Figure~\ref{fig:style_transfer_2d_example_corrected}, Figure~\ref{fig:style_transfer_rescaled_example} shows that loss rescaling before optimization also visually does not improve solutions of existing methods (LS, Pareto MTL, EPO). The images generated by HV maximization show a visually less pronounced, and thus less desirable, change in style intensity which might be resolved by shifting the reference point.
As can be observed, the scale of losses at initialization does not provide sufficient information to adjust the MO problem so that existing methods reliably yield diverse approximations of Pareto fronts. On the other hand, we show here that HV maximization can achieve a set of diverse solutions with and without loss rescaling at initialization.
Certainly, this basic rescaling procedure could be improved by rescaling losses with an updated factor $w_{t}$ at every iteration $t$. This would, however, result in another complex dynamic loss function with unclear advantages over the presented HV maximization.

\begin{figure*}[h!]
\begin{subfigure}{0.45\textwidth}
    \centering
    \includegraphics[width=\textwidth]{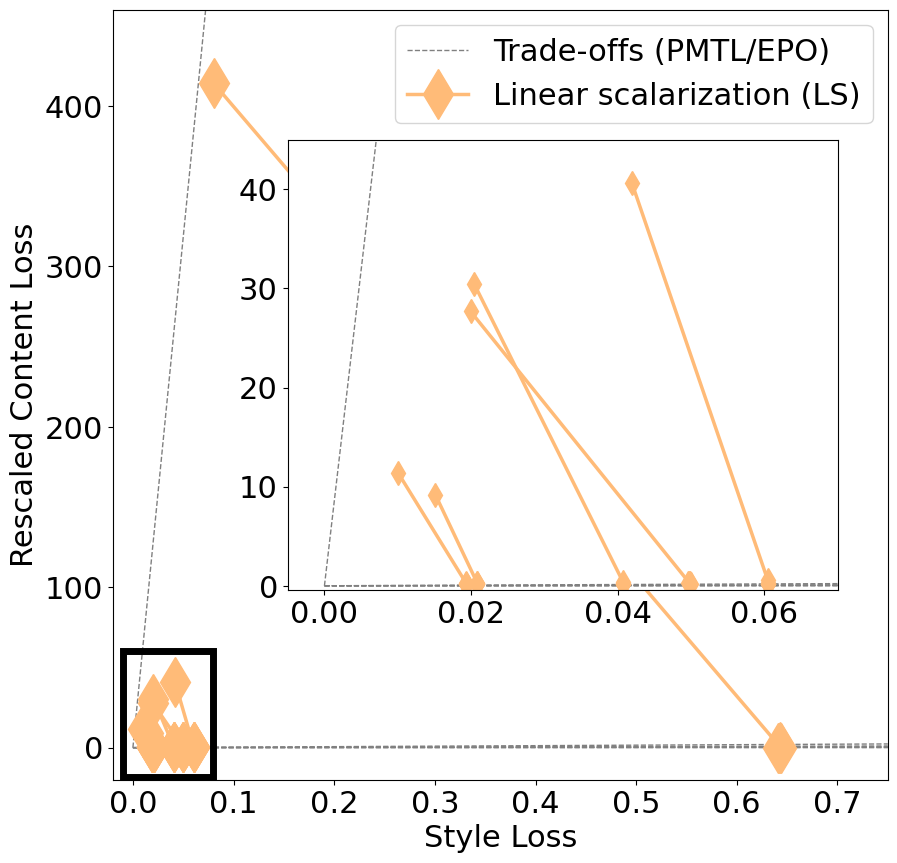}
    \caption{}
    \label{fig:2d_style_transfer_rescaled_comparison_multi_fig_lin_scal}
\end{subfigure}
\begin{subfigure}{0.45\textwidth}
    \centering
    \includegraphics[width=\textwidth]{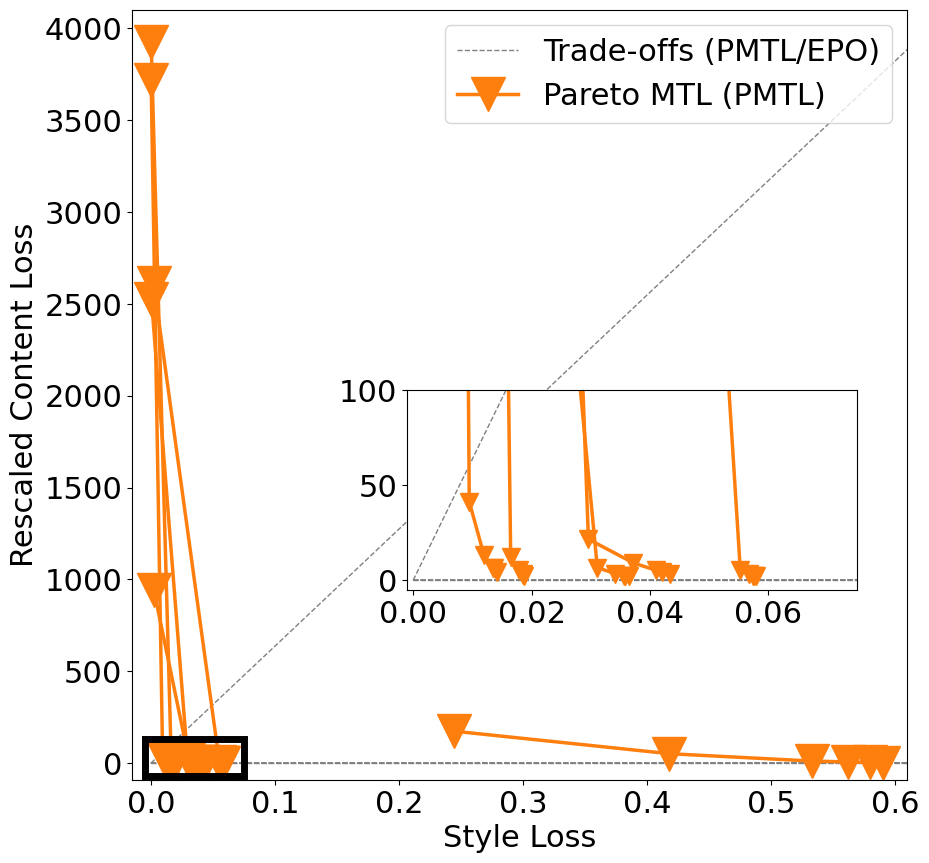}
    \caption{}
    \label{fig:2d_style_transfer_rescaled_comparison_multi_fig_pareto_mtl}
\end{subfigure}
\begin{subfigure}{0.45\textwidth}
    \centering
    \includegraphics[width=\textwidth]{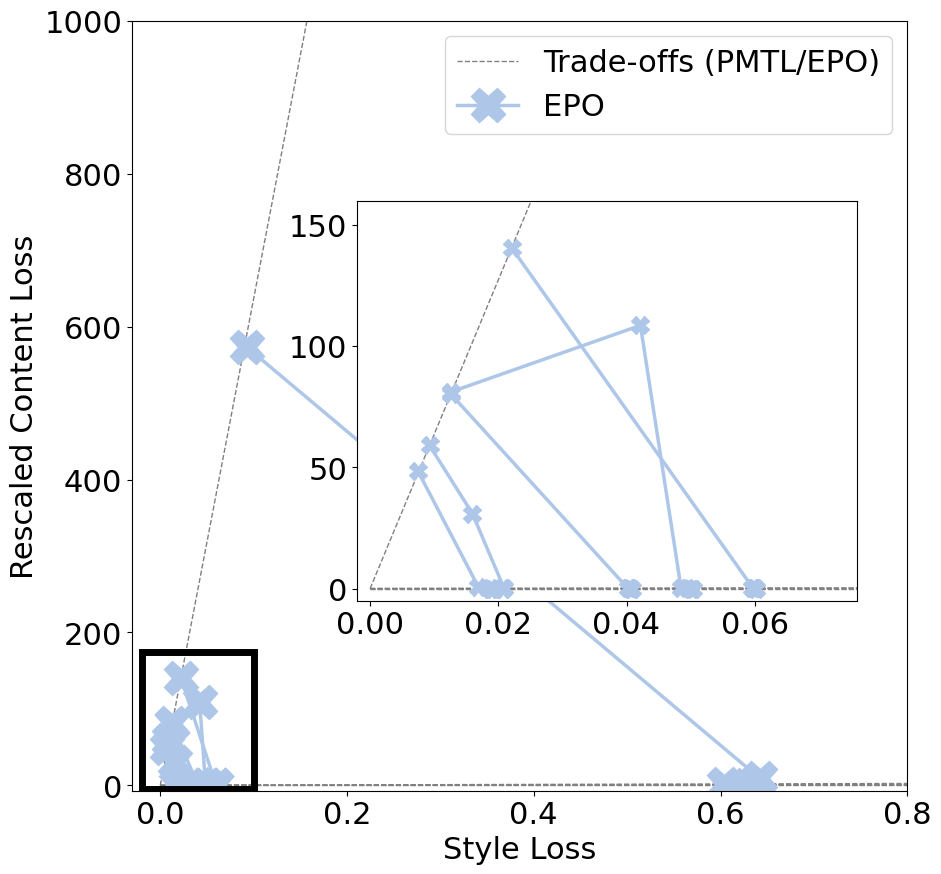}
    \caption{}
    \label{fig:2d_style_transfer_rescaled_comparison_multi_fig_epo}
\end{subfigure}
\begin{subfigure}{0.45\textwidth}
    \centering
    \includegraphics[width=\textwidth]{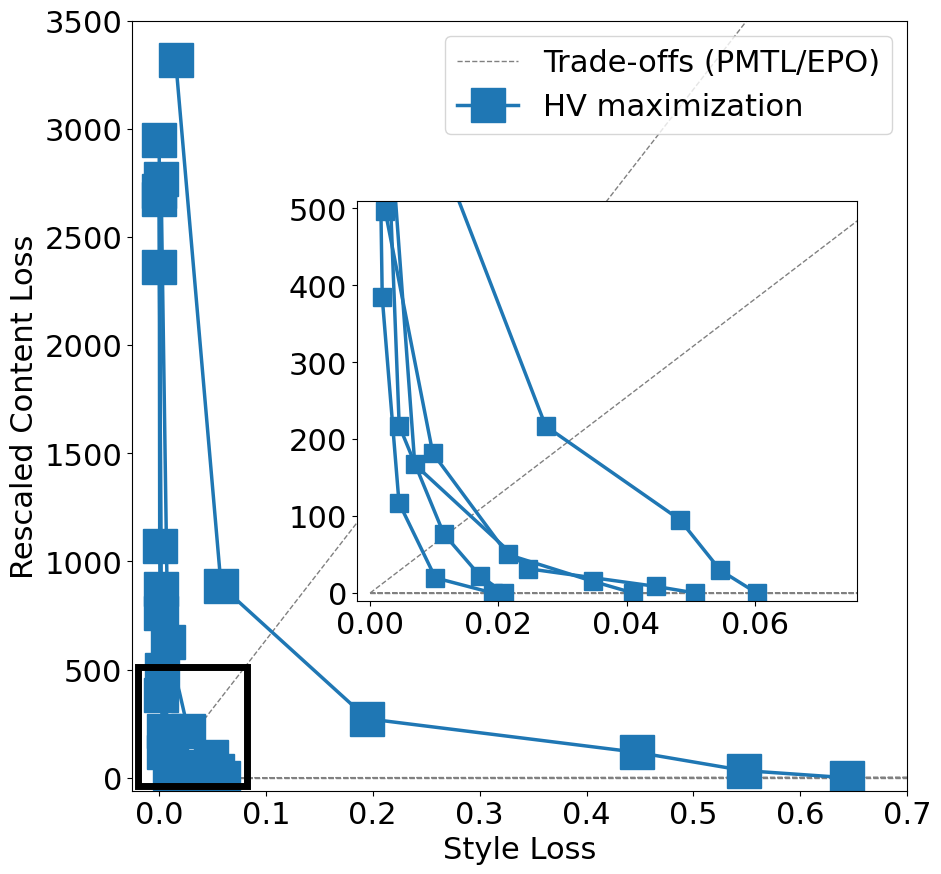}
    \caption{}
    \label{fig:2d_style_transfer_rescaled_comparison_multi_fig_higamo_hv}
\end{subfigure}
    \centering
    \caption{Pareto front estimates in loss space for neural style transfer, with content loss rescaled by the estimated ratio of losses at initialization. Four approaches were used on six image sets: (a) Linear scalarization (b) Pareto MTL, (c) EPO, and (d) HV maximization. Sections within the black frames are magnified.}
    \label{fig:2d_style_transfer_rescaled_comparison_multi_fig}
\end{figure*}

\begin{figure*}[h!]
\centering
\includegraphics[width=1.00\textwidth]{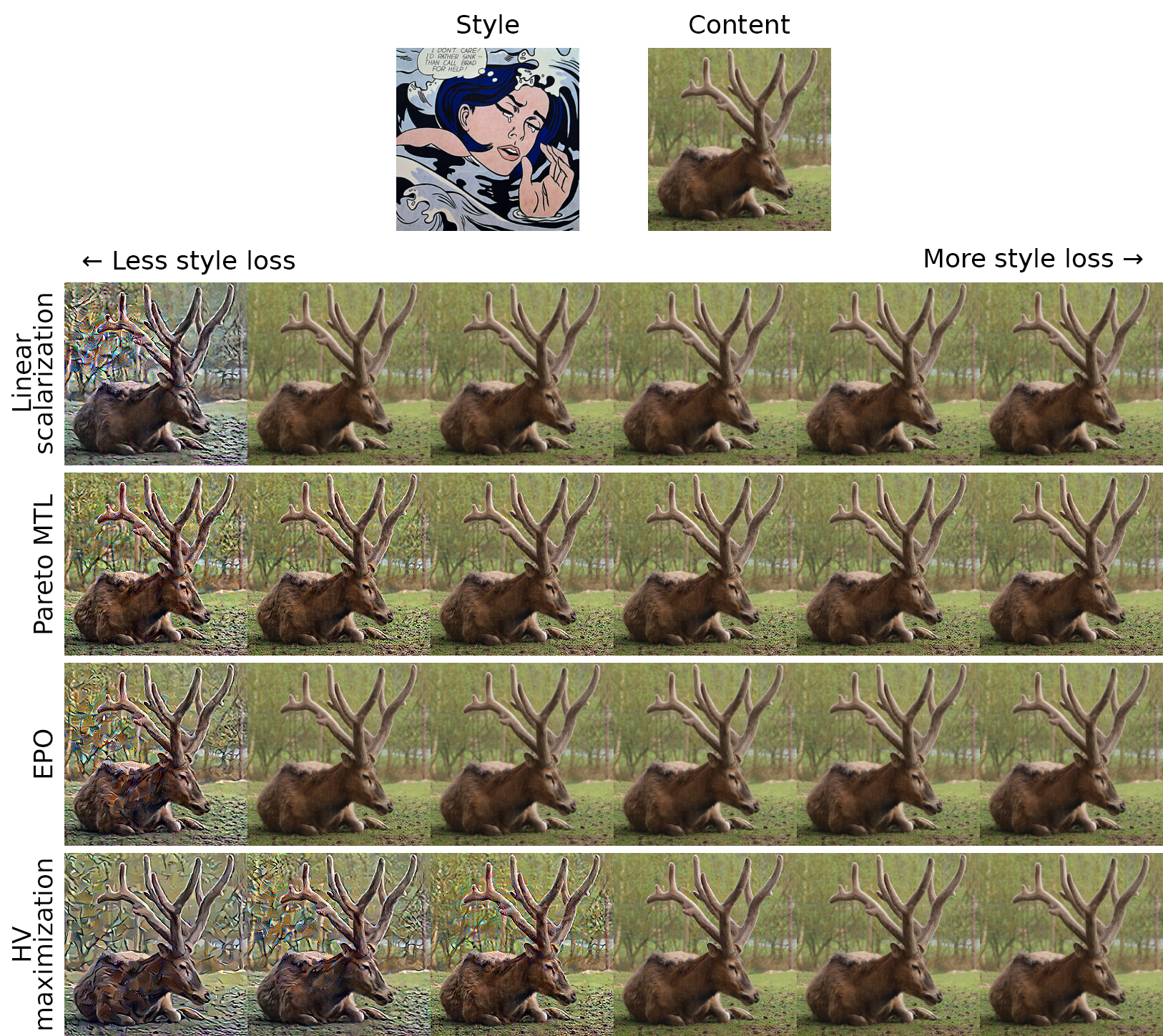}
\caption{Images generated for image set B19 with loss rescaling using all four approaches.} 
\label{fig:style_transfer_rescaled_example}
\end{figure*}



\end{document}